\documentclass{article} 

\usepackage{xspace}
\newcommand{\framework}{\textsc{Eureka}\xspace}  
\newcommand{\collection}{\textsc{Eureka-Bench}\xspace}

\newcommand{\ClaudeSonnet}{{Claude 3.5 Sonnet}\xspace}  
\newcommand{\ClaudeOpus}{{Claude 3 Opus}\xspace}  
\newcommand{\GeminiPro}{{Gemini 1.5 Pro}\xspace}  
\newcommand{\LlamaThree}{{Llama 3 70B}\xspace}  
\newcommand{\LlamaThreeOne}{{Llama 3.1 70B}\xspace}  
\newcommand{\LlamaThreeOneLarge}{{Llama 3.1 405B}\xspace}  
\newcommand{\MistralLargeTwo}{{Mistral Large 2407}\xspace}  
  
\newcommand{\GPTFourPrev}{{GPT-4 1106 Preview}\xspace}  
\newcommand{\GPTFourO}{{GPT-4o 2024-05-13}\xspace}  
\newcommand{\GPTFourVisionPreview}{{GPT-4 Vision Preview}\xspace} 
\newcommand{\GPTFourTurboApril}{{GPT-4 Turbo 2024-04-09}\xspace} 
\newcommand{\Llava}{{Llava 1.6 34B}\xspace}

\date{}
\usepackage[hyphens]{url}
\usepackage{hyperref}
\usepackage{times}

\hypersetup{colorlinks=true,
breaklinks=true,
allcolors={blue}}
\usepackage{array} 

\usepackage{graphicx}
\usepackage{framed}
\usepackage{comment}
\usepackage{hyperref}
\usepackage{array}
\usepackage{multirow}
\usepackage{booktabs}
\usepackage{enumitem}
\usepackage[letterpaper,top=2cm,bottom=2cm,left=3cm,right=3cm,marginparwidth=1.75cm]{geometry}
\usepackage{subfigure}
\usepackage{xcolor}
\definecolor{cadmiumgreen}{rgb}{0.2, 0.82, 0.24}
\definecolor{forestgreen}{rgb}{0.13, 0.55, 0.13}
\definecolor{redfrontier}{HTML}{B50575}
\definecolor{greenfrontier}{HTML}{188814}
\definecolor{bluefrontier}{HTML}{4C66DD}

\begin{document}

\title{\textsc{Eureka}: Evaluating and Understanding \\ Large Foundation Models}
\author{Vidhisha Balachandran\quad
 Jingya Chen\quad
 Neel Joshi\quad
  Besmira Nushi\quad \\
  Hamid Palangi\quad
  Eduardo Salinas\quad
  Vibhav Vineet\quad   \\
  James Woffinden-Luey\quad
  Safoora Yousefi\quad
  \\
  {Microsoft Research}
\vspace{0.4em}
  \\
  {\small 
  \includegraphics[width=1.2em, trim=0 0 0 0, clip]{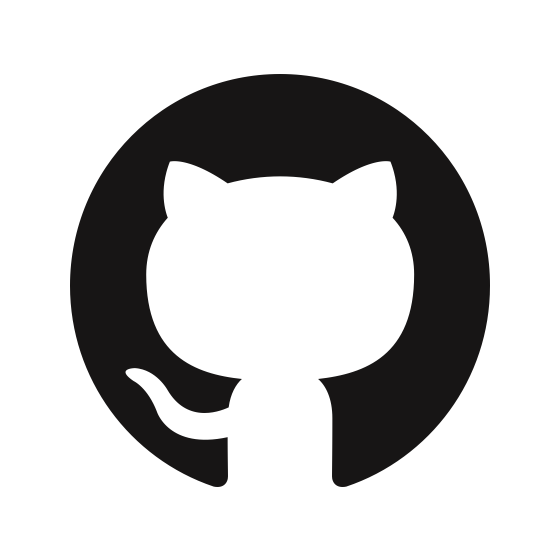} Code: \texttt{\url{https://github.com/microsoft/eureka-ml-insights}}
  } 
}
\maketitle

\begin{abstract}
Rigorous and reproducible evaluation of large foundation models is critical for assessing the state of the art, informing next steps in model improvement, and for guiding scientific advances in Artificial Intelligence (AI). Evaluation is also important for informing the increasing number of application developers that build services on foundation models. The evaluation process has however become challenging in practice due to several reasons that require immediate attention from the community, including benchmark saturation, lack of transparency in the methods being deployed for measurement, development challenges in extracting the right measurements for generative tasks, and, more generally, the extensive number of capabilities that need to be considered for showing a well-rounded comparison across models. In addition, despite the overwhelming numbers of side-by-side capability evaluations available, we still lack a deeper understanding about when and how different models fail for a given capability and whether the nature of failures is similar across different models being released over time.

We make three contributions to alleviate the above challenges. First, we present \framework, a reusable and open evaluation framework for standardizing evaluations of large foundation models beyond single-score reporting and rankings. Second, we introduce \collection as an extensible collection of benchmarks testing capabilities that (i) are still challenging for state-of-the-art foundation models and (ii) represent fundamental but overlooked capabilities for completing tasks in both language and vision modalities. The available space for improvement that comes inherently from non-saturated benchmarks, enables us to discover meaningful differences between models at a capability level. Third, using the framework and \collection, we conduct an analysis of 12 state-of-the-art models, providing in-depth insights for failure understanding and model comparison by disaggregating the measurements across important subcategories of data. Such insights uncover granular weaknesses of models for a given capability and can then be further leveraged to plan more precisely on what areas are most promising for improvement. \framework is available as open-source to foster transparent and reproducible evaluation practices. 

In contrast to recent trends in evaluation reports and leaderboards showing absolute rankings and claims for one model or another to be the best, our analysis shows that there is no such best model. Different models have different strengths, but there are models that appear more often than others as best performers for several capabilities. Despite the many observed improvements, it also becomes obvious that current models still struggle with a number of fundamental capabilities including detailed image understanding, benefiting from multimodal input when available rather than fully relying on language, factuality and grounding for information retrieval, and over refusals.

\end{abstract}
\renewcommand{\thefootnote}{\fnsymbol{footnote}}
\footnotetext[2]{Correspondence to \texttt{\{benushi,neel,sayouse,vidhishab,vivineet\}@microsoft.com}.}
\footnotetext[3]{Currently at Google. Work done while at Microsoft Research.}
\clearpage
\renewcommand{\thefootnote}{\arabic{footnote}}

{
  \hypersetup{linkcolor=black}
  \tableofcontents
}
\clearpage
\section{Introduction}
The evaluation of Large Foundation Models (LFMs) presents several methodical and practical challenges, many of which stem from the generative and general-purpose nature of recent models. The rapid progress in AI has also introduced many new capabilities as part of the model skills portfolio, which need to be assessed alongside traditional capabilities. \framework is a framework and a collection of challenging benchmarks that aims at scaling up such evaluations for LFMs in an open and transparent manner. The framework itself provides a library for flexibly customizing evaluation pipelines that combine a series of components necessary for evaluation including data preprocessing, prompt templates, model inference, data postprocessing, metric computation, and reporting. \collection is the collection of benchmarks whose implementation is currently supported in \framework and for which we provide extensive evaluation and analysis reports.

\begin{table}[htbp]
\footnotesize
\centering
\begin{tabular}{ccccc}
\toprule
\textbf{Modality} &  
\textbf{\begin{tabular}[c]{@{}c@{}}Benchmark \\ \#prompts\end{tabular}} & 
\textbf{Capability} &
\textbf{\begin{tabular}[c]{@{}c@{}}Experimental Conditions\end{tabular}} & 
\textbf{Subcategories} \\ \midrule

Image $\rightarrow$ Text &  
\begin{tabular}[c]{@{}c@{}} GeoMeter \\ 1086 
\end{tabular} &
Geometric Reasoning &
 & \begin{tabular}[c]{@{}c@{}}Depth\\Height\end{tabular}
\\ \hline

Image $\rightarrow$ Text & 
\begin{tabular}[c]{@{}c@{}} MMMU \\ 900 
\end{tabular}  &
Multimodal QA &
&
\begin{tabular}[c]{@{}c@{}}Disciplines\\ Subjects\end{tabular} 
\\ \hline

Image $\rightarrow$ Text &
\begin{tabular}[c]{@{}c@{}} Image Understanding \\ 10,240

\end{tabular}  & 
\begin{tabular}[c]{@{}c@{}}Object Recognition \\ Object Detection\\ Visual Prompting \\ Spatial Reasoning \end{tabular}    &
\begin{tabular}[c]{@{}c@{}}Single Object\\ Two Objects\end{tabular} 
&  \begin{tabular}[c]{@{}c@{}}Object Recognition \\ Object Detection\\ Visual Prompting \\ Spatial Reasoning \end{tabular}   
\\ \hline

Image $\rightarrow$ Text &
\begin{tabular}[c]{@{}c@{}} Vision Language \\Understanding \\ 13,500 
\end{tabular}  &
\begin{tabular}[c]{@{}c@{}}Spatial Understanding\\ Navigation\\ Counting \end{tabular}   &
\begin{tabular}[c]{@{}c@{}}Image Only \\ Text Only\\ Image and Text \end{tabular}  &
\begin{tabular}[c]{@{}c@{}}Spatial Map\\ Maze Navigation\\ Object Counting \end{tabular}  
\\ \hline

Text $\rightarrow$ Text &
\begin{tabular}[c]{@{}c@{}} IFEval \\ 541 
\end{tabular}  &
Instruction Following &
&
Instruction Category  \\ \hline

Text $\rightarrow$ Text &
\begin{tabular}[c]{@{}c@{}} FlenQA \\ 12,000 
\end{tabular}  &
Long Context multi-hop QA&
\begin{tabular}[c]{@{}c@{}} Context Length \\ Info Placement \end{tabular} &
\begin{tabular}[c]{@{}c@{}} Monotonic Relations\\ People in Rooms \\ Ruletaker \end{tabular}   
\\ \hline

Text $\rightarrow$ Text &
\begin{tabular}[c]{@{}c@{}} Kitab \\ 34,217 
\end{tabular}  &
Information Retrieval &
\begin{tabular}[c]{@{}c@{}} Context Availability \\ Constraint Count \end{tabular} &
\begin{tabular}[c]{@{}c@{}} Constraint type \\ Author popularity \\ Query constrainedness \end{tabular}  
\\ \hline

Text $\rightarrow$ Text &
\begin{tabular}[c]{@{}c@{}} Toxigen \\ 10,500 
\end{tabular}  &
\begin{tabular}[c]{@{}c@{}} Toxicity Detection \\ Safe Language Generation\end{tabular}&
\begin{tabular}[c]{@{}c@{}} Discriminative \\ Generative \end{tabular} &
\begin{tabular}[c]{@{}c@{}} Demographic groups \end{tabular} \\ 
\bottomrule
\end{tabular}
\caption{Benchmarks currently available in \textsc{Eureka-Bench}.}
\label{tab:all_benchmarks}
\end{table}
\noindent {\bfseries Evaluation Framework.} The evaluation process for complex and generative capabilities has made traditional practices for evaluation obsolete. For example, the concept of a fixed, closed-form definition of a metric does not apply anymore to many capabilities either because several different sub metrics need to be computed before reaching a final score, or because many data transformations and answer extraction operations need to be applied to the output prior to computing a metric. Some of these data transformations often are custom to the model being evaluated. In addition, part of the evaluation also needs to be handed over to other model judges for scaling up~\cite{chang2024survey,zheng2024judging}. This new landscape leaves practitioners with the necessity of creating a rich combination of data processing steps, code execution, and model inferences as evaluators, all in function of producing a final score for the model under test. \framework provides a flexible library for composing these functionalities into shareable evaluation pipelines and gives full control to practitioners to handle and log the details of each experiment. These functionalities also enable reproducibility and backtracking of experimentation details (e.g. prompt templates, inference parameters, API and model versions) in a transparent manner. Given that such details can change measurements significantly, we believe it is important for the research community to have access to both the code and logs behind evaluations.  Thus, we provide the actual code and logs used in evaluations.

\noindent {\bfseries Benchmark selection.} A pressing issue in the evaluation of state-of-the-art LFMs is the fact that many of the benchmarks commonly reported in technical reports of model releases are either close to \emph{saturation} or already saturated, where models have reached close to 100\% accuracy on the benchmarks. For example, several recent models~\cite{openai2024gpt4,dubey2024llama,ClaudeSonnet,reid2024gemini,MistralLarge2} have been reported to have an accuracy higher than 85\% on benchmarks like MMLU~\cite{HendrycksBBZMSS21}, GSM8K~\cite{cobbe2021training}, HumanEval~\cite{chen2021evaluating}, DROP~\cite{DuaWDSS019}, BigBench-Hard~\cite{srivastava2022beyond}, MGSM~\cite{ShiSF0SVCTRZ0W23}, ChartQA~\cite{MasryLTJH22}, AI2D~\cite{KembhaviSKSHF16}. Many of the benchmarks in this list represent tasks that were important for testing fundamental capabilities at the time of their release. However, saturation of performance does not leave ample space for discovering major failure modes and for comparing different models. While saturation itself may originate either from inherent model improvements or from memorization, the challenge from a scientific communication perspective is reflected as lack of clarity in quantifying and characterizing improvements over time. In fact, many of the recent improvements in such benchmarks can be as small as 1-2 percentage points, which leaves one wondering whether we are indeed experiencing a true more general saturation in terms of progress in AI or whether the benchmarks and methods being used for measurement are insufficient. 

To create space for deeper analysis, the benchmarks in \collection (Table~\ref{tab:all_benchmarks}) are chosen such that either the whole benchmark, or an important experimental condition within that benchmark, remains challenging for even the most capable models. As a simplified rule of thumb, we choose to include in \collection benchmarks for which overall model performance (or performance on an important experimental condition) is less than 80\% for either all models or at least roughly half of the models studied in this work.

Another consideration for benchmark selection is capability and modality coverage. While \collection is not an exhaustive list of capabilities, it aims at covering  diverse fundamental language and multimodal capabilities that are currently overlooked in traditional evaluations but that are critical for more complex tasks. For example, spatial and geometric understanding are not evaluated often in recent reports but they are fundamental to real-world tasks such as navigation and planning. Table~\ref{tab:benchmark_role} provides a brief summary of each selected benchmark and capability, as well as a justification to why that benchmark is included in \collection. 

Ultimately, we consider the current list of benchmarks a contribution that needs to be maintained over time, with the assumption that the list needs to be completed (e.g. with math and planning benchmarks) and refreshed with new capabilities. Some benchmarks will also need to be deprecated as they get saturated with new releases.  

\noindent {\bfseries Models.} Given the challenging nature of the selected benchmarks, and the goal of assessing frontier developments in AI, we consider a broad range of six model families but we only pick the most advanced models within each family. Table \ref{tab:all_models} shows all models included in the evaluation by modality.

\begin{table}[t]
\footnotesize
\centering
\setlength{\tabcolsep}{3.5pt}
\begin{tabular}{lllllll}
\toprule
\textbf{Modality}                                                                      & \textbf{Claude}                                                           & \textbf{Gemini} & \textbf{GPT}                                                                                              & \textbf{Llama}                                                                       & \textbf{Llava} & \textbf{Mistral}   \\ \midrule
\textbf{\begin{tabular}[c]{@{}l@{}}Multimodal\\ Image $\rightarrow$ Text\end{tabular}} & \begin{tabular}[c]{@{}l@{}}\ClaudeOpus\\ \ClaudeSonnet\end{tabular} & \GeminiPro  & \begin{tabular}[c]{@{}l@{}}\GPTFourVisionPreview\\ \GPTFourTurboApril\\ \GPTFourO\end{tabular} &                                                                                      & \Llava  & \\ \hline
\textbf{\begin{tabular}[c]{@{}l@{}}Language\\ Text $\rightarrow$ Text\end{tabular}}    & \begin{tabular}[c]{@{}l@{}}\ClaudeOpus\\ \ClaudeSonnet\end{tabular} & \GeminiPro  & \begin{tabular}[c]{@{}l@{}}\GPTFourPrev\\ \GPTFourO\end{tabular}                            & \begin{tabular}[c]{@{}l@{}}\LlamaThree\\ \LlamaThreeOne\\ \LlamaThreeOneLarge\end{tabular} &                & \MistralLargeTwo 
\\ \bottomrule
\end{tabular}
\caption{Models being evaluated via \textsc{Eureka-Bench} for language and multimodal capabilities.}
\label{tab:all_models}
\end{table}

\noindent {\bfseries Methodology and analysis.} Rather than reporting overall single scores for each benchmark and model, we extract more granular and deeper insights that can better characterize failures of each model and also conduct meaningful comparisons. In particular, this report contributes three distinct types of analysis:
\vspace{-2mm}
\begin{itemize}[leftmargin=*]
    \item \textbf{Disaggregated reports of model performance across important experimental conditions and subcategories of data.} Previous work in model evaluation and error analysis \cite{EyubogluVSDLD0R22,singla2021understanding,nushi2018towards,barocas2021designing} has shown that single-score evaluations can hide important failure modes. Here, we build upon the prior work and disaggregate performance across input attributes (i.e. subcategories of data) and experimental conditions relevant for the given capability. While analysis for subcategories slices the benchmark based on the content and nature of input prompts, experimental conditions (when applicable) add another dimension that relates to the nature of the task itself or how the model is prompted to perform the task. Table~\ref{tab:all_benchmarks} shows all experimental conditions and subcategories studied per benchmark, and in-depth results are in sections~\ref{sec:multimodal} and~\ref{sec:language}. 
    \item \textbf{Analysis of model non-determinism across identical runs.} For applications that use foundation models as a basis to their services, determinism of output is important for assuring reliable output and providing consistent experiences to end users. Throughout our experiments we observe that this property is not guaranteed from most generative models, even when the prompt is identical and generation temperature is set to zero. In Section~\ref{sec:non_determinism}, we compare all models in this report with respect to this phenomenon and show that, for many of them, there are major variations at the example level for identical runs.
    \item \textbf{Backward compatibility analysis within model families for measuring progress and regress upon model updates.} Backward compatibility was first studied for predictive machine learning~\cite{bansal2019updates,srivastava2020empirical,trauble2021backward,matsuno2023robust,shen2020towards} for measuring whether model updates can cause regressions in performance for individual examples or whole groups. This work showed that regressions can happen during updates even when overall accuracy increases. More recent findings have shown that the phenomenon can also happen for generative models with brittleness cases when identical prompts that had worked in older versions of the model do not work in newest updates~\cite{echterhoff2024muscle,chen2023chatgpt,ma2024my}. Section~\ref{sec:backward_compatibility} shows that across three model families (GPT, Claude, and Llama) incompatibility can be observed at both the example level as well as per data subcategory.
\end{itemize}
\section{Results Summary}
\vspace{-2mm}
\begin{figure}
    \centering
    \includegraphics[width=0.46\linewidth]{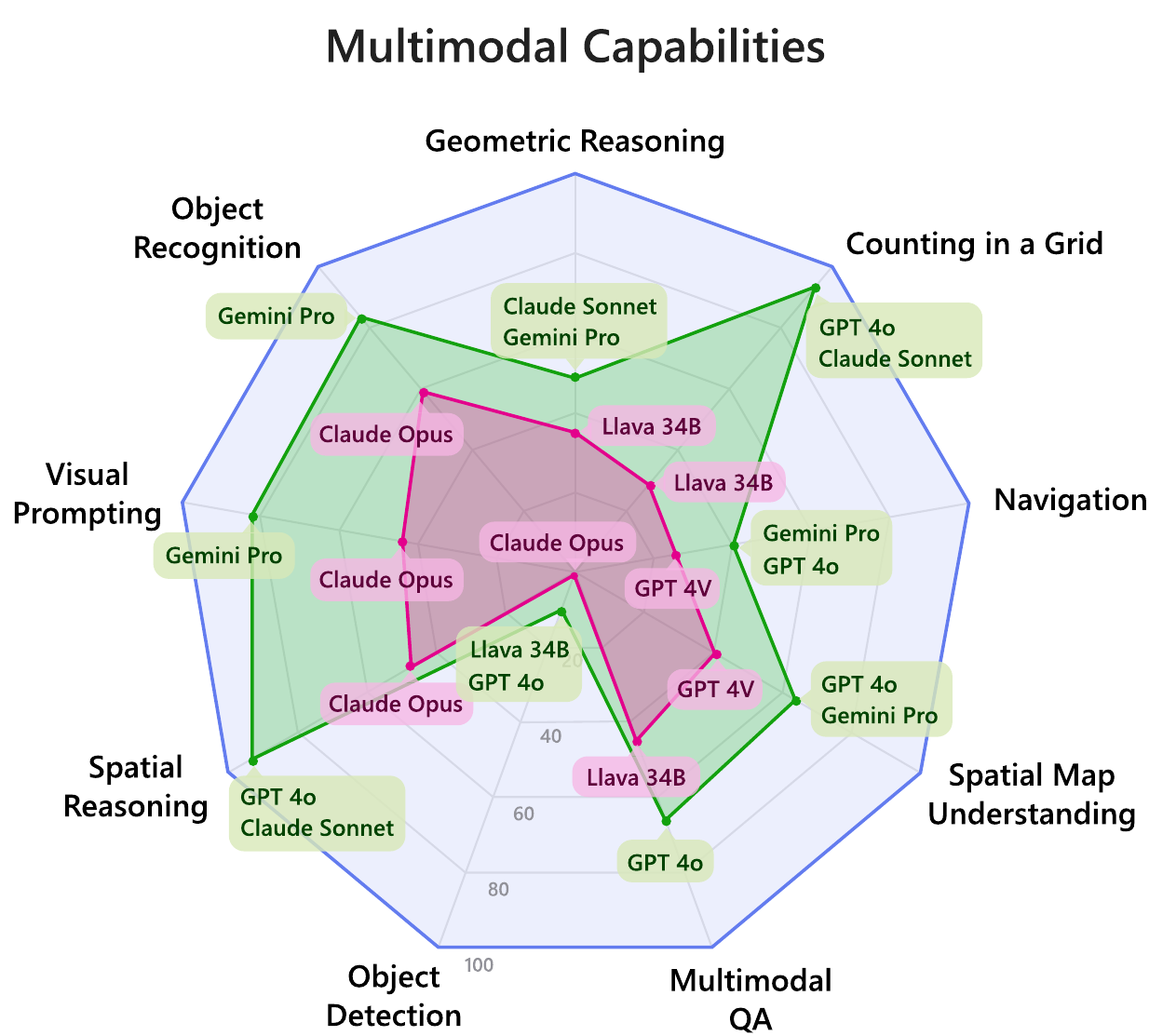}
    \includegraphics[width=0.52\linewidth]{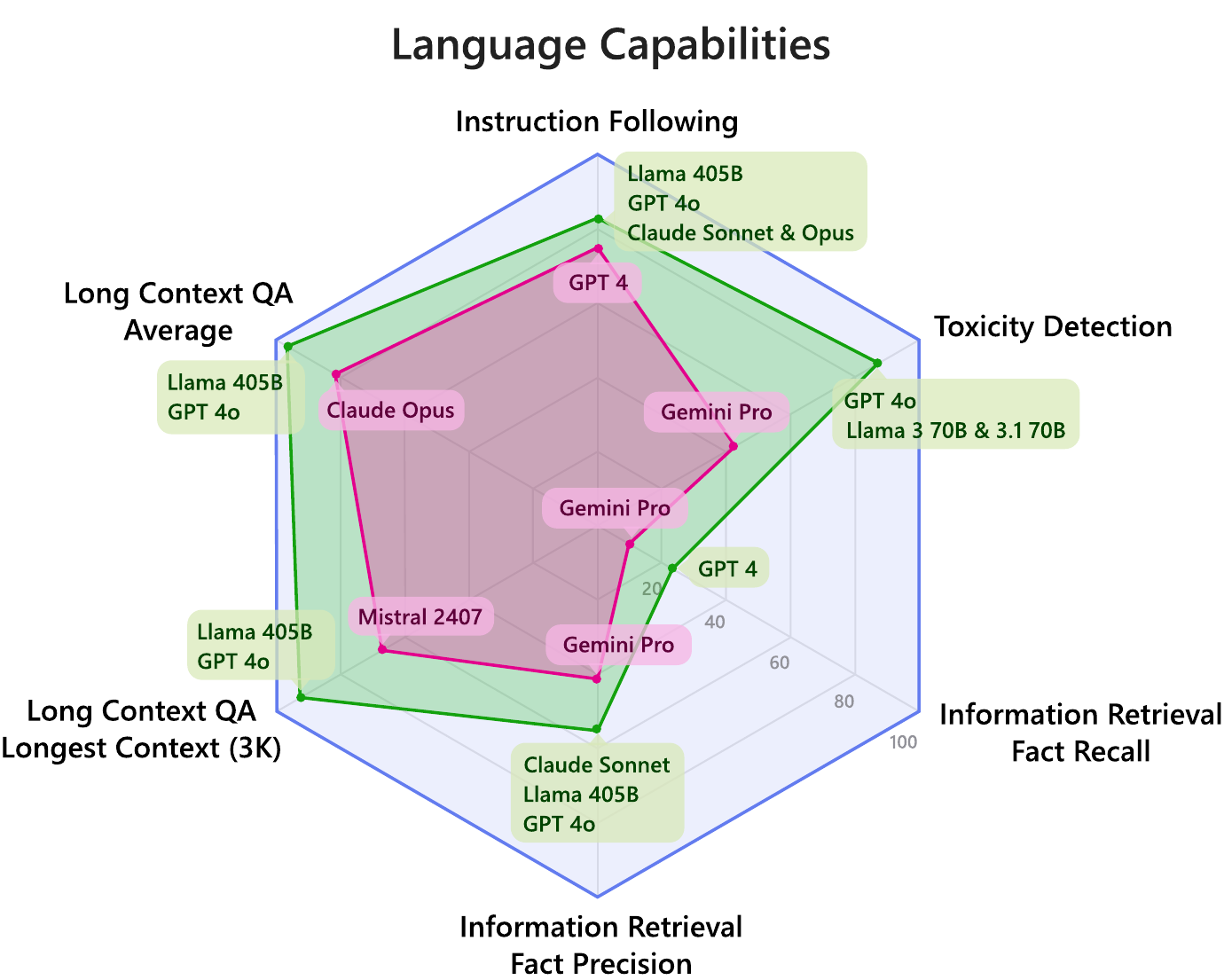}
    \caption{Performance of best and worse models for multimodal (left) and language (right) datasets in in \collection. The \textcolor{redfrontier}{red} frontier shows the performance of the worse model, indicating the area that is already solved for the set of capabilities. The \textcolor{greenfrontier}{green} frontier shows the performance of the best model, indicating the best known result with current technology. The \textcolor{bluefrontier}{blue} horizon between the best model and the maximum performance shows the room for improvement for mastering the capability. The best performance sets indicated in the green border include all models that perform within 2\% of the best observed result.}
    \label{fig:summary}
\end{figure}
Figure~\ref{fig:summary} is a high-level illustration of the state of the art in AI for \collection, showing the best and the worst performance per capability. These results show a complementary picture of capabilities of different models and that there is no single model that outperforms all others in most tasks. However, \ClaudeSonnet, \GPTFourO, and \LlamaThreeOneLarge repeatedly outperform others in several capabilities.

\noindent \textbf{Multimodal Evaluation:} Evaluations on important vision-language capabilities such as geometric and spatial reasoning, object recognition and detection, multimodal question answering, and navigation demonstrate increased capabilities of most recent models when compared to their previous versions. For example, \GPTFourO improvements over \GPTFourVisionPreview range between 3\%-20\%. Yet, state-of-the-art models are still fairly limited in their multimodal abilities, specifically when it comes to detailed image understanding (e.g. localization of objects, geometric and spatial reasoning, and navigation), which is most needed in truly multimodal scenarios that require physical awareness, visual grounding, and localization.   
\vspace{-1mm}
\begin{enumerate}[leftmargin=*]
    \item \textbf{State-of-the-art multimodal models struggle with geometric reasoning.}
    Reasoning about height is more difficult than about depth. \ClaudeSonnet and \GeminiPro are the best performing models for this task with \ClaudeSonnet being the most accurate model for depth ordering and \GeminiPro the most accurate for height ordering.
    \item \textbf{Multimodal capabilities lag language capabilities.} On tasks which can be described either as a multimodal task or as language-only, the performance of most tested models is higher for the language-only condition. \GPTFourO is the only model that consistently achieves better results when presented with both vision and language information, showing therefore that it can better fuse the two data modalities.
    \item \textbf{Complementary performance across models for fundamental multimodal skills.} For example, \ClaudeSonnet, \GPTFourO, and \GPTFourTurboApril have comparable performance in multimodal question answering (MMMU) but they outperform all other models by at least 15\%.
    There are tasks like object recognition and visual prompting where the performance of \ClaudeSonnet is better or comparable to \GPTFourO, but \GeminiPro outperforms them both. Finally, in tasks like object detection and spatial reasoning, \GPTFourO is the most accurate model.
\end{enumerate}

\noindent \textbf{Language Evaluation:} The evaluation through \collection shows that there have been important advances from state-of-the-art LFMs in the language capabilities of instruction following, long context question answering, information retrieval, and safety.

\begin{enumerate}[leftmargin=*]
    \item \textbf{Faster improvements in instruction following across all model families.} Amongst the studied language capabilities, instruction following is where most models are improving faster, potentially due to strong investments in instruction tuning processes, with most models now having an instruction following rate of higher than 75\%.
    \item \textbf{All models' performance in question answering drops with longer context.} When state-of-the-art models are compared in “needle-in-a-haystack” tasks, they seem to all perform equally well. However, testing the models on tasks that involve reasoning over long-context, reveals that all models’ performance drops as context size grows. Amongst all models, \GPTFourO and \LlamaThreeOneLarge have the lowest drop in performance for longer context.
    \item \textbf{Major gaps in factuality and grounding for information retrieval from parametric knowledge or input context.} For example, we observe query constraint satisfaction rates (i.e. fact precision) of lower than 55\%, completeness rates of lower than 25\% (i.e. fact recall), and information irrelevance rates of higher than 20\% (potentially information fabrication). \LlamaThreeOneLarge, \GPTFourO, and \ClaudeSonnet are the best performing models in this task across different conditions. \GPTFourO and \ClaudeSonnet in particular have significantly lower information irrelevance rates (associated with better factuality). \LlamaThreeOneLarge has better constraint satisfaction rates (associated with better constrained text generation and grounding).
    \item \textbf{High refusal rates. Lower accuracy in detecting toxic content vs. neutral content for most models.} While several models have high accuracy rates for toxicity detection, others (\GeminiPro, \ClaudeSonnet, \ClaudeOpus, and \LlamaThreeOneLarge) exhibit low accuracy in classifying toxic content and a high amount of refusal. During the safe language generation evaluation, models like \GPTFourPrev and \MistralLargeTwo have the highest toxicity rates. \GPTFourO is the only model that has both a high toxicity detection accuracy and a low toxicity score for safe language generation, as shown in the discriminative and generative evaluations respectively.
\end{enumerate}
\noindent \textbf{Several models have highly non-deterministic output for identical runs.} We study outcome determinism for all models by running three identical runs (temp=0, top\_p= 0.95), and then report different measures of non-determinism such as disagreement, variation, and entropy of outcomes. \GeminiPro, \GPTFourPrev, \GPTFourVisionPreview, and \GPTFourTurboApril show high non-determinism of outcomes. For example, there exists a 26\% and a 14\% disagreement across three runs of \GeminiPro on a random sample of MMMU and IFEval respectively. This can translate to 1\%-4\% fluctuations in performance at the subcategory level for these benchmarks. These results raise important questions regarding the stability of user and developer experiences when repeatedly inferencing with identical queries using the same prompt templates. \LlamaThree, \LlamaThreeOne, and \MistralLargeTwo are almost perfectly deterministic. The Claude family and \GPTFourO, and \LlamaThreeOneLarge are more deterministic than \GeminiPro and GPT-4 versions prior to \GPTFourO but yet non-zero and sometimes highly non-deterministic for longer generative tasks.

\noindent \textbf{Backward incompatibility for shifts within the same model family is prevalent across all state-of-the-art models.} This is reflected in high regression rates for individual examples and at a subcategory level. This type of regression can lead to breaking trust with users and application developers during model updates. Regression varies per task and metric, but we observe several cases when it is higher than 10\% across three model families (Claude, GPT, Llama), and sometimes they can dominate progress rates for whole subcategories of data.

The complementary nature of these results shows that there is space and demonstrated opportunity to improve current models in different areas, at least to the level of the best performing model for each individual capability in this challenge set. Despite this, several tasks in the challenge set we evaluate on remain difficult even for the most capable models and it is important to discuss and explore whether these gaps can be addressed with current technologies, architectures, and data synthesis protocols in place.

\begin{table}[htbp]
\scriptsize
\centering
\begin{tabular}{llp{8cm}}
\toprule
\textbf{\begin{tabular}[c]{@{}l@{}}Benchmark\\ Modality\end{tabular}}                          & \textbf{Capability}                                                                                                  & \textbf{Role in EUREKA-BENCH}                                                                                                                                                                                                                                                                                                                                              \\ \midrule
\textbf{\begin{tabular}[c]{@{}l@{}}GeoMeter\\ Image → Text \\ Section~\ref{sec:geometric_reasoning}\end{tabular}}             & Geometric Reasoning                                                                                                  & \begin{tabular}[c]{@{}p{8cm}}{\bf Task}: Predicting depth and height orderings.\\ {\bf Capability importance}: Depth and height understanding are a fundamental cognitive capability \\ for natural language interaction with physical environments and for navigation. \\ {\bf State-of-the-art}: Even most capable models have an accuracy of less than 50\% in the task.\end{tabular} \\ \hline
\textbf{\begin{tabular}[c]{@{}l@{}}MMMU\\ Image → Text \\ Section~\ref{sec:mmmu}\end{tabular}}                            & Multimodal QA                                                                                                        &                                                                          \begin{tabular}[c]{@{}p{8cm}}
{\bf Task}: Question answering for image content across different disciplines.\\ 
{\bf Capability importance}: Reasoning over image content is important for assessing multimodal knowledge and for measuring multimodal understanding skills. 
\\ {\bf State-of-the-art}: Even most capable models have an accuracy of less than 70\% in the task.
\end{tabular}                                                                                                                                                                                                                                                                                                  \\ \hline
\textbf{\begin{tabular}[c]{@{}l@{}}Image Understanding\\ Image → Text\\ Section~\ref{sec:image_understanding}\end{tabular}}             & \begin{tabular}[c]{@{}l@{}}Object Recognition\\ Object Detection\\ Visual Prompting\\ Spatial Reasoning\end{tabular} &    
\begin{tabular}[c]{@{}p{8cm}}
{\bf Task}: Recognizing and detecting objects in an image or a given section of the image, reasoning about the spatial relationships between them.\\ 
{\bf Capability importance}: Most tasks in this benchmark are representative of classic vision tasks for which traditional ML and vision methods have matured, but generative models have not yet caught up. All tasks are important for  understanding and localizing image content, relevant to almost all multimodal applications. 
\\ {\bf State-of-the-art}: There is a large variance across models and capabilities in this benchmark. The goal of the evaluation in this benchmark is to characterize this variance. While there are a few cases and models where performance is higher than 80\%, for most models this is not the case.
\end{tabular}                                                                                                                                                                                                                                                                                                                                                                                                            \\ \hline
\textbf{\begin{tabular}[c]{@{}l@{}}Vision Language\\ Understanding\\ Image → Text\\ Section~\ref{sec:vl_understanding}\end{tabular}} & \begin{tabular}[c]{@{}l@{}}Spatial Understanding\\ Navigation\\ Counting\end{tabular}                                &                 \begin{tabular}[c]{@{}p{8cm}}
{\bf Task}: A set of synthetic tasks whose input can be phrased either in language, image, or both. \\ 
{\bf Capability importance}: The goal here is to disentangle the role and model performance for each modality, for models that support both vision and language as input. Spatial Understanding and Navigation are chosen as reasoning capabilities. Counting is chosen as a basic capability for which several models still struggle with, in at least one of the modalities. 
\\ {\bf State-of-the-art}: There is a large variance across models, modalities, and capabilities in this benchmark, with most models performing at less than 70\% spatial understanding and less than 50\% in navigation. The goal of the evaluation in this benchmark is to characterize this variance.
\end{tabular}                                                                                                                                                                                                                                              \\ \hline
\textbf{\begin{tabular}[c]{@{}l@{}}IFEval\\ Text → Text \\ Section~\ref{sec:ifeval}\end{tabular}}                           & Instruction Following                                                                                                &                                    \begin{tabular}[c]{@{}p{8cm}}
{\bf Task}: Instruction following for formatting, styling, and organizing generated text.\\ 
{\bf Capability importance}: Horizontal language task that impacts several generative writing scenarios that require generation with specific user constraints. 
\\ {\bf State-of-the-art}: Overall accuracy for most capable models ranges between 70\% and 85\%. However, there are several instruction subcategories and models for which performance is lower and for which there is higher variance.
\end{tabular}                                                                                                                                                         \\ \hline
\textbf{\begin{tabular}[c]{@{}l@{}}FlenQA\\ Text → Text\\ Section~\ref{sec:flenqa}\end{tabular}}                           & Long Context QA                                                                                                      &                                    \begin{tabular}[c]{@{}p{8cm}}
{\bf Task}: Answering logical reasoning questions on long context.\\ 
{\bf Capability importance}: Reasoning upon several statements distributed across long context is relevant for long document understanding that goes beyond merely retrieving simple facts from context (i.e. needle-in-the-haystack tasks). 
\\ {\bf State-of-the-art}: While most models perform well in this benchmark for short input context, as context length increases, many models are not able to maintain good accuracy with several of them having an accuracy of less than 80\%.
\end{tabular}                                                                              \\ \hline
\textbf{\begin{tabular}[c]{@{}l@{}}Kitab\\ Text → Text\\ Section~\ref{sec:kitab}\end{tabular}}                            & Information Retrieval                                                                                                &                                    \begin{tabular}[c]{@{}p{8cm}}
{\bf Task}: Retrieving long-form information from the model's parametric knowledge or from given input context with filtering constraints.\\ 
{\bf Capability importance}: All information retrieval tasks involve some form of constraint that defines the retrieval query. However, other simpler IR benchmarks only test for short-form generation (finding a single fact) and for a single constraint. Being able to answer more complex queries is relevant to the factuality and grounding of advanced search and information finding.
\\ {\bf State-of-the-art}: Constrained retrieval from parametric knowledge is still prone to major irrelevance and fact fabrication with constraint satisfaction being less than 60\%. Constrained retrieval from given input context is significantly better in overall, but for queries with more than one constraint constraint satisfaction and completeness drop to less than 70\% and 60\% respectively.
\end{tabular}                                                                                 \\ \hline
\textbf{\begin{tabular}[c]{@{}l@{}}Toxigen\\ Text → Text\\ Section~\ref{sec:toxigen}\end{tabular}}                          & \begin{tabular}[c]{@{}l@{}}Toxicity Detection\\ Safe Language Generation\end{tabular}                                &                            \begin{tabular}[c]{@{}p{8cm}}
{\bf Task}: Classifying whether a given text is toxic or neutral (discriminative case). Prompting the model with questions/statements that could lead to unsafe language generation, and testing the safety of the generated language (generative case).\\ 
{\bf Capability importance}: Ability to distinguish between toxic and neutral language is relevant for dialogue comprehension and also for establishing the model utility on content moderation scenarios. Safe language generation itself is an important capability that impacts different aspects of responsible AI including representational fairness, inclusion, and safety.
\\ {\bf State-of-the-art}: State-of-the art models face two types of challenges in this benchmark. First, there are discrepancies between performance of models across different demographic groups. Second, some of the models exhibit a high refusal rate for the toxicity detection task, which makes these models not useful for content moderation and other scenarios where toxicity detection is key.
\end{tabular}                                                                                                                                                                             \\ \bottomrule
\end{tabular}
\caption{The role of each benchmark in \textsc{Eureka-Bench} and the importance of each capability for measuring progress in AI.}
\label{tab:benchmark_role}
\end{table}

\section{Evaluation Framework}
\label{sec:framework}
In the fast-paced space of AI research, where new models and benchmarks are introduced and others are deprecated frequently, it is important for evaluation and understanding efforts to be able to reuse existing evaluation pipelines with minimal adjustments to efficiently accommodate new models and benchmarks. This calls for a modular design that allows the users to onboard a new benchmark or model by inheritance of pipeline definitions from existing experiments and implementing changes only where overriding the existing pipeline is necessary.

All experiments reported in this paper were conducted through \framework, a software framework that unifies all of the benchmarks and models used in this report. This framework was designed to ensure reproducibility, composability, and reusability. We have opened the source code of \framework to maximize transparency into the details of all benchmarks and evaluation settings. 

The \framework framework currently supports both language and multimodal (text and image) data, and allows the user to define custom pipelines for data processing, inference, and evaluation, with the possibility to inherit from existing pipelines to minimize development work. We have released the experiment pipelines for all benchmarks listed in Table \ref{tab:all_benchmarks}.

Each experiment under the \framework framework is defined using a \verb|Pipeline|. These pipelines are comprised of a set of \verb|Component|s. This modular design not only improves readability, but also allows extendability and reusability as users can inherit from an existing \verb|Pipeline| and only implement minimal changes.

Currently, \framework operates with the following set of \verb|Component|s: 
\begin{itemize}[leftmargin=*]
    \item \verb|PromptProcessing|: This component is used to prepare data for inference, apply data manipulations, or apply complex prompt templates. It enables flexible and reproducible experimentation with different prompt templates and data pre-processing steps.
    \item \verb|Inference|: This component performs model inference on any processed data. For example, inference can be done for the model that is subject to evaluation, or a model that is involved in the evaluation pipeline as an evaluator of the original model's output.
    \item \verb|DataProcessing|: This component is used to post-process the model outputs to extract the model response from the generated text. This is necessary for example when a regex search of the option selected is needed in multiple-choice scenarios or when the generated text includes tags that were used in training (e.g. \texttt{|assistant|}) and need to be removed before metric calculation.
    \item \verb|EvalReporting|: This component facilitates evaluation of the processed model outputs using various metrics and an arbitrary number of aggregations, and logs the metric results for individual prompts as well as aggregation results.
    \item \verb|DataJoin|: This component is used to join two sources of data, for example to join the model outputs with the ground truth data for evaluation.
\end{itemize}
\begin{figure}[t]
    \centering
    \includegraphics[width=\linewidth]{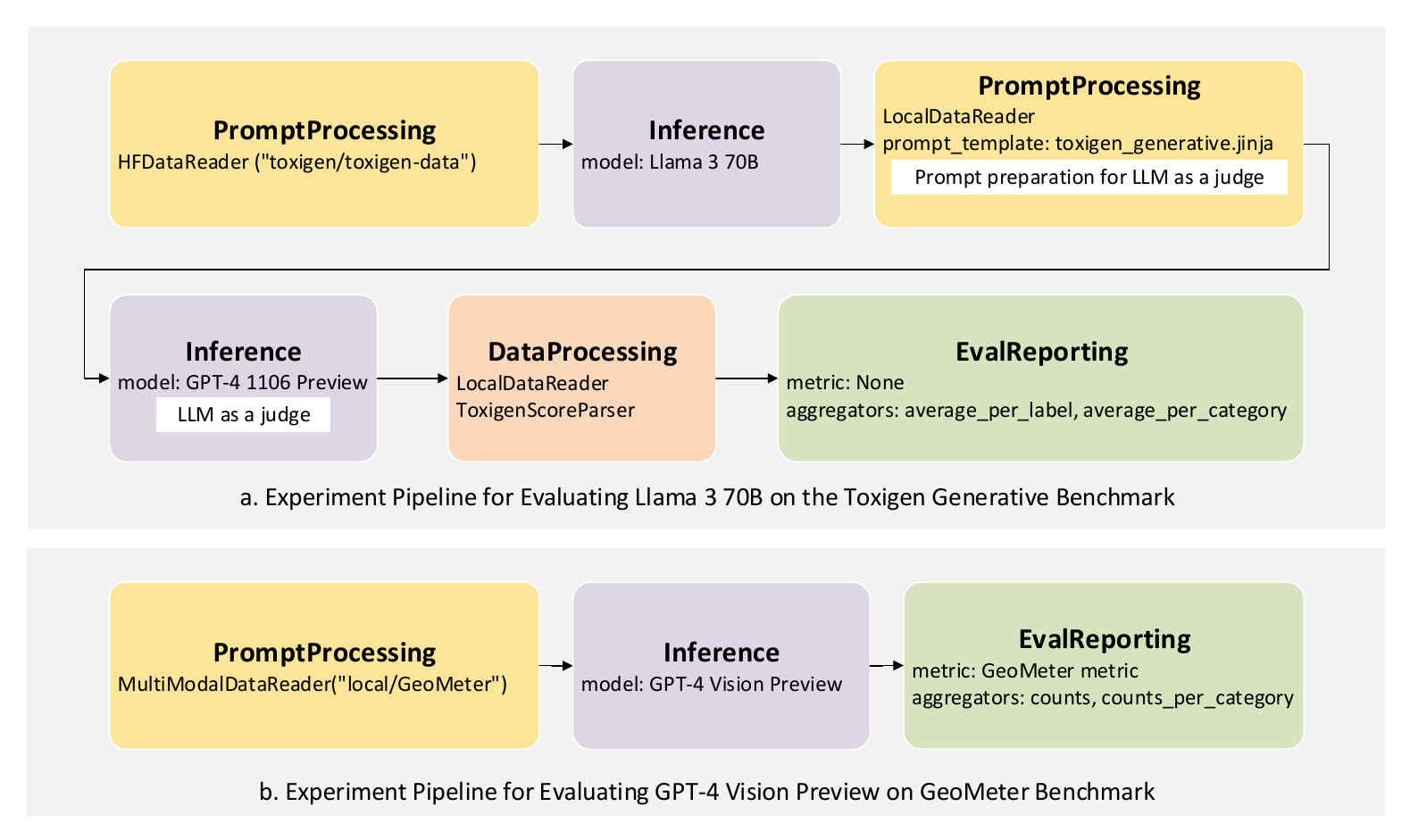}
    \caption{Overview of experiment pipelines for two example evaluation experiments: Toxigen Generative (a) and GeoMeter (b). Components are configurable at instantiation time to maximize code reuse and enable controlled experimentation. For example, the PromptProcessing component is shown here to use different data readers or prompt templates in different contexts.
    }
    \label{fig:framework}
\end{figure}

Importantly, all \texttt{Components} log their outputs in standardized \texttt{jsonl} files, which increases transparency into each step of the evaluation, facilitates error analysis, and streamlines result analysis and visualization across multiple experiments.

Furthermore, the \texttt{Component}s make use of utility classes, including but not limited to \texttt{DataLoader}s, \texttt{Model}s, \texttt{Metric}s, and \texttt{Aggregator}s. Each utility class is configurable using its corresponding \texttt{Config} class. All \texttt{Config}s and individual parameters defining an experiment \verb|Pipeline| are logged in the experiment directory as parts of the \verb|Pipeline| config to ensure reproducibility. \framework also provides implementations of \texttt{Model}s for inference either through APIs or local deployments. 

For an overview of two example experiment pipelines available in \framework, see Figure \ref{fig:framework}. The top part of the figure shows the evaluation pipeline for the Toxigen Generative benchmark. First, the PromptProcessing component reads the Toxigen data from HuggingFace and prepares the prompts for inference. Next, the Inference component is used to inference the model under evaluation, in this example \LlamaThree. After that, the PromptProcessing component is reused, this time to load the inference results and prepare prompts for the judge model using a prompt template. The Inference component is then reused to inference the judge model and score the original inference results. Finally, the DataProcessing component is used to extract scores from the judge model inference results. The scores are aggregated in several ways in the EvalReporting component. 

The bottom part of the figure shows the components comprising the GeoMeter experiment pipeline. This pipeline is different from Toxigen in two important regards: it deals with multimodal data and it does not use another model as a judge, instead it offers a metric class to score the inference results. Despite these differences, the same components can be reused with minimal adjustments to accomodate this scenario. Namely, the PromptProcessing component is set to read multimodal data from a local directory, and the EvalReporting component is set to use the GeoMeter metric class. As shown in both pipelines, the Inference component can be easily configured to use the desired model, enabling fair comparison between models without changing the rest of the pipeline, and with maximum code reuse.
\section{Multimodal Evaluation}
In this section, we provide detailed analysis and results for the capabilities of geometric reasoning (GeoMeter), multimodal question answering (MMMU), object recognition, object detection, visual prompting, spatial understanding and reasoning, navigation, and counting. 

To account for the impact of non-determinism 
 (discussed in Section~\ref{sec:non_determinism}), all experiments reported here were repeated three times and we report the mean and corresponding standard error across the three repeated runs with temperature set to zero and top\_p = 0.95.

\label{sec:multimodal}
\subsection{Geometric Reasoning - GeoMeter}
\label{sec:geometric_reasoning}
\noindent \textbf{Motivation:} The ability to understand visual properties such as size, shape, depth, and height is fundamental to visual understanding, yet many existing Visual Question Answering (VQA) benchmarks \cite{johnson2017clevr, chen2024spatialvlm, liu2023visual, diwan2022winoground, thrush2022winoground} do not specifically focus on the depth and height perception capabilities of Vision Language Models (VLMs). Accurate perception of these dimensions is vital for practical applications like scene understanding, navigation, monitoring, and assistive technologies. The lack of accurate depth and height understanding in VLMs can lead to serious consequences, such as misjudging the proximity of objects, which could result in catastrophic outcomes in real-world scenarios. 

Despite VLMs' abilities to recognize object shapes and sizes, their depth and height reasoning often relies on learned size/shape cues rather than actual geometric analysis, potentially influenced by biases from training data \cite{jayaraman2024d}. Alternatively, models might estimate the depth based on the apparent size of objects, without genuine inter-object reasoning. Additionally, when faced with multiple choices, VLMs might also show bias towards certain answers, influenced by the prevalence of similar data during training. Thus, it becomes important working with focused benchmarks that enhance understanding of true depth and height perception in VLMs, ensuring they perform reliably in complex, real-world environments.

Here, we use \textbf{GeoMeter}, a geometric reasoning benchmark derived from previous work \cite{azad2024dhbench}, which is specifically designed to evaluate the depth and height reasoning capabilities of Vision Language Models (VLMs).
GeoMeter comprises approximately \textit{1086 unique image-text pairs} across two tasks: depth and height. 
The data consists of synthetic examples depicted by 2D shapes like triangles, squares, rectangles, circles etc.
The development of synthetic datasets featuring basic shapes aims to genuinely test the visual reasoning capabilities of models, focusing on their ability to process visual information without relying on familiar real-world cues and biases. Our motivation comes from concerns about test time data leakage that could arise when models, trained on vast existing datasets, encounter images during testing that they may have already seen during training. By using unique datasets, we seek to ensure a more accurate evaluation of the model's true visual reasoning abilities.

\begin{figure}[t]
    \centering
    \includegraphics[width=\linewidth]{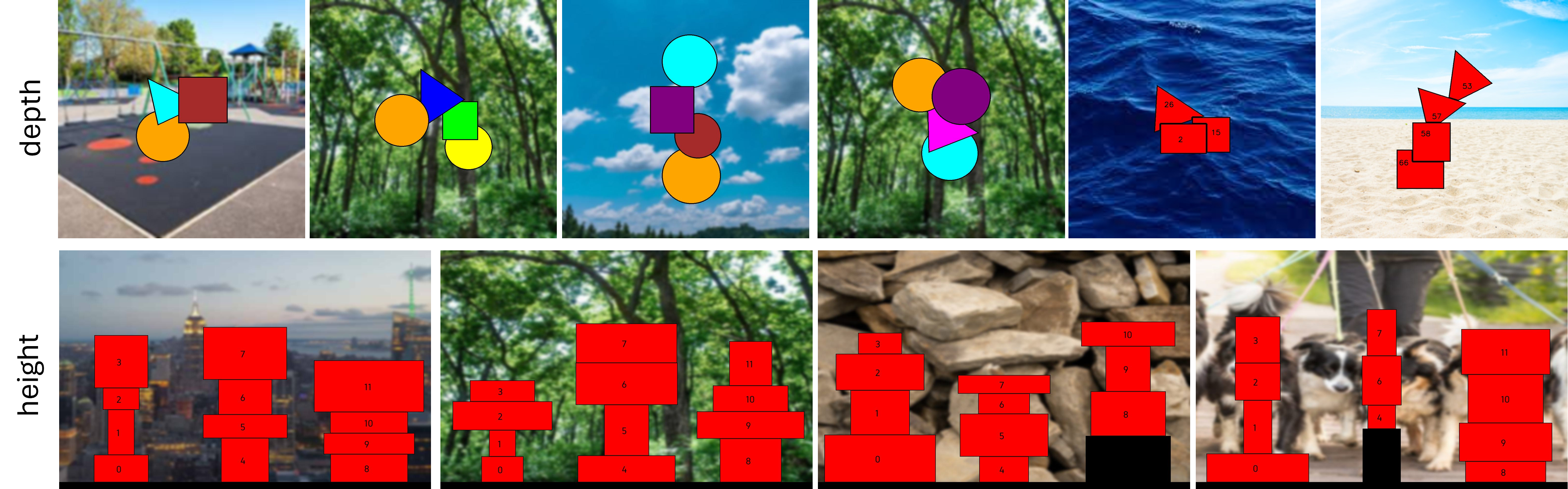} 
    \caption{Samples from the GeoMeter dataset. Here, each sample is shown with random query attributes including color - numeric label and color - shape label.}
    \label{fig:geobench_samples}
\end{figure}

\begin{figure}[t]
    \centering
    \includegraphics[width=\linewidth]{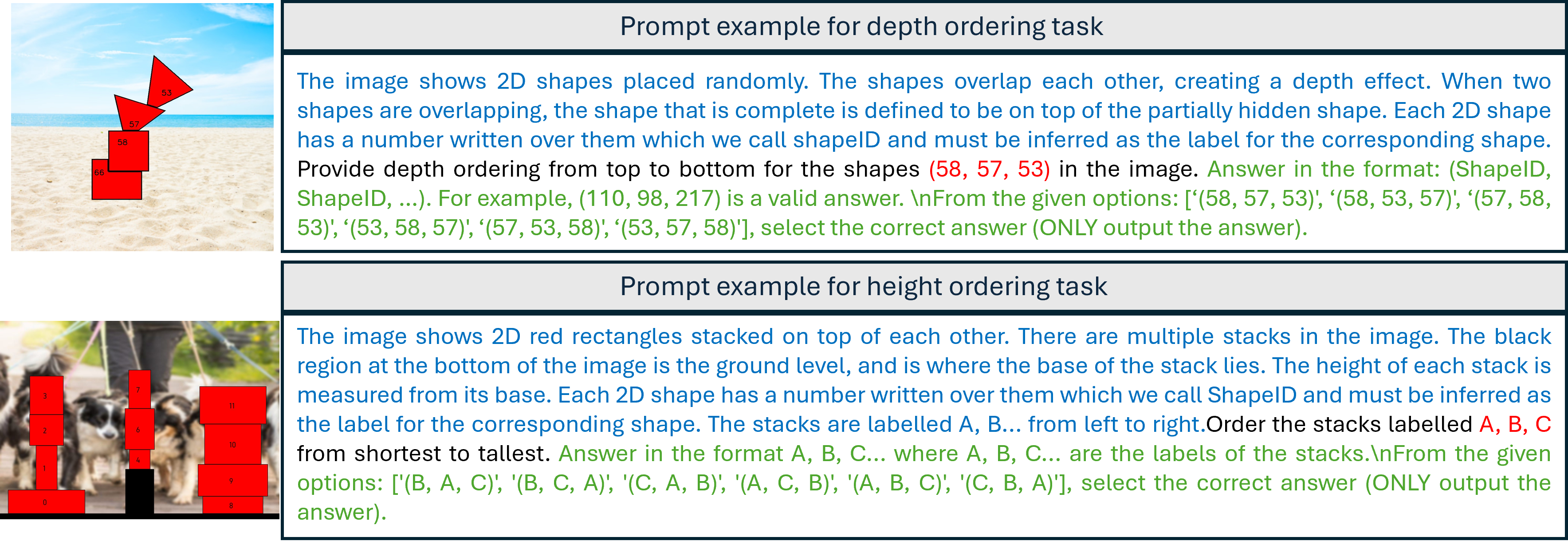}
    \caption{Sample image-text pair from the GeoMeter dataset. Here the image contains 5 shapes labeled with random numeric labels which are used as query attributes in the prompt. Prompt template shows the basic template for each image-text pair of all our benchmark, where the prompt example is the actual prompt for this image. The prompt example is appended with either MCQ or True/False type question.}
    \label{fig:geobench_prompt}
\end{figure}

\noindent \textbf{Benchmark Description:} GeoMeter consists of synthetic 2D images to test model performance on depth and height perception tasks. Table \ref{tab:geobench_stat}, Figure \ref{fig:geobench_samples} and Figure \ref{fig:geobench_prompt} respectively show the dataset statistics, sample images and sample image-text pair of our proposed datasets. The dataset generation can be divided into two parts - Image generation and Question generation.

\begin{table}[t]
    \centering
    \resizebox{\linewidth}{!}{
    \begin{tabular}{cccccp{3.3cm}c}
        \toprule
         \textbf{Dataset} & \textbf{Subcategory} & \textbf{Task}& \textbf{Question Type}& \textbf{Questions}& \textbf{Query attributes}& \textbf{Img-Text pairs}\\
         \midrule
         \multirow{4}{*}{GeoMeter} & \multirow{2}{*}{Depth} & \multirow{4}{*}{VQA}& \multirow{4}{*}{MCQ}& \multirow{2}{*}{986}& Color, Numeric label (random and patterned)&\multirow{4}{*}{1086}\\
         & \multirow{2}{*}{Height} & & & \multirow{2}{*}{100}& Color, Numeric label (random)\\
         \bottomrule
    \end{tabular}}
    \caption{Dataset statistics of GeoMeter. Here, query attributes are unique identifiers for the object of interest. MCQ denotes Multiple Choice Questions.}
    \label{tab:geobench_stat}
\end{table}

\noindent\emph{Image Generation.} The synthetic dataset is divided into two subcategories - \textit{depth} and \textit{height}, with each image containing a real-world image as a background to enhance realism. The dataset consists of \textit{1086 image-questions} pairs. The \textit{depth} category consists of \textit{986 images}, featuring rectangles, triangles, or circles that partially overlap to create a depth illusion, with unique identifiers such as colors, and numeric labels.
The \textit{height} category has \textit{100} images, where each tower consists of four rectangles with random dimensions. Further, in these images, towers are placed on a horizontal black strip that is treated as a raised platform. This category includes two sets: one with all towers placed at the same height level and another with a randomly chosen tower on a raised platform, with unique identifiers being label.  All towers are labeled sequentially.

\noindent\emph{Question Generation:} The following method for generating questions is applied consistently across all images. Each question is composed of two key components:  \textit{Description Prompt} and an \textit{Answer Format Instruction}. The \textit{Description Prompt} provides general information about the scene, offering semantic cues related to the image. This is then followed by the actual question and the \textit{Answer Format Instruction}. For instance, the \textit{Description Prompt} might be: \textit{``[additional information] Provide depth/height ordering for the shapes $<$question items$>$ in the image. [additional information]"}. This is then followed by an \textit{Answer Format Instruction} such as: \textit{``From the given options: $<$answer set$>$, select the correct answer [additional information]."} Each question is constructed by sequentially combining the \textit{Description Prompt} and \textit{Answer Format Instruction}. 

The \textit{question items} is a list containing \textit{$<$query attribute$>$} appended by \textit{$<$shape$>$}. Here \textit{$<$query attribute$>$} is one of the unique identifiers of the dataset. 
For example in the question item \textit{``red circle"}, \textit{``red"} is the \textit{$<$query attribute$>$} and \textit{$<$circle$>$} is the shape.  The \textit{answer set} contains all possible valid values (\textit{$<$query attribute$>$} + \textit{$<$shape$>$}) to that given prompt. To generate both the question items and answer set, we read through the scene graph and run depth-first search on it to generate valid unambiguous values of object-pair relationship. For each image, there are several multiple choice questions. For MCQ, the order of the given options is randomly generated, and ground truth is randomly placed in one of those options. Additionally, the answer for each question has been manually checked by the dataset creators.

\begin{table}[t]
    \centering
\begin{tabular}{lrrr}
         \toprule
          \bf{Model} & \bf{Depth ordering} (\%) & \bf{Height ordering} (\%) & \bf{Overall (\%)} \\          
          \midrule           
       \ClaudeOpus   &   42.42   $\pm$ 0.03 & 17.00 $\pm$ 0.00  & 40.10  $\pm$ 0.00 \\
       \ClaudeSonnet   &   \textbf{50.70}    $\pm$ 0.00   & 28.66 $\pm$ 0.33 & \textbf{48.66} $\pm$ 0.03 \\
       \hline       
       \GeminiPro    &  47.59 $\pm$ 0.14 & \textbf{32.00} $\pm$ 1.00 &  46.16 $\pm$ 0.18 \\
       \hline
       \GPTFourTurboApril &   38.84  $\pm$ 0.27 & 13.00 $\pm$ 1.00 & 36.46 $\pm$ 0.32 \\       
    \GPTFourVisionPreview     &   40.12 $\pm$ 0.44 & 13.33 $\pm$ 0.33 & 37.66 $\pm$ 0.41 \\
       \GPTFourO &    43.91  $\pm$ 0.00  & 19.33 $\pm$ 0.67 & 41.63 $\pm$ 0.07 \\
       \hline
       \Llava &   37.01  $\pm$ 0.10 & 15.00 $\pm$ 0.00 & 35.00  $\pm$ 0.10 \\
       \bottomrule
\end{tabular}
\caption{Performance comparison of the studied models on proposed benchmark. The reported results are average accuracy and standard error on three runs across depth and height subcategories. Top scores are in bold.}
\label{tab:geobench_avg}
\end{table}

\noindent\textbf{Aggregate and subcategory results:} We evaluate our benchmark on the task of visual question answering (VQA), with accuracy being the performance metric on MCQ type questions.
Evaluation is done across query attributes and number of shapes on probing the VLMs' depth and height perception. The performance of the selected models on the VQA task for MCQ type questions on the proposed benchmarks is shown in Table \ref{tab:geobench_avg}, where each row corresponds to the average accuracy across all different query attributes and shapes. Depth and height subcategory results are also presented in Table \ref{tab:geobench_avg}.

Main observations are as follows. First, \ClaudeSonnet achieves the best performance compared to the other models: \GPTFourO, \GPTFourTurboApril, \GeminiPro. Additionally, \ClaudeSonnet is also more consistent across multiple runs using the same prompts, as shown by the low standard error across three different runs.

\subsubsection*{Analysis and Discussion}
\noindent \textbf{Models generally struggle in depth and height perception tasks.} Results from Table \ref{tab:geobench_avg} highlight that the foundation multi-modal models struggle significantly with depth and height perception tasks involving similar shapes. This discrepancy underscores our benchmark's value in identifying gaps in VLMs' capabilities to handle more complex geometric reasoning, beyond mere shape recognition. 
To further support our claim that the low performance of models on the GeoMeter data is due to VLMs' deficiencies in depth and height reasoning, we also provide reference to relevant observations from prior work \cite{azad2024dhbench}. First, they note that VLMs generally perform well on basic geometric tasks such as line understanding, shape recognition, and shape counting, but fail on advanced perception tasks like depth and height perception tasks. Further, they also observe that VLMs struggle with depth and height reasoning tasks on real-world data as well.

\noindent \textbf{Models generally struggle more in height perception than depth.} Table \ref{tab:geobench_avg} indicates that both open and closed models struggle more with height perception compared to depth questions, with depth perception generally aided by occlusion cues and height perception challenged by the complexity of assessing vertically stacked objects. This difference indicates that models may find it somewhat easier to interpret scenarios with partial obstructions than to precisely evaluate complex vertical arrangements.

Overall, while models perform better in depth perception, they still show limitations in comprehensively handling more complex geometric understanding tasks, underscoring an area for improvement in advanced geometric task perception. 

\subsubsection*{Main takeaways}
\noindent\fbox{%
    \parbox{\textwidth}{%
        \begin{itemize}[leftmargin=*]
        \item State-of-the-art multimodal models struggle in depth and height perception tasks.
        \item Generally models show better depth perception than height.
        \item \ClaudeSonnet and \GeminiPro are the best performing models for this task with \ClaudeSonnet being the most accurate model for depth ordering and \GeminiPro the most accurate for height ordering.
        \end{itemize}
    }%
}
\begin{figure}[t]
    \centering
    \includegraphics[width=0.98\linewidth]{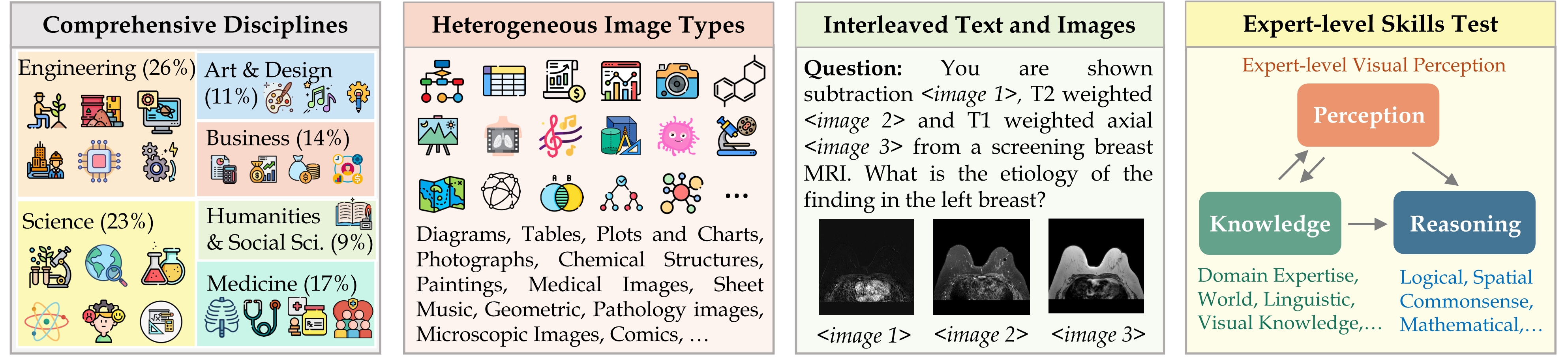}
    \caption{The MMMU dataset is a set of visual question-answering task that is comprehensiveness across 11.5K college-level problems across six broad disciplines and 30 subject-areas, and it requires detailed image understanding and reasoning requiring in deep subject knowledge.}   
    \label{fig:mmmu_overview}
\end{figure}

\subsection{Multimodal Question Answering - MMMU}
\label{sec:mmmu}
\noindent \textbf{Motivation:} A key use case for multimodal models is to serve as an expert assistant to answer queries and provide information and context about images.  This Visual Question Answering setup is one of the core tasks for multimodal models. It combines the abilities of understanding images at a high and detailed level with the ability of reasoning using that understanding.  MMMU~\cite{yue2024mmmu} is a popular dataset that tests these capabilities across a broad range of topics requiring deep image understanding and domain-specific knowledge.  We have included it in our evaluations due to this broad and deep coverage, wide adoption in the research and industry communities, and that it remains a challenging dataset that no models have yet mastered. Thus it provides a good high-level measure of multimodal reasoning performance.

\noindent \textbf{Benchmark Description:} MMMU tests multimodal multi-discipline reasoning in six core disciplines: Art and Design, Business, Science, Health and Medicine, Humanities and Social Science, and Tech and Engineering. The questions span 30 subject areas and 183 subfields, with a wide-variety of image types, such as charts, diagrams, maps, tables, music sheets, and chemical structures.  Questions are both multiple-choice and open-ended.  An illustration of the disciplines, subjects, images, and questions appears in Figure \ref{fig:mmmu_overview}.  For our evaluations, we use the 900-question validation set that spans all subject-areas.  

\noindent \textbf{Aggregate Baseline Results:} As a baseline evaluation we use the prompt formatting that is provided by the MMMU evaluation codebase~\cite{MMMUEvals}, which concatenates each question with the appropriate answer choices:
\begin{itemize}[leftmargin=*]
    \item Multiple-choice prompt example: ``A recent study found that the demand and supply schedules for Frisbees are as follows:\textless~image 1\textgreater Frisbee manufacturers persuade the government that Frisbee production improves scientists' understanding of aerodynamics and thus is important for national security. A concerned Congress votes to impose a price floor \$2 above the equilibrium price. What is the new market price? (A) 8 (B) 9 (C) 10 (D) 11."
    \item Open-ended prompt example: ``The graph below shows the supply and demand curves and the world price for bagels. \textless~image 1\textgreater What is the equilibrium price if this country does not trade?" 
\end{itemize}
The instruction prompts are as follows:
\begin{itemize}[leftmargin=*]
    \item Multiple-choice: ``\{prompt\} Answer with the option's letter from the given choices directly."
    \item Open-ended ``\{prompt\} Answer the question using a single word or phrase." 
\end{itemize}
This combined prompt is fed to each model as a user-prompt.

5\% of these questions require reasoning over more than one image.  This is supported for all models except \Llava, and thus for that model these questions are marked as unanswerable and thus incorrect.  

\begin{figure}[t]
    \centering
    \includegraphics[width=0.6\linewidth]{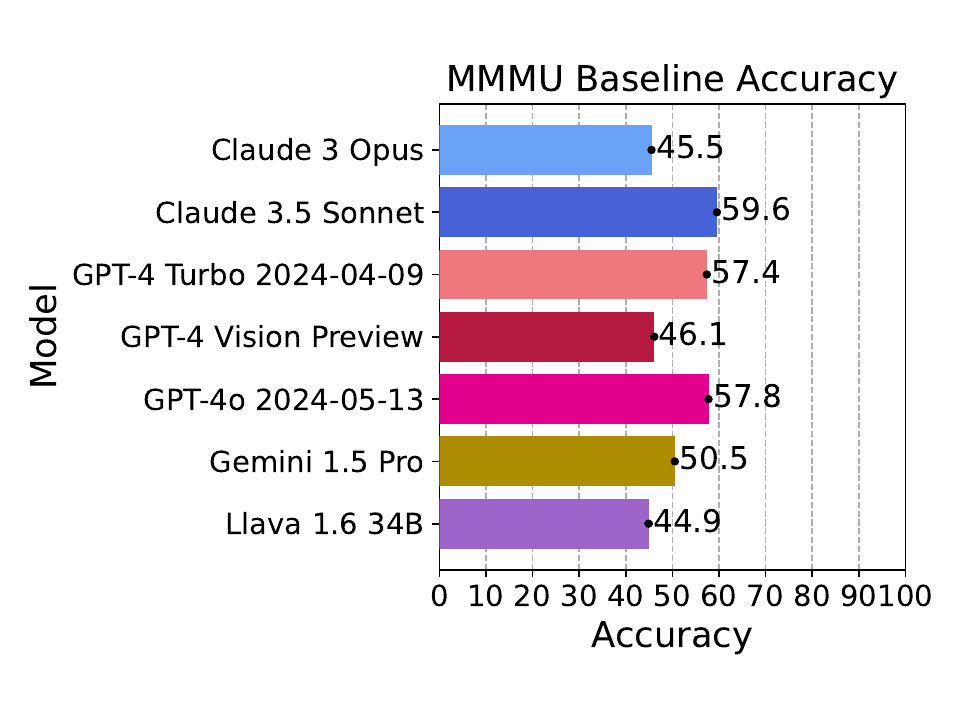}\vspace{-25pt}
    \caption{Aggregate accuracy reported across three different runs per model.}
    \label{fig:mmmu_baseline_accuracy}
\end{figure}

Figure~\ref{fig:mmmu_baseline_accuracy} presents the average accuracy for the entire validation dataset across the seven multimodal models evaluated in this report.  \ClaudeSonnet is the best performing model out-performing \GPTFourO by 2.2\%, followed closely by \GPTFourTurboApril and then \GeminiPro.  \Llava is the worst performing model overall, but as an open source model fairs well and is close in performance to \ClaudeOpus.

\begin{figure}[t!]
    \centering
    \includegraphics[width=0.7\linewidth]{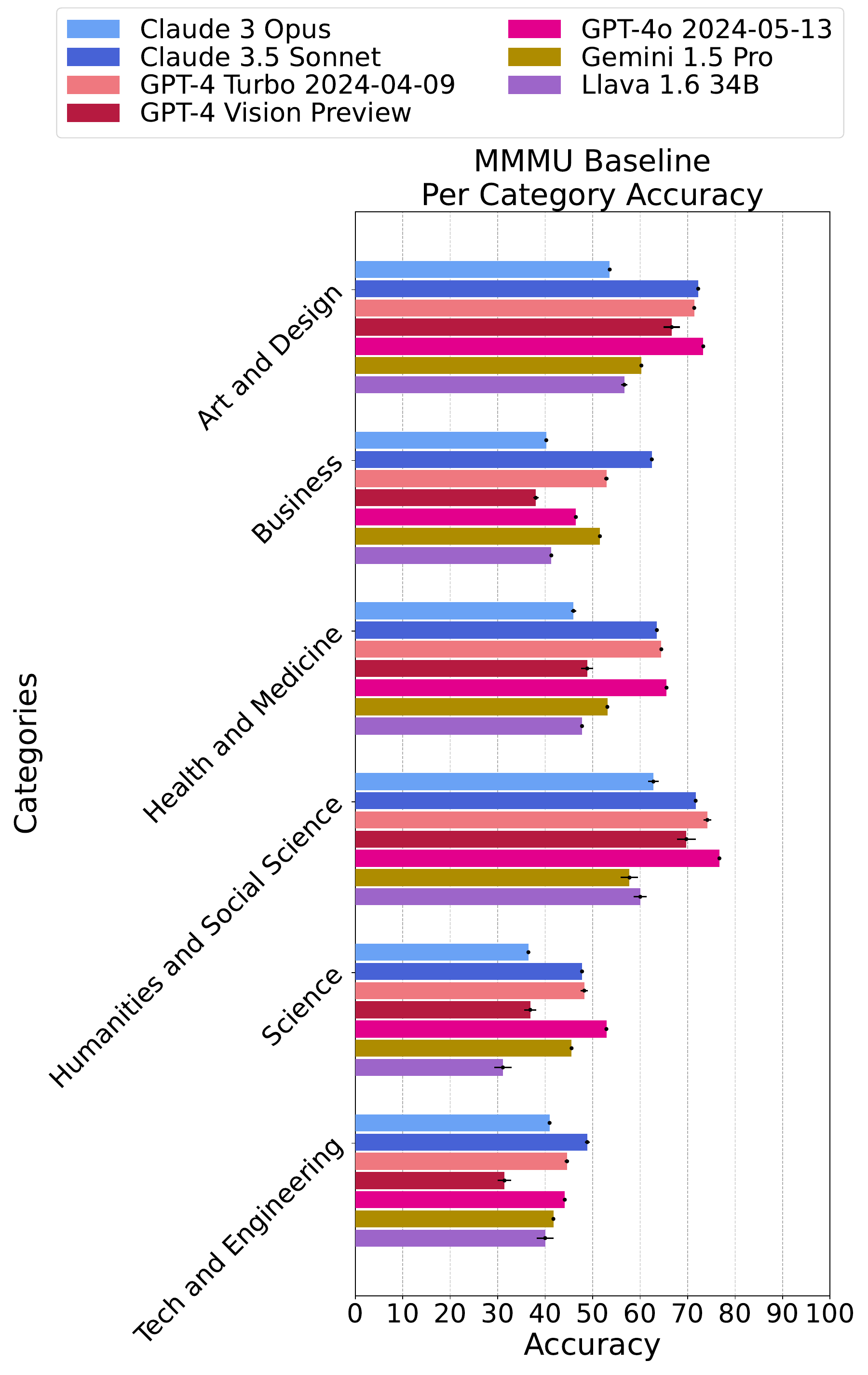}\vspace{-15pt}
    \caption{Accuracy per discipline reported across three different runs per model.}
    \label{fig:mmmu_discipline_baseline_accuracy}
\end{figure}

\noindent \textbf{Discipline Level Baseline Results:} Figure~\ref{fig:mmmu_discipline_baseline_accuracy} present the discipline-level accuracy, as defined by the six core disciplines in the benchmark: Art \& Design, Business, Science, Health \& Medicine, Humanities \& Social Science, and Tech \& Engineering.  Here we see the close performance of the top two models \ClaudeSonnet and \GPTFourO, with different per-subject wins. \ClaudeSonnet has big leads in Business and Tech and Engineering but lags \GPTFourO in all other disciplines, with \GPTFourO having the best relative performance in Science and Humanities and Social Sciences. Science, Tech and Engineering remain the most difficult disciplines across all models.

\noindent \textbf{Aggregate Zero-shot CoT and Expert Prompt Results:} We note than many current evaluations on MMMU use engineered prompts for better performance, leading to higher accuracy results than shown in the previous section. Thus, to provide some insight into this we performed our evaluations on the top model in each family using a modified system-level \emph{expert} instruction prompt, as specified in the OpenAI Evals codebase~\cite{OpenAIEvals}.

\begin{itemize}[leftmargin=*]
    \item Multiple-choice: \textit{You are an expert in \{subject\} whose job is to answer questions from the user using images. First, reason about the correct answer. Then write the answer in the following format where X is exactly one of A,B,C,D: ``ANSWER: X". If you are uncertain of the correct answer, guess the most likely one.}
    \item Open-ended: \textit{You are an expert in \{subject\} whose job is to answer questions from the user using images. First, reason about the correct answer. Then write the answer in the following format where X is only the answer and nothing else: ``ANSWER: X"}
\end{itemize}
Here \{subject\} is one of the 30 subjects associated with each question, e.g., Art, Chemistry, Clinical Medicine, etc.

Next, we remove the instruction component from the user prompt that was used in the baseline results, as there is now an instruction prompt in the system prompt, thus the following prompts:

\begin{itemize}[leftmargin=*]
    \item Multiple-choice: ``\{prompt\} Answer with the option's letter from the given choices directly."
    \item Open-ended ``\{prompt\} Answer the question using a single word or phrase." 
\end{itemize}


are changed to have a Zero-Shot Chain-of-Thought (CoT) prompt, as such:

\begin{itemize}[leftmargin=*]
    \item Multiple-choice and Open-ended: \textit{\{prompt\} Let’s think step by step.}
\end{itemize}
\begin{figure}[t!]
    \centering
    \includegraphics[width=0.49\linewidth]{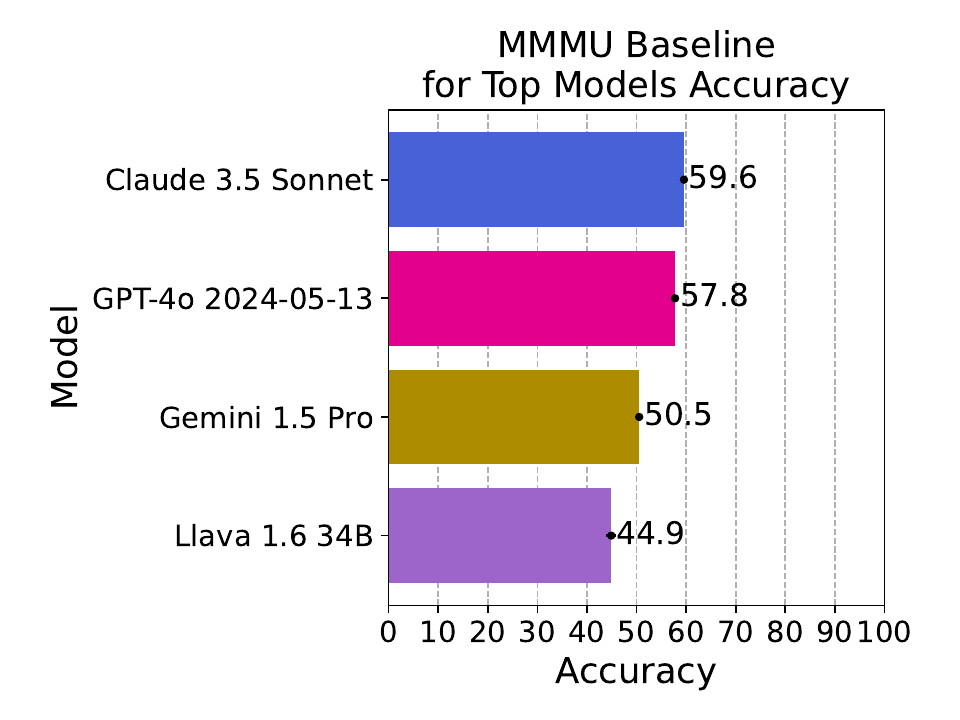}
    \includegraphics[width=0.49\linewidth]{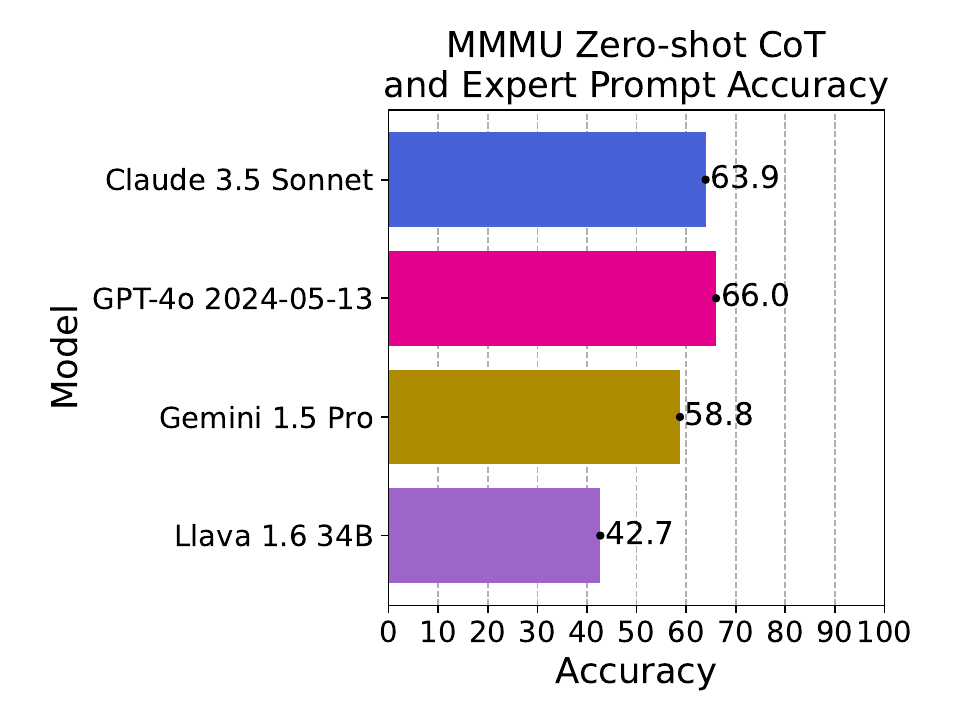}
    \caption{Aggregate accuracy for top models per family comparing the Baseline and Zero-shot CoT and Expert Prompt conditions. Reported across three different runs per model.}
    \label{fig:mmmu_expert_baseline_accuracy}
\end{figure}

\begin{figure}[t!]
    \centering
    \includegraphics[width=0.98\linewidth]{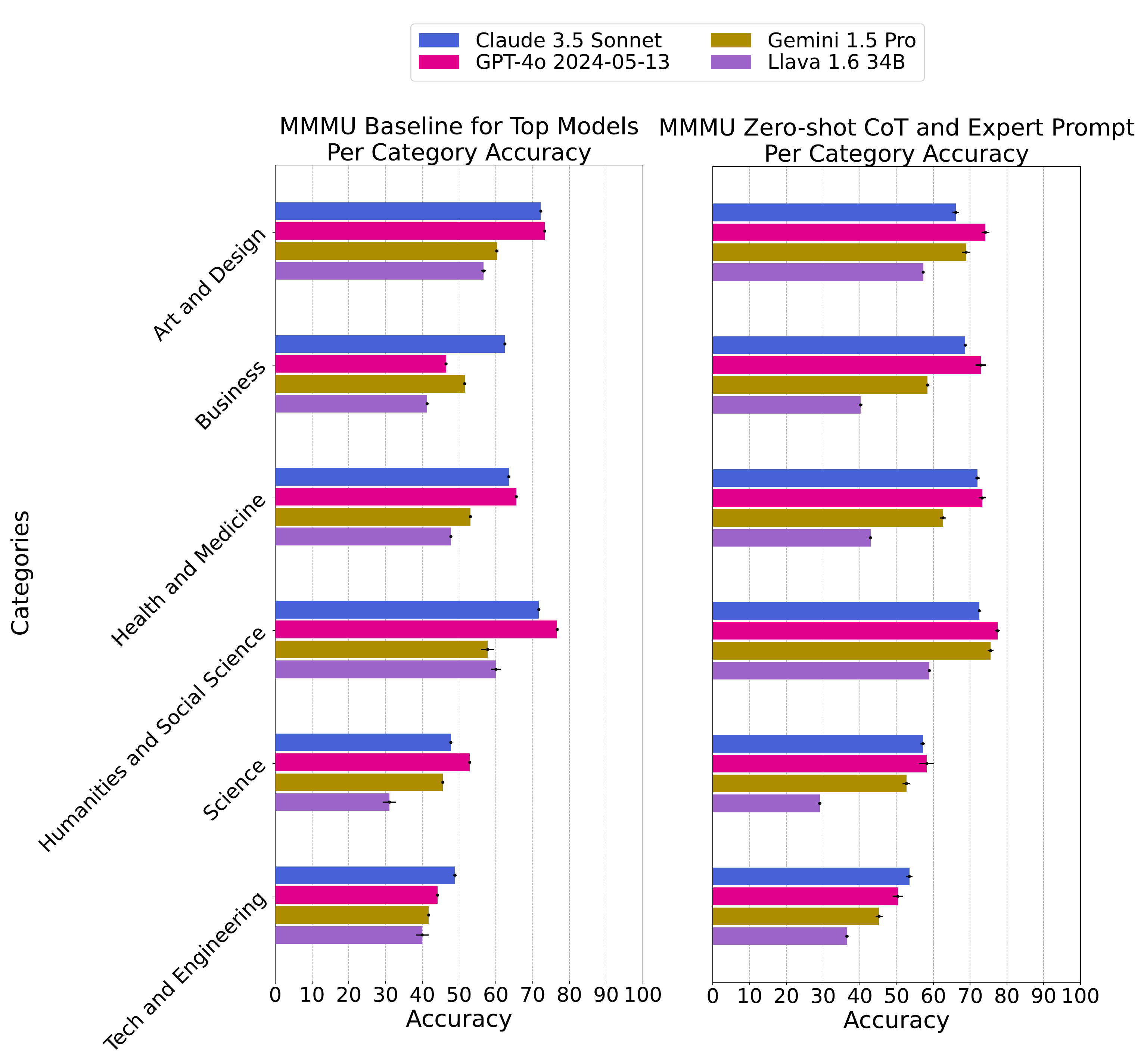}\vspace{-10pt}
    \caption{Accuracy per discipline for top models per family comparing the Baseline and Zero-shot CoT and Expert Prompt conditions. Reported across three different runs per model.}
    \label{fig:mmmu_discipline_expert_accuracy}
\end{figure}

As shown in Figure~\ref{fig:mmmu_expert_baseline_accuracy}, with the Zero-shot CoT Expert Prompt strategy, most models have a significant performance boost over the baseline. Largest increases are observed for \GeminiPro and \GPTFourO, which have a 8.3\% increase in accuracy each. \ClaudeSonnet has the smallest increase of 4.3\%. \GPTFourO is the best performing model by 2.1\%. 
\clearpage
The \Llava model is the only one whose accuracy decreases. \Llava is not specifically trained to handle a system role, thus the system prompt is concatenated with the user prompt~\cite{LLAVAHF}, which is a likely explanation for the different behavior over other models.  

\noindent \textbf{Discipline Level Zero-shot CoT and Expert Prompt Results:} As shown in Figure ~\ref{fig:mmmu_discipline_expert_accuracy}, the accuracy of \GPTFourO increases noticeably in every discipline, from which the largest increases are observed in Business. \ClaudeSonnet has more modest gains and has a regression on Art \& Design.

\subsubsection*{Main takeaways}
\noindent\fbox{%
    \parbox{\textwidth}{%
        \begin{itemize}[leftmargin=*]
        \item \ClaudeSonnet and \GPTFourO are the leading models for multimodal question answering as measured by the MMMU dataset, indicating better multimodal understanding skills and knowledge.
        \item The MMMU benchmark remains a challenging task for all models, with the best performance in the mid 60s percentage range.
        \item The performance of models is highly dependent on how they are prompted, with improvements and regressions across topics. The role of the prompt in evaluations cannot be ignored. This is an area that requires further investigation.
        \end{itemize}
    }%
}
\begin{figure}[t!]
\centering
    \includegraphics[width=.98\textwidth]{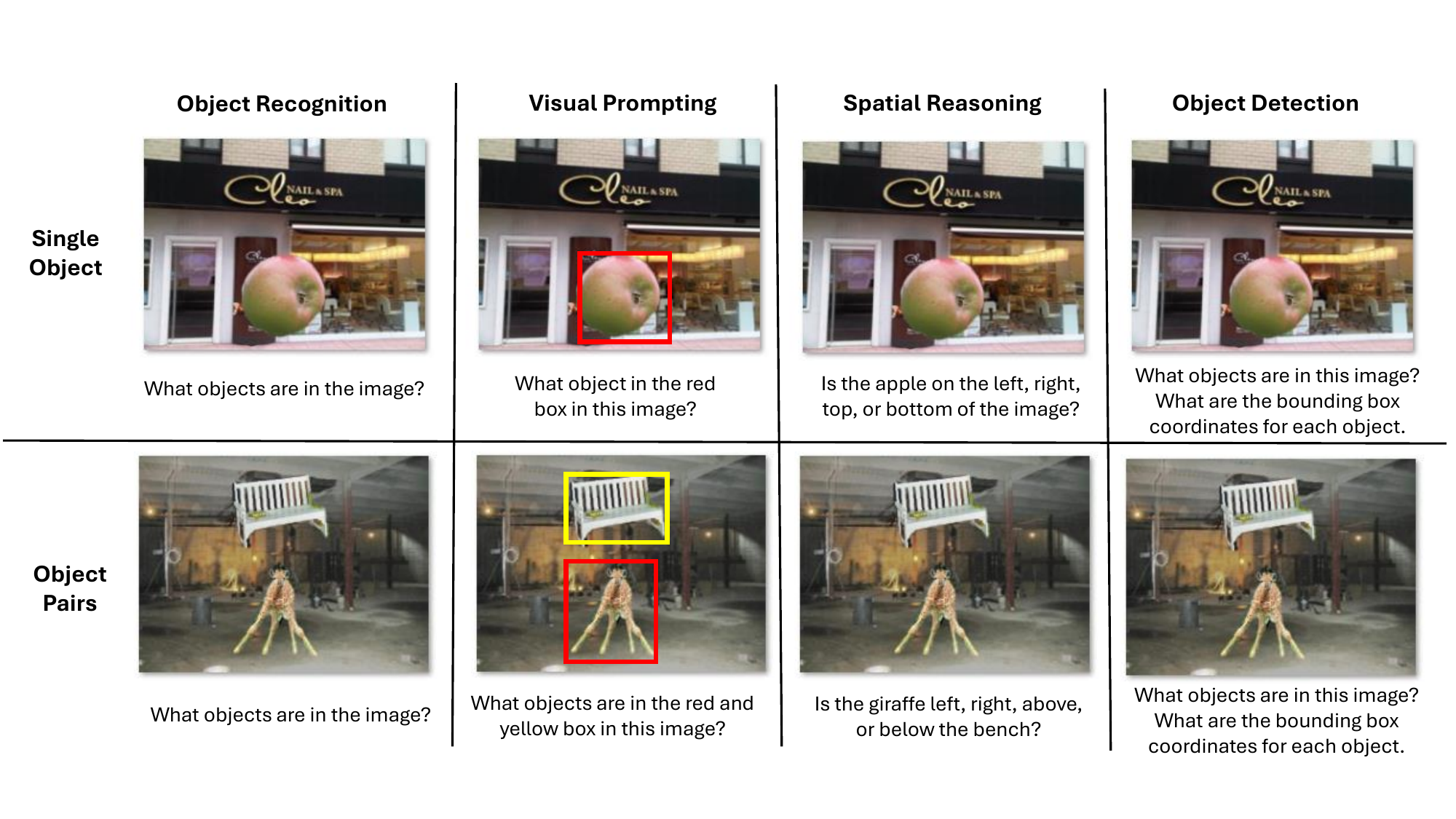}
\caption{Illustration of the tasks in the Image Understanding Benchmark: Object Recognition, Visual Prompting, Spatial Reasoning, and Object Detection. For each task, the images consist of images of pasted objects on random images. For each there are 2 conditions ``single object'' and ``object pairs''. Each test set has 1280 images and text pairs.\vspace{-.2cm}}
\label{fig:image_understanding}
\end{figure}

\subsection{Image Understanding}
\label{sec:image_understanding}
\noindent \textbf{Motivation:} A key question for understanding multimodal performance is analyzing the ability for a model to have basic vs. detailed understanding of images. These capabilities are needed for models to be used in real-world tasks, such as an assistant in the physical world. While there are many dataset for object detection and recognition, there are few that test spatial reasoning and other more targeted task such as visual prompting.  The datasets that do exist are static and publicly available, thus there is concern that current AI models could be trained on these datasets, which makes evaluation with them unreliable. Thus we created a dataset that is  procedurally generated and synthetic, and tests spatial reasoning, visual prompting, as well as object recognition and detection~\cite{ImageUnderstanding}. The datasets are challenging for most AI models and by being procedurally generated the benchmark can be regenerated ad infinitum to create new test sets to combat the effects of models being trained on this data and the results being due to memorization.

\noindent \textbf{Benchmark Description:} This dataset has 4 sub-tasks: Object Recognition, Visual Prompting. Spatial Reasoning, and Object Detection. For each sub-task, the images consist of images of pasted objects on random images.  The objects are from the COCO~\cite{lin2014microsoft} object list and are gathered from internet data.  Each object is masked using the DeepLabV3 object detection model~\cite{DeepLabV3} and then pasted on a random background from the Places365 dataset~\cite{zhou2017places}. The objects are pasted in one of four locations, top, left, bottom, and right, with small amounts of random rotation, positional jitter, and scale.  

There are 2 conditions `` single'' and `` pairs'', for images with one and two objects. Each test set uses 20 sets of object classes (either 20 single objects or 20 pairs of objects), with four potential locations and four backgrounds classes, and we sample 4 instances of object and background. This results in 1280 images per condition and sub-task.  An example of each is shown in Figure~\ref{fig:image_understanding}, and examples of the prompts are as follows:

\begin{itemize}[leftmargin=*]
    \item Object Recognition: \textit{What objects are in this image?}
    \item Visual Prompting:
    \begin{itemize}
    \item One Object: \textit{What object is in the red box in this image?}
    \item Two Objects: \textit{What objects are in the red and yellow box in this image?}    
    \end{itemize}       
    \item Spatial Reasoning: 
    \begin{itemize}
    \item One Object: \textit{Is the potted plant on the right, top, left, or bottom of the image? Answer with one of (right, bottom, top, or left) only.}
    \item Two Objects: \textit{Is the bottle above, below, right, or left of the keyboard in the image? Answer with one of (below, right, left, or above) only.}    
    \end{itemize}       
    \item Object Detection: \textit{You are an object detection model that aims to detect all the objects in the image.  Definition of Bounding Box Coordinates:  The bounding box coordinates (a, b, c, d) represent the normalized positions of the object within the image:  a: The x-coordinate of the top-left corner of the bounding box, expressed as a percentage of the image width. It indicates the position from the left side of the image to the object's left boundary. The a ranges from 0.00 to 1.00 with precision of 0.01. b: The y-coordinate of the top-left corner of the bounding box, expressed as a percentage of the image height. It indicates the position from the top of the image to the object's top boundary. The b ranges from 0.00 to 1.00 with precision of 0.01. c: The x-coordinate of the bottom-right corner of the bounding box, expressed as a percentage of the image width. It indicates the position from the left side of the image to the object's right boundary. The c ranges from 0.00 to 1.00 with precision of 0.01. d: The y-coordinate of the bottom-right corner of the bounding box, expressed as a percentage of the image height. It indicates the position from the top of the image to the object's bottom boundary. The d ranges from 0.00 to 1.00 with precision of 0.01.  The top-left of the image has coordinates (0.00, 0.00). The bottom-right of the image has coordinates (1.00, 1.00).  Instructions: 1. Specify any particular regions of interest within the image that should be prioritized during object detection. 2. For all the specified regions that contain the objects, generate the object's category type, bounding box coordinates, and your confidence for the prediction. The bounding box coordinates (a, b, c, d) should be as precise as possible. Do not only output rough coordinates such as (0.1, 0.2, 0.3, 0.4). 3. If there are more than one object of the same category, output all of them. 4. Please ensure that the bounding box coordinates are not examples. They should really reflect the position of the objects in the image. 5. Report your results in this output format: (a, b, c, d) - category for object 1 - confidence (a, b, c, d) - category for object 2 - confidence ... (a, b, c, d) - category for object n - confidence.}
\end{itemize}

\begin{figure}[t!]
    \centering
    \includegraphics[width=0.48\linewidth]{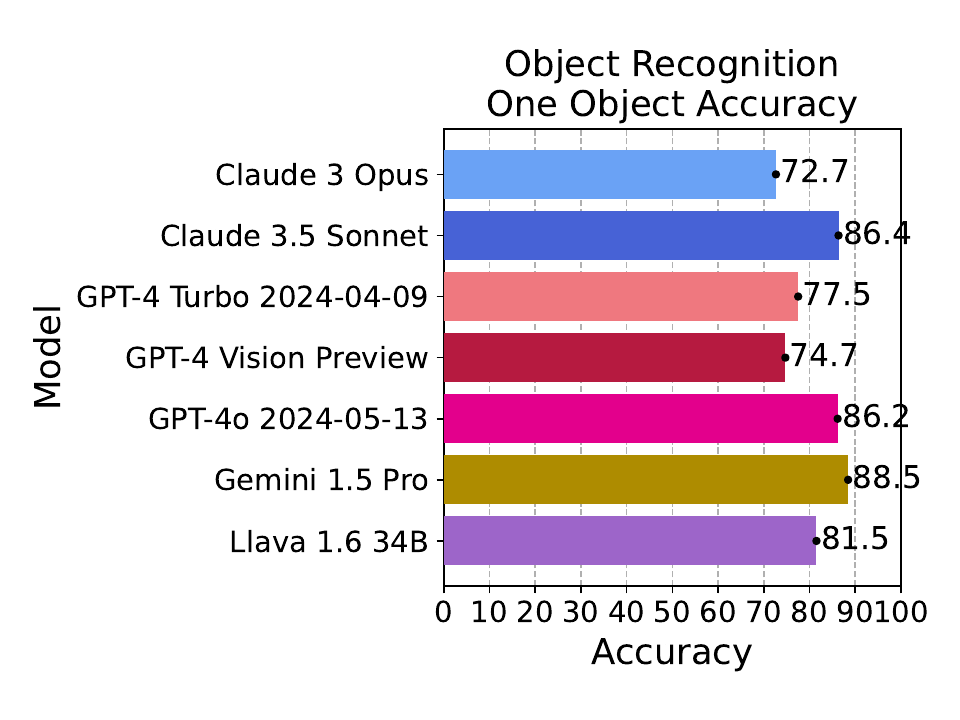}
    \includegraphics[width=0.48\linewidth]{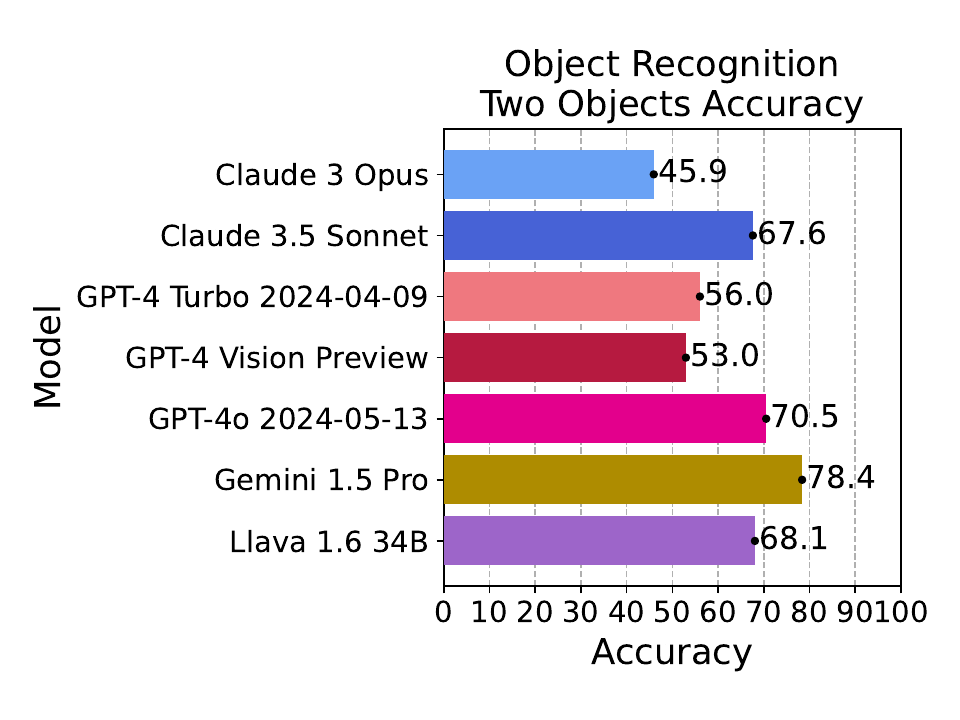}\\\vspace{-10pt}
    \includegraphics[width=0.48\linewidth]{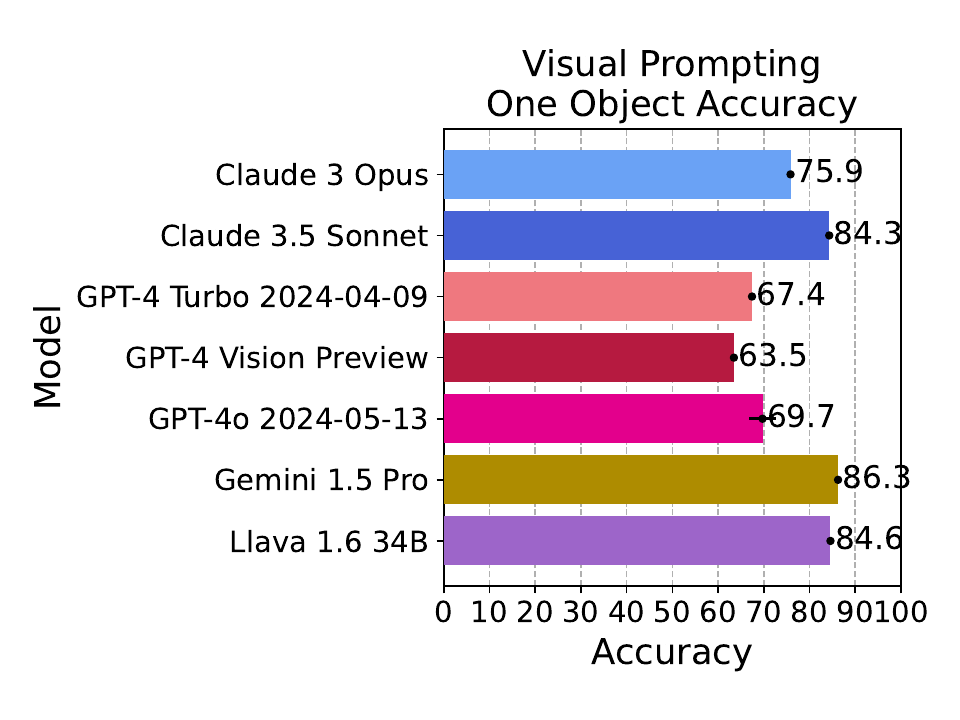}
    \includegraphics[width=0.48\linewidth]{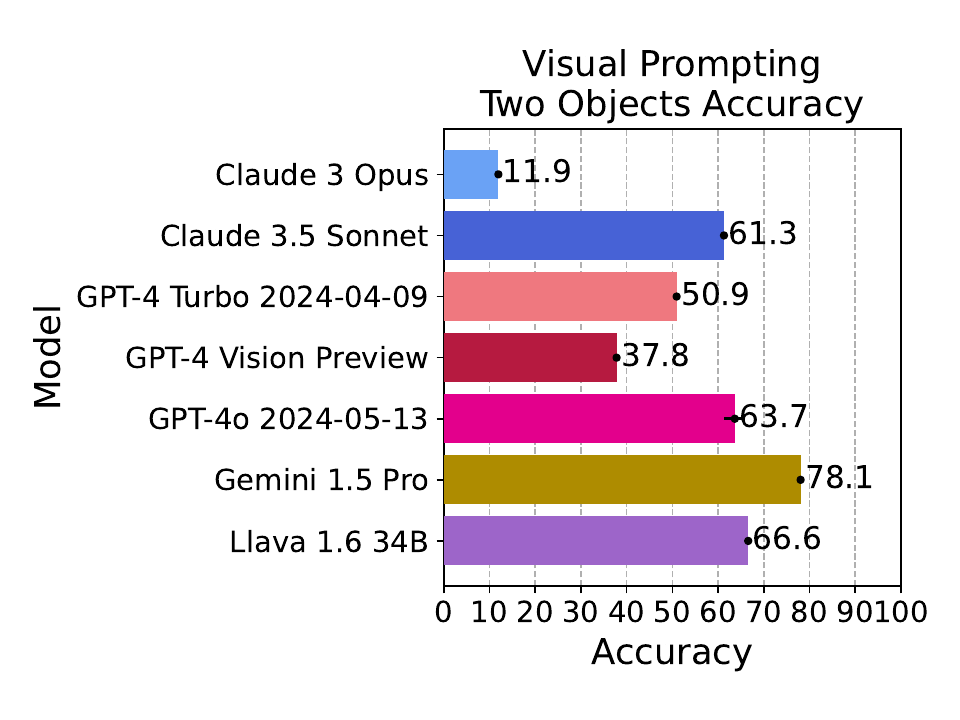}\\\vspace{-10pt}      
    \includegraphics[width=0.48\linewidth]{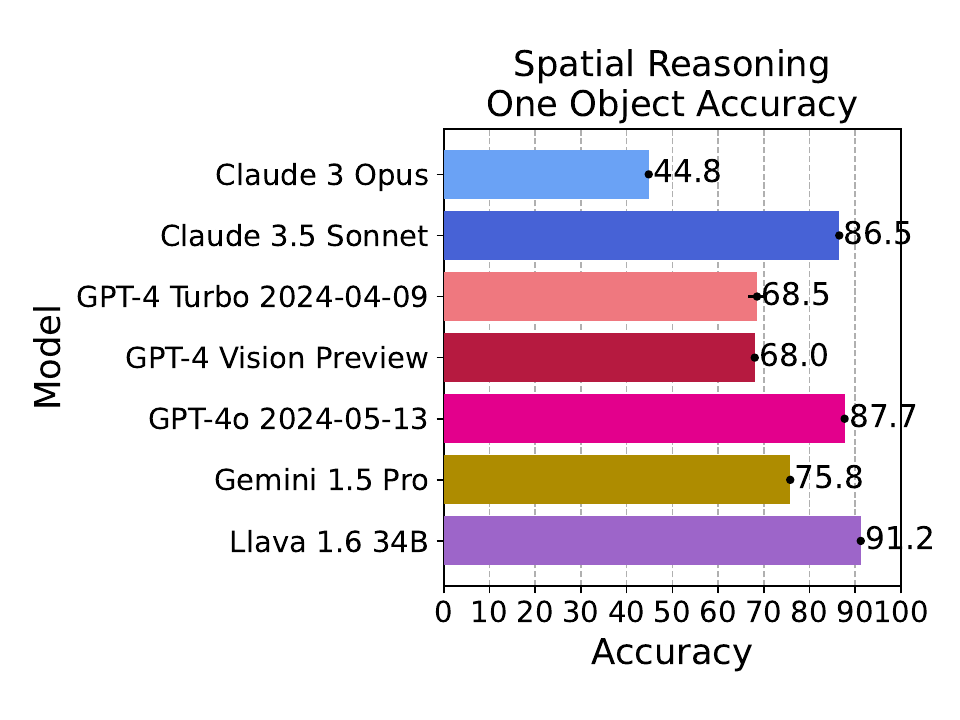}
    \includegraphics[width=0.48\linewidth]{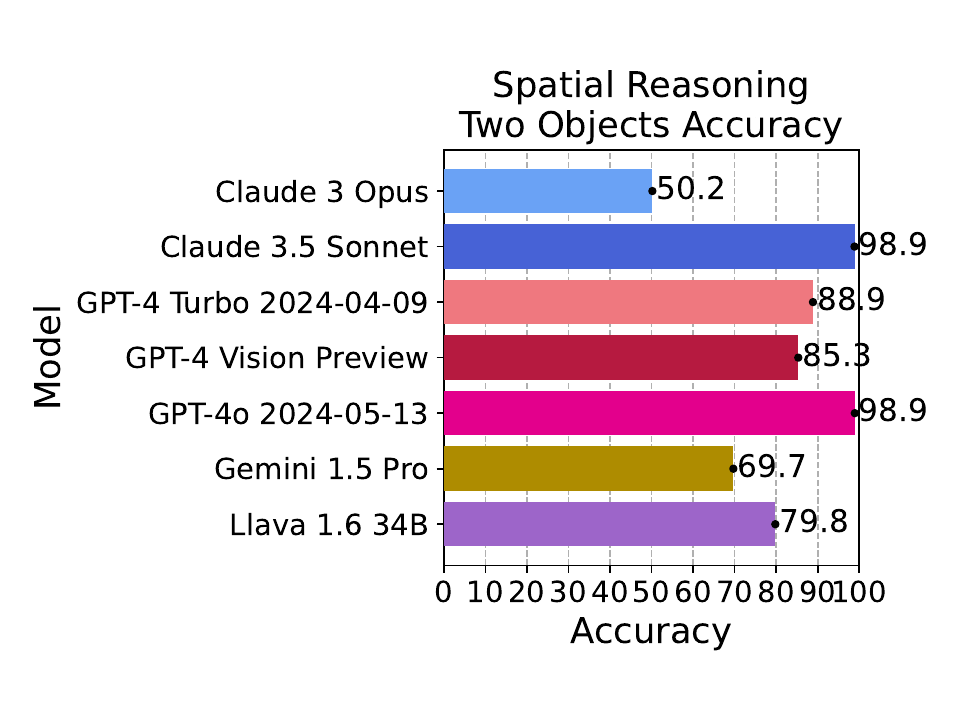}\\\vspace{-10pt}  
    \includegraphics[width=0.48\linewidth]{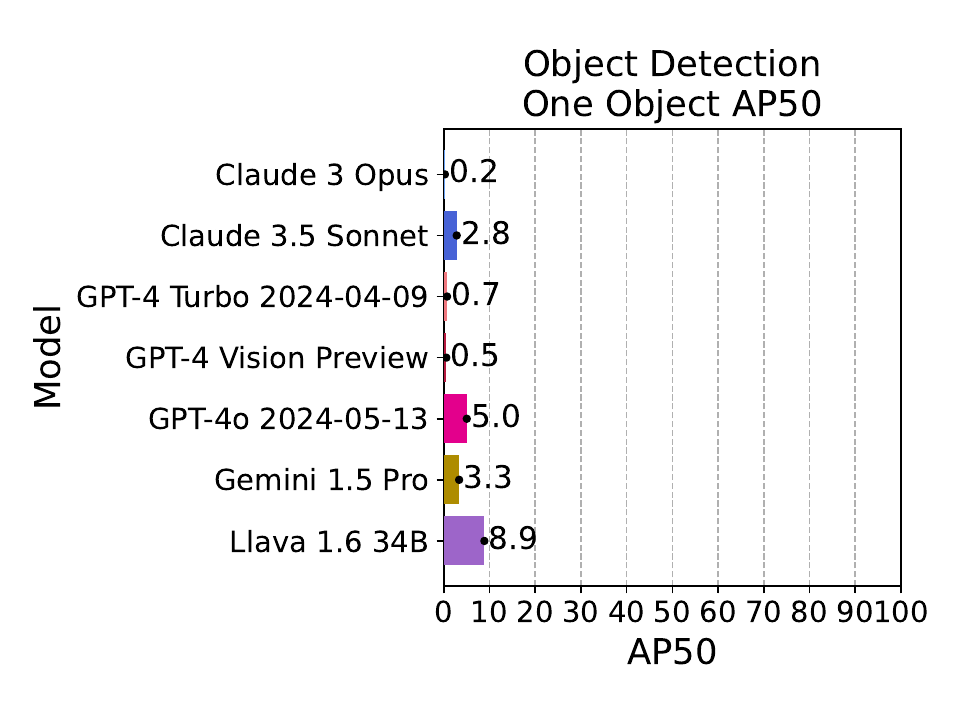}
    \includegraphics[width=0.48\linewidth]{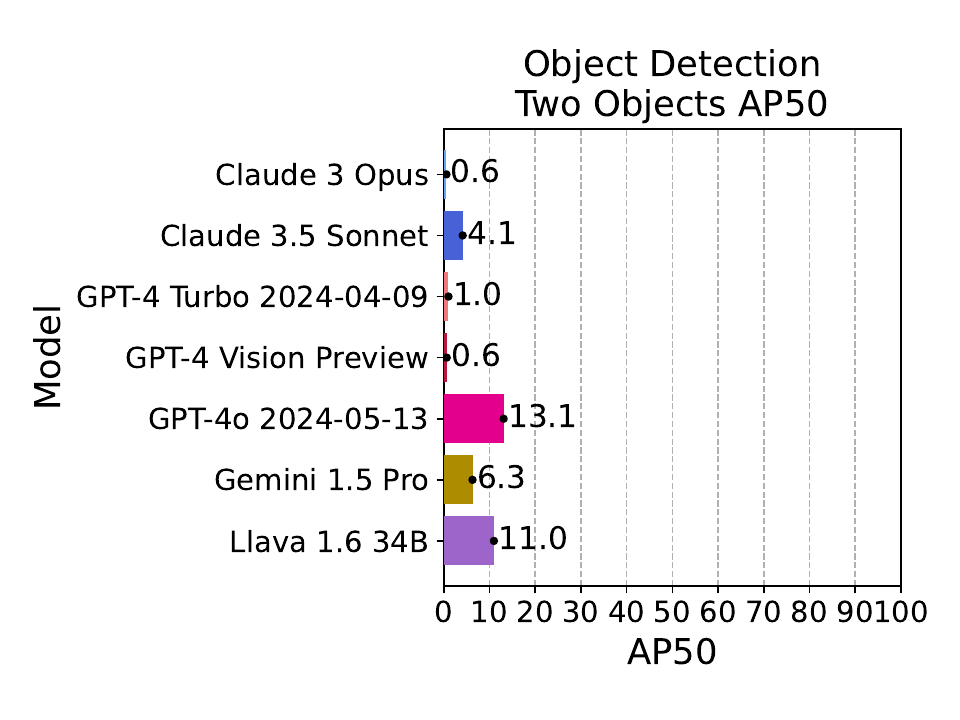}    
    \caption{Image Understanding accuracy broken down per sub-task and condition reported across three different runs per model. 
    }
    \label{fig:image_understanding_subtask}
\end{figure}

\noindent \textbf{Results:} As shown in Figure~\ref{fig:image_understanding_subtask}, for Object Recognition and Visual Prompting, \GeminiPro is consistently the best model by a range of 2-12\% across these tasks and the one and two object conditions. \GPTFourO, \ClaudeSonnet, and \Llava all come in second within a few percentage points of each other, without a consistent clear winner between them.

One interesting observation is that Visual Prompting leads to a small drop in model performance relative to Object Recognition. In cases involving two objects, we see a substantial decline across nearly all models, while in one-object cases, there is a modest drop for some models. This is surprising given we would expect the visual prompt to help focus the model and thus improve results~\cite{subramanian2022reclipstrongzeroshotbaseline, yao2022cptcolorfulprompttuning}. 

For Spatial Reasoning, we see \GPTFourO as the overall best across both the one and two object conditions; however, within each condition there is a different story.  \Llava excels at the one object condition, with 91.2\% accuracy, 3.5\% better than \GPTFourO, but is far behind \GPTFourO and \ClaudeSonnet with two objects, which are tied at 98.9\%. The Spatial Reasoning task is interesting as there are incredible gains in performance with new models in each family. For example, the one object condition had 50-60\% accuracy in earlier models (\ClaudeOpus, \GPTFourTurboApril), and is now in the 80-th percentiles for accuracy.

The average accuracy for the top models across Object Recognition, Visual Prompting, and Spatial Reasoning and the one and two object conditions are \ClaudeSonnet at 80.83\%, \GeminiPro at 79.47\%, \GPTFourO at 79.45\%, and \Llava at 78.63\%. \GeminiPro has the most wins for these three sub-tasks and conditions, but \ClaudeSonnet has the best average performance by a little over 1\%.

Object Detection is a task that requires very detailed level image understanding and localization and this is one task where all models perform poorly. The best performing model is \GPTFourO at AP50 13.1 in the two object condition and this is far below the performance of a state-of-the-art object detector~\cite{CoDETR}.

When we add Object Detection in the mix and consider average performance across all sub-tasks and conditions, the order flips, due to the stronger object detection performance of \GPTFourO and \Llava with \GPTFourO at 61.85\%, \ClaudeSonnet at 61.49\%, \Llava at 61.46\%, and \GeminiPro at 60.8\%.

\subsubsection*{Main takeaways}
\noindent\fbox{%
    \parbox{\textwidth}{%
        \begin{itemize}[leftmargin=*]
        \item \GPTFourO, \ClaudeSonnet, \GeminiPro, and \Llava all perform well for Image Understanding as measured by our dataset. \GeminiPro has the most wins for these sub-tasks and conditions, but \ClaudeSonnet has the best average performance by a little over 1\% when excluding Object Detection. \GPTFourO has the best average performance for all sub-tasks and conditions.
        \item Object Detection is still quite challenging for all models.  The best performing model is \GPTFourO at AP50 13.1 in the two object condition and this is far below the performance of a state-of-the-art object detector.
        \item Object Recognition and Visual Prompting become more difficult for more than one object, but Spatial Understanding and Object Detection become easier.  Models perform slightly work on Visual Prompting vs. Object Recognition, which is unexpected.
        \end{itemize}
    }%
}
\begin{figure}[t!]
\centering
    \includegraphics[width=.8\textwidth]{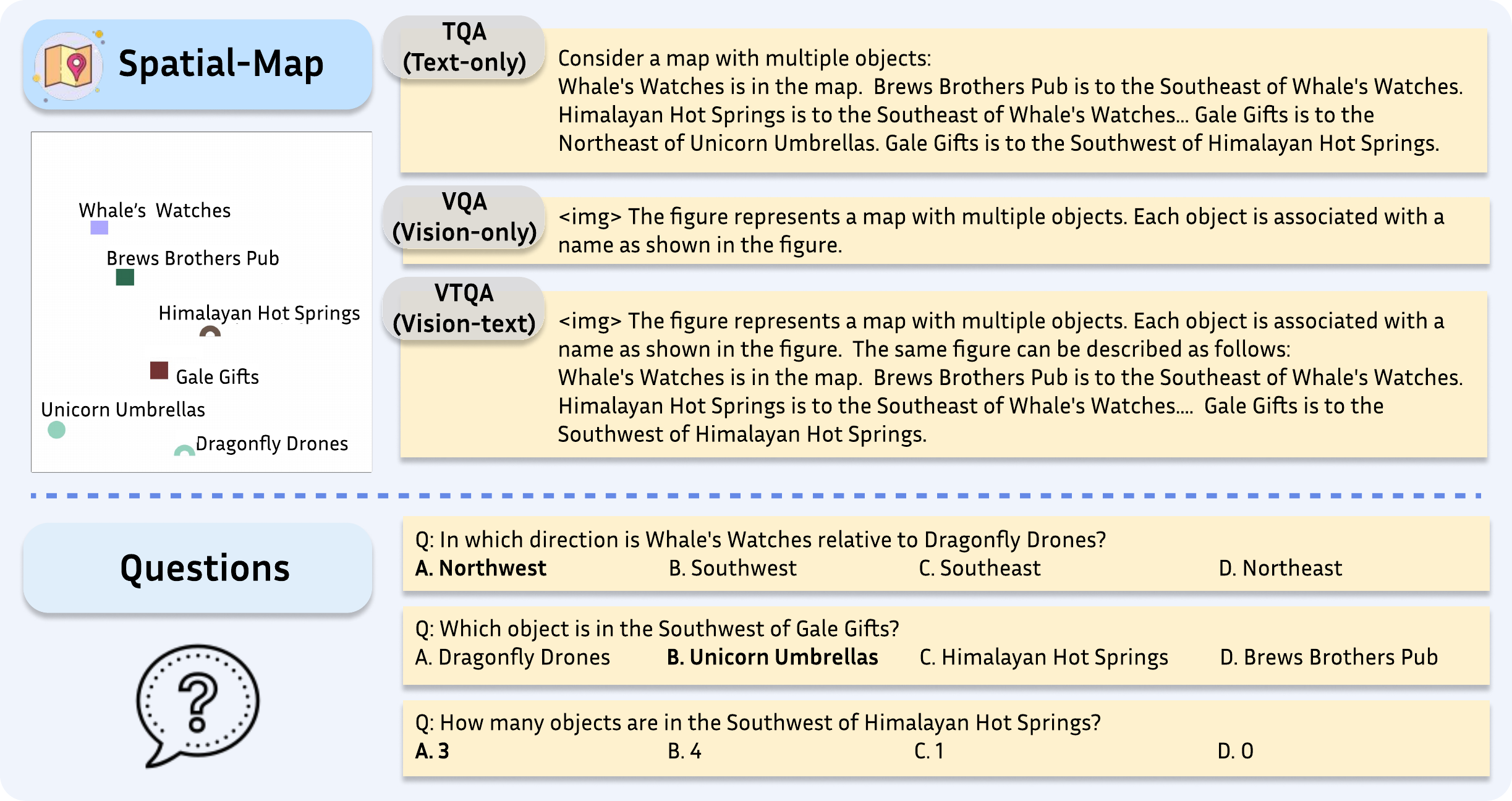}\\
    \includegraphics[width=.8\textwidth]{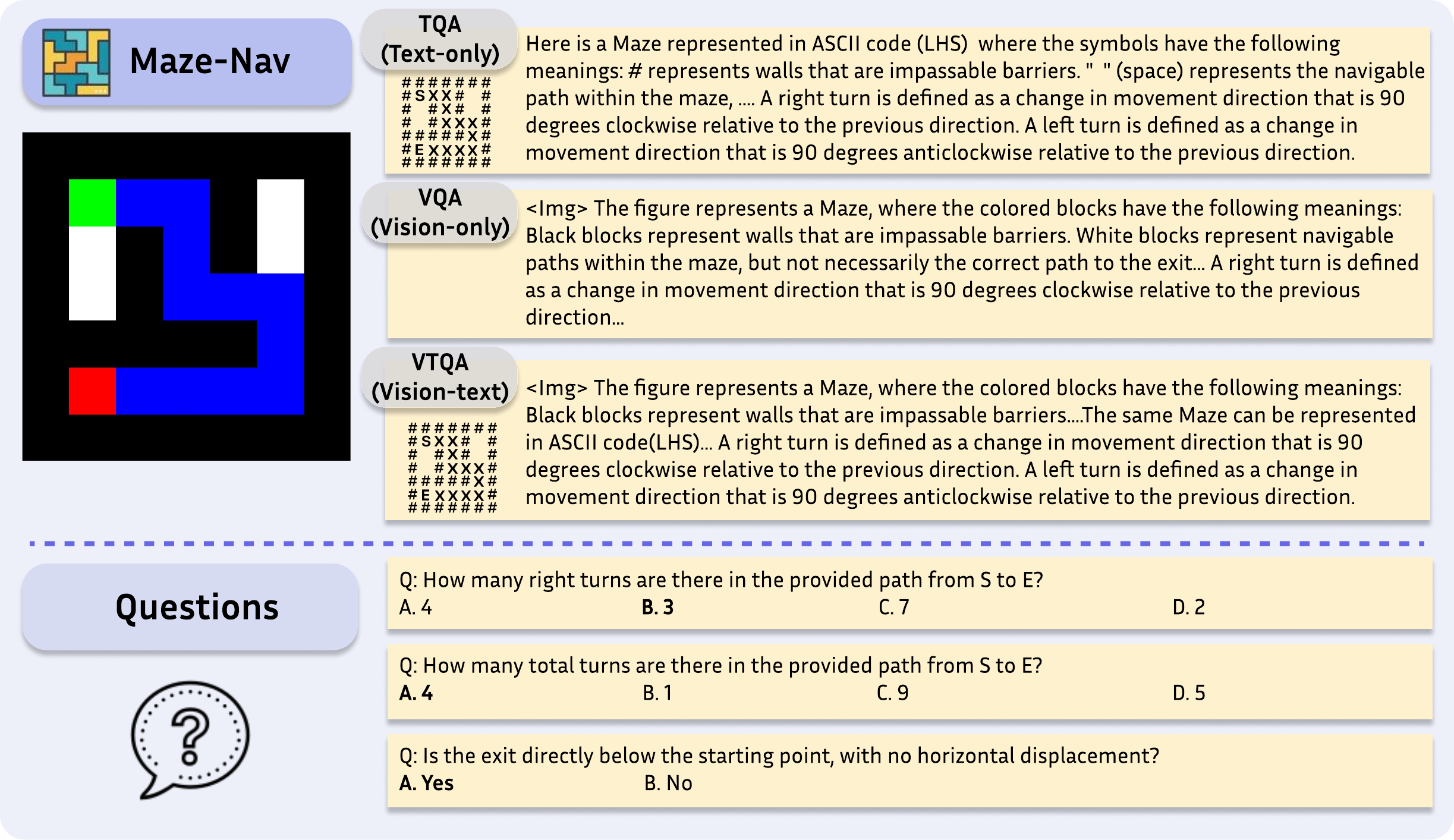} \\
    \includegraphics[width=.8\textwidth]{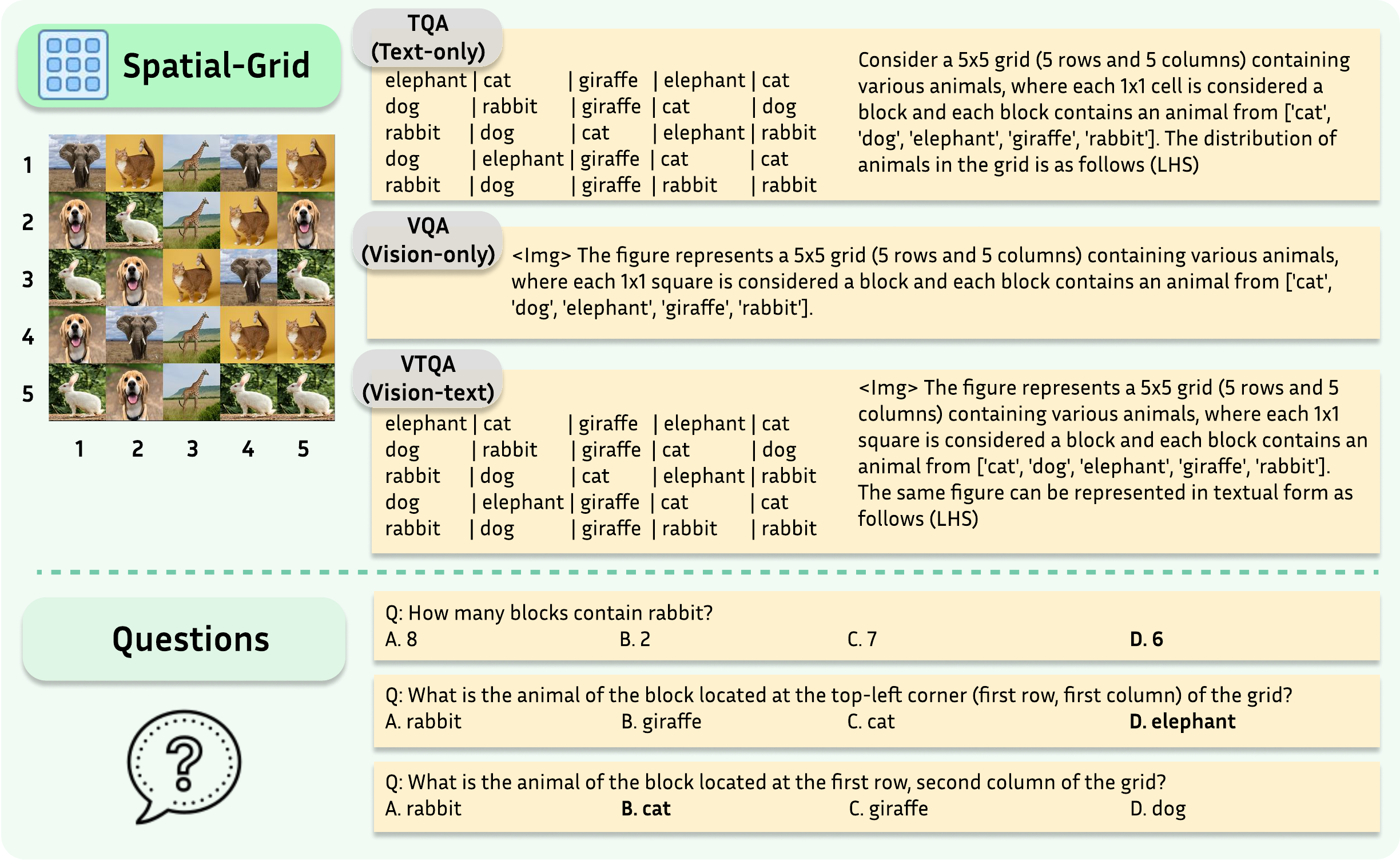}
\caption{Illustration of the Spatial-Map (spatial understanding), Maze-Nav (navigation), and Spatial-Grid (counting) tasks. To investigate the impact of modality, we consider three input formats: Text-only, Vision-only, and Vision-text.}
\label{fig:vl}
\end{figure}

\subsection{Vision Language Understanding}
\label{sec:vl_understanding}

\noindent \textbf{Motivation:} A key question for understanding multimodal vs. language capabilities of models is what is the relative strength of the spatial reasoning and understanding in each modality, as spatial understanding is expected to be a strength for multimodality? To test this we use the procedurally generatable, synthetic dataset of Wang et al.~\cite{wang2024pictureworththousandwords} to testing spatial reasoning, navigation, and counting. These datasets are challenging and by being procedurally generated new versions can easily be created to combat the effects of models being trained on this data and the results being due to memorization. For each task, each question has an image and a text representation that is sufficient for answering each question. 

\noindent \textbf{Benchmark Description:} This dataset has three tasks that test: Spatial Understanding (Spatial-Map), Navigation (Maze-Nav), and Counting (Spatial-Grid).  Each task has three conditions, with respect to the input modality, 1) text-only, input and a question, 2) vision-only, which is the standard task of visual-question answering that consists of a vision-only input and a question, and 3) vision-text includes both text and image representations with the question. See Figure~\ref{fig:vl} for an illustration of each task. Each condition includes 1500 images and text pairs for a total of 4500.
\begin{itemize}[leftmargin=*]
\item \textbf{Spatial Map:} The dataset consists of spatial relationships for random layouts of symbolic objects with text names on  white background. 
Each object is associated with a unique location name, such as Unicorn Umbrellas and Gale Gifts. To study the impact of modality,
the textual representation of each input consists of pairwise relations such as ``Brews Brothers Pub
is to the Southeast of Whale’s Watches''. The questions include asking about the spatial
relationships between two locations and the number of objects that meet specific spatial criteria.

\item \textbf{Maze Navigation:} The dataset consists of small mazes with questions asked about the maze. Each sample can be 
represented as colored blocks where different colors signify distinct elements: \emph{``a green block marks
the starting point (S), a red block indicates the exit (E), black blocks represent impassable walls,
white blocks denote navigable paths, and blue blocks trace the path from S to E. The objective is to
navigate from S to E following the blue path, with movement permitted in the four cardinal directions
(up, down, left, right).''} Alternatively, each input can be depicted in textual format using ASCII code.
The questions asked include counting the number of turns from S to E and determining the spatial relationship 
between S and E. 

\item \textbf{Grid Counting:} Each input consists of a grid of cells, each cell containing an image (e.g., a rabbit). Alternatively, this grid 
can also be represented in a purely textual format; for instance, the first row might be described as: 
elephant | cat | giraffe | elephant | cat. The evaluations focus on tasks such as counting specific objects (e.g., rabbits) and
identifying the object located at a specific coordinate in the grid (e.g., first row, second column).
\end{itemize}

For more details on this dataset and further in-depth analysis please refer to~\cite{wang2024pictureworththousandwords}.

\begin{figure}[t!]
\centering
    \includegraphics[width=.99\textwidth]{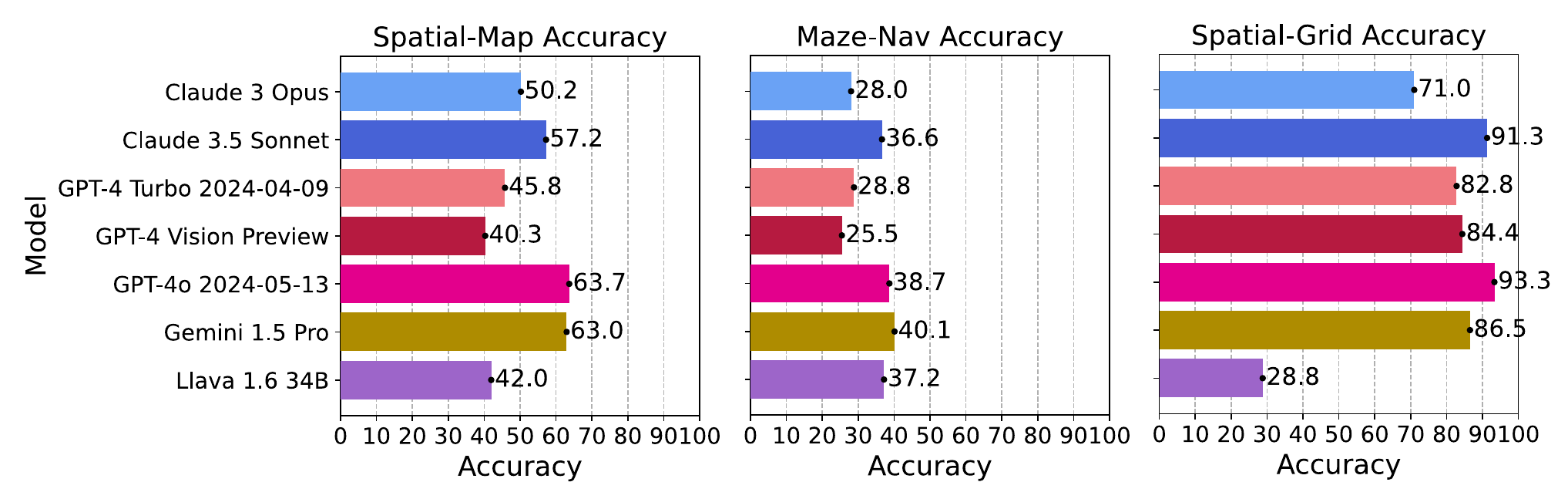}
\caption{Aggregate accuracy on the Spatial-Map, Maze-Nav, and Spatial-Grid tasks.}
\label{fig:vl_results}
\end{figure}

\begin{figure}[t!]
\centering
    \includegraphics[width=1\textwidth]{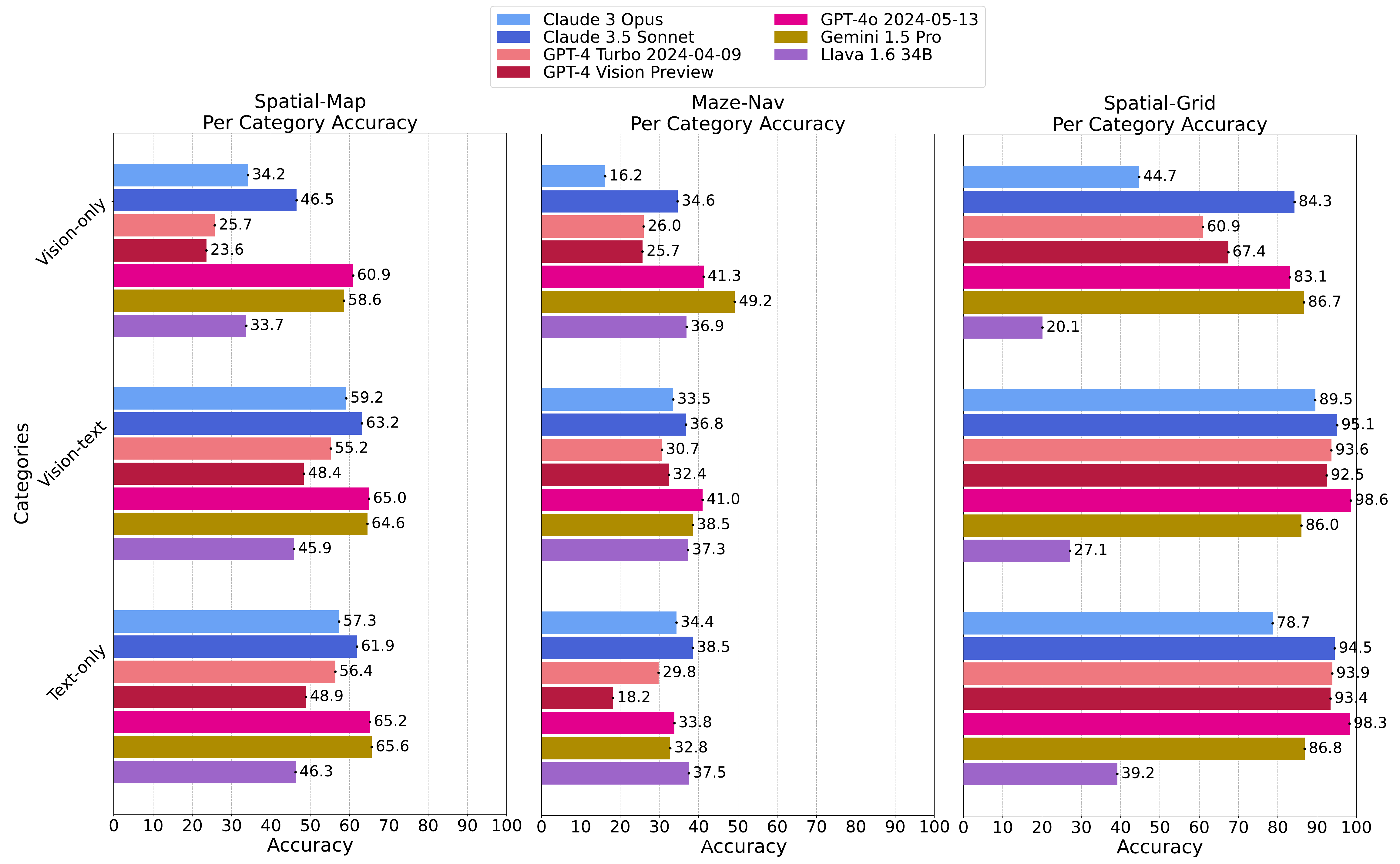}
\caption{Vision-Only, Text-Only, and Vision-Text accuracy on the Spatial-Map, Maze-Nav, and Spatial-Grid. }
\label{fig:vl_disaggregation_results}
\end{figure}

\noindent\textbf{Aggregate Results:} As seen in Figure~\ref{fig:vl_results}, when examining the aggregate results across all conditions, we see that \GPTFourO has the best performance by 0.7\% over \GeminiPro for Spatial-Map, while \GeminiPro performs the best on Maze-Nav by 1.4\% over \GPTFourO, and \GPTFourO edges out \ClaudeSonnet by 2\% on Spatial-Grid.  Overall, we see that Spatial-Map and Maze-Nav are challenging task for all models, with \GPTFourO and \GeminiPro being fairly comparable.  Most models do well on Spatial-Grid except for \Llava, which is far behind all models on all tasks. 

\noindent\textbf{Vision-Only, Text-Only, and Vision-Text Results:} Figure~\ref{fig:vl_disaggregation_results} shows the vision-only, text-only, and vision-text results for Spatial-Map, Maze-Nav, and Spatial-Grid tasks, respectively.  For the Spatial-Map and Spatial-Grid there is a consistent trend that the vision-only condition under-performs when compared to the text-only and vision-text conditions, with no consistent winner between the text-only and vision-text conditions.  For Maze-Nav there is a similar trend for most models, except \GeminiPro which has far better vision-only performance than with the other conditions, including the vision-text condition.

\subsubsection*{Main takeaways}
\noindent\fbox{%
    \parbox{\textwidth}{%
        \begin{itemize}[leftmargin=*]
        \item The vision-only conditions generally under-perform when compared to the text-only and vision-text conditions, with no consistent winner between the text-only and vision-text conditions. 
        \item When both textual and visual information are available, multi-modal language models appear to rely less on visual information if sufficient textual clues are provided. This opens important questions for multimodal learning, as for humans most tasks in this benchmark are easier in the vision modality but current models do not seem to benefit from it.
        \end{itemize}
    }%
}

\subsection{High Level vs. Detailed Image Understanding - Discussion}

Current frontier models excel on tasks that require high-level image understanding, such as general object recognition, counting in a grid, and basic spatial reasoning (with 1 or 2 objects), with results in the 80-90\% accuracy range. In contrast, these models struggle when it comes to tasks that require detailed image understanding, including tasks such as object detection, complex spatial reasoning, maze navigation, and depth and height perception tasks. For example, \emph{the best object detection result we measured of 13.1 AP50, is around 30 points below a Computer Vision detector from almost 10 years ago} ~\cite{NIPS2015_14bfa6bb}. Spatial reasoning with more than two objects tops out at 63.7\%. The best-case performance on our maze navigation benchmark is only 15\% better than random guessing. Accuracy on geometric reasoning that requires depth and height reasoning max out at round 50\%. 

At the same time, detailed image understanding is also the type of competency that is critically needed in a truly multimodal scenario requiring physical awareness, localization, and grounded perception. For example, reliable object detection is required for  numerous safety monitoring and remote navigation use cases. These have been main drivers of the flagship advances in computer vision for specialized object localization. Low performance in these types of tasks means that replacement of traditional models like YOLO~\cite{redmon2016you} and Faster R-CNN~\cite{NIPS2015_14bfa6bb} with more recent LFMs is not yet realistic. Beyond important traditional object-recognition tasks, other rising scenarios motivate the critical need for accurate inferences about object recognition. For example, the configuration and spatial relationships of objects over time is essential in scenarios of rising importance centering on human-AI interaction on physical tasks \cite{bohus2024isthisit}.  
\section{Language Evaluation}
\label{sec:language}
In this section, we provide detailed analysis and results for the capabilities of instruction following, question answering for long context, information retrieval, as well as toxicity detection and safe language generation. 

To account for the impact of non-determinism (discussed in Section~\ref{sec:non_determinism}), all experiments for the smaller datasets (i.e. IFEval and Toxigen - generative) were repeated three times and we report the mean and corresponding standard error across the three repeated runs with temperature set to zero and top\_p = 0.95. For the larger datasets (i.e. Kitab, FlenQA, Toxigen - discriminative) we run the experiment two times to investigate the impact of non determinism at the overall dataset level and the subcategory level, and only observe minimal differences (of less than 0.5 percentage points). Therefore, for these datasets we henceforth report results on a single run.

\subsection{Instruction Following - IFEval}
\label{sec:ifeval}
\noindent \textbf{Motivation:} A critical skill for frontier models is the ability to follow instructions provided in the input prompt. Users provide increasingly complex instructions to LLMs in order to specify details about tasks they intend the model to perform, teach the model problem solving techniques and format the model's responses under specific requirements. Model training pipelines now often include a dedicated instruction tuning phase for specifically teaching models to follow complex instructions for real-world scenarios. 

Consequently, evaluating how well models follow such instructions is crucial when assessing overall model behaviour. While real instructions provided by users can be very varied and complex, a predominant category are instructions to control the format or style of the output. IFEval~\cite{zhou2023instruction} is a benchmark designed to evaluate a model's ability to follow instructions about the output style, structure and form. Recent model evaluations report $\sim$70-80\% accuracy on IFEval on average, showing headroom for further analysis and progress on challenging instruction categories.

\noindent \textbf{Benchmark Description:} The benchmark includes instruction based prompts for a category of ‘verifiable instructions’, which are defined as instructions amenable to objective verification of compliance. Examples of such instructions are: ‘write 450 to 500 words’, ‘your entire output should be in JSON output’, ‘include a title, and put it into two square brackets such as [[ title ]]'. The benchmark consists of nine broad instruction categories with 25 fine-grained types focusing on various output content and format constraint-based instructions. An input prompt can contain multiple instructions and can support fine-grained instruction level analysis.

Prior evaluations accompanying model releases, often report a single aggregate number by averaging the different metrics proposed in \cite{zhou2023instruction}, which often fail to reveal meaningful differences between models. Instead, the evaluation for IFEval in this report separately reports two understandable metrics at two levels of granularity: i) Overall Accuracy - reports dataset level accuracy of a model following 'all' instructions in an input prompt and percentage of instructions followed, both under strict criteria, across all categories and ii) Instruction Category Level accuracy - reports accuracy of following instructions under strict criteria per instruction category. 
\begin{figure}[t]
    \centering
{{\includegraphics[width=0.49\linewidth]{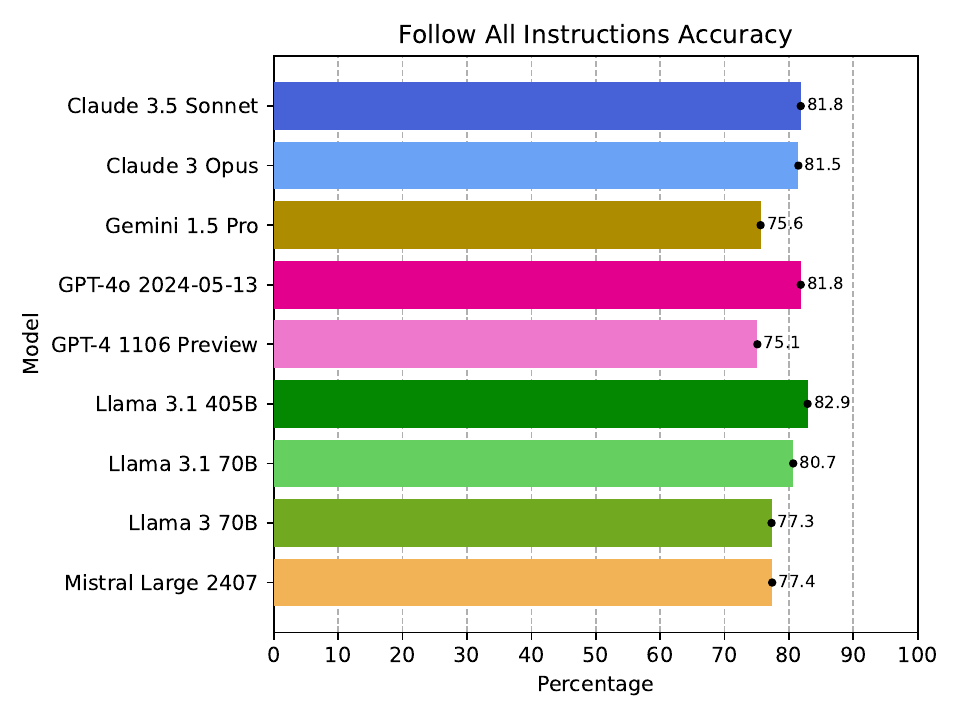} }}%
{{\includegraphics[width=0.49\linewidth]{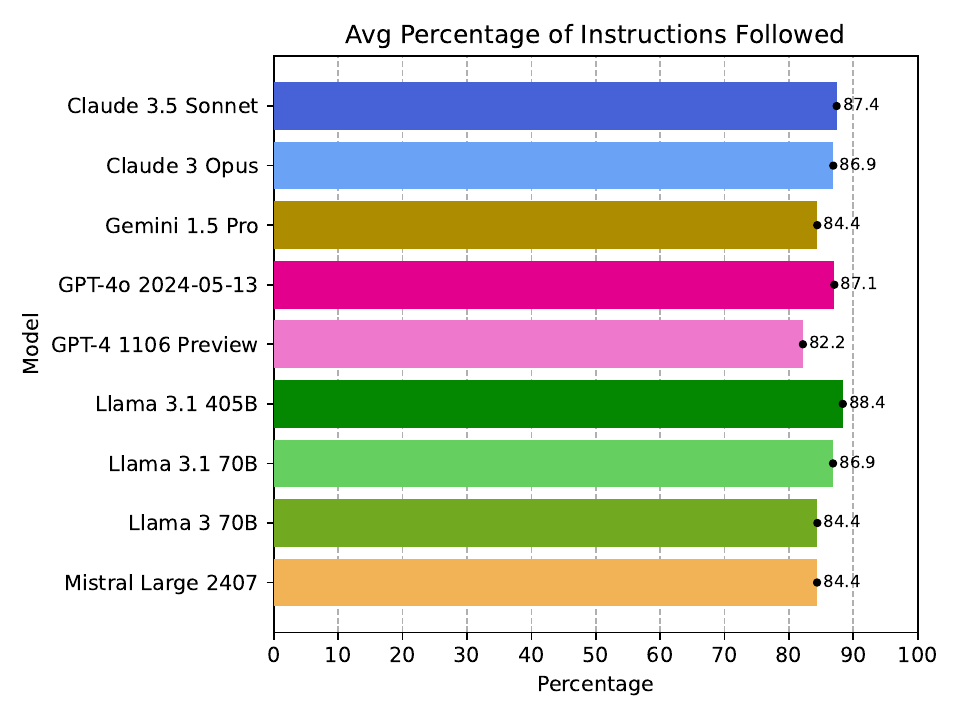} }}%
    \caption{Aggregate Prompt Level Instruction following accuracy and Percentage of Instructions followed under strict criteria reported across three different runs per model.}%
    \label{fig:ifeval_dataset_level_accuracy}%
\end{figure}

\noindent {\bf {Aggregate Results:}} Figure \ref{fig:ifeval_dataset_level_accuracy} presents instruction following accuracy and percentage of instructions followed for the entire dataset across the nine language models evaluated in this report. While all model perform significantly better in average number of instructions followed in a query, they all still lag in accuracy of strictly following all instructions included in the input. In the follow all instruction accuracy, performance of the strongest models in each family is similar in the range of 80-82\%, potentially indicating similar instruction following capabilities. Recent versions of large models in the Claude family (\ClaudeSonnet and \ClaudeOpus), GPT family (\GPTFourO) and Llama family (\LlamaThreeOneLarge and \LlamaThreeOne) all score above 80\% on the dataset showing strong instruction following abilities. Most recent GPT models (\GPTFourO) and Llama models (\LlamaThreeOne, \LlamaThreeOneLarge) show significant improvement from earlier versions (\GPTFourPrev and \LlamaThree) confirming increasing focus on instruction following during training. Interestingly, performance in the Claude family remains similar between earlier and recent models.

\begin{figure}[t!]
    \centering
    \includegraphics[width=0.75\linewidth]{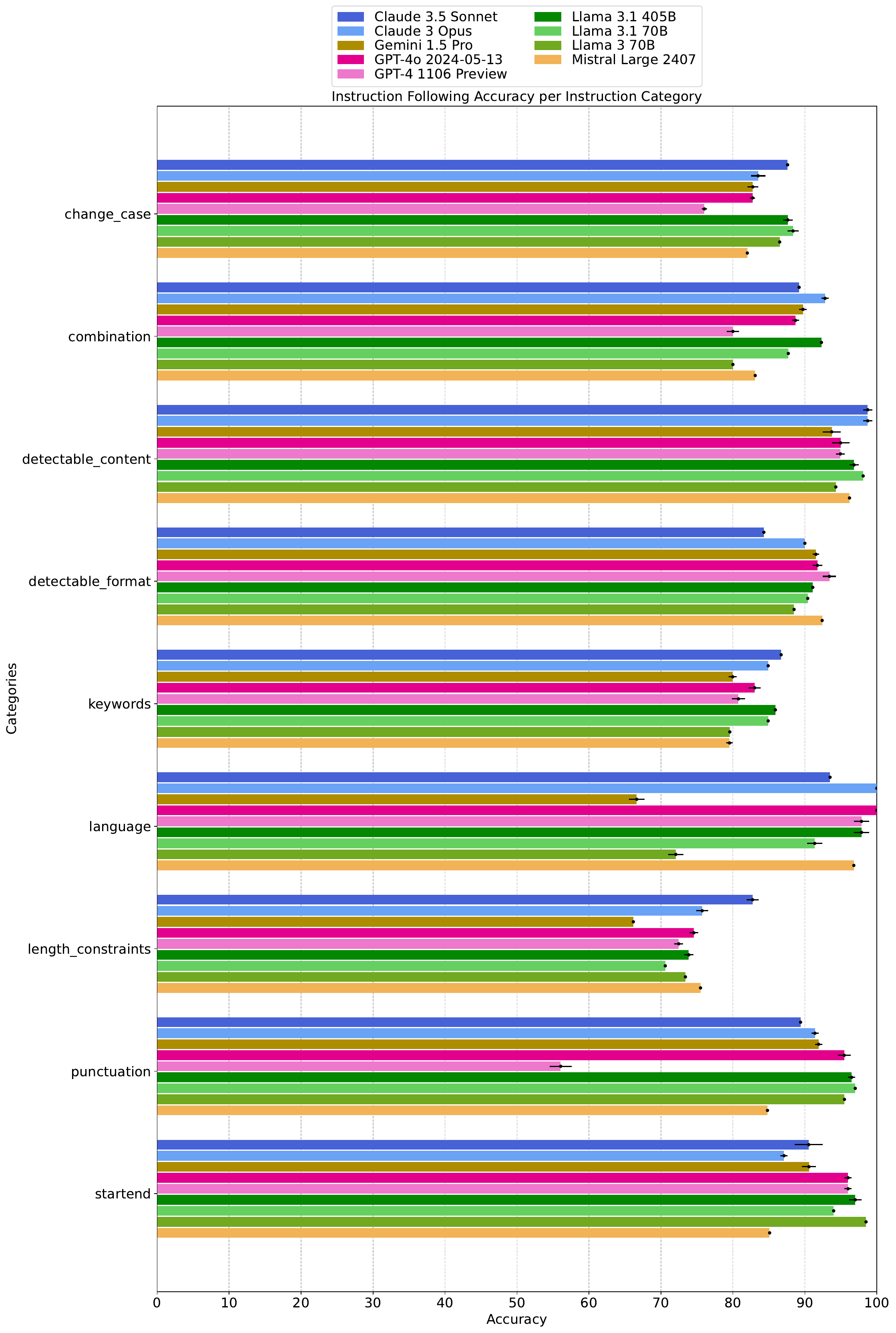}
    \caption{Instruction Category Level accuracy reported across three runs per model.}
    \label{fig:ifeval_instr_level_accuracy}
\end{figure}
\noindent {\bf Instruction Category Level Results:} Results of performance disaggregated by instruction category are presented in Figure \ref{fig:ifeval_instr_level_accuracy}. Instruction level breakdown reveals significantly varying performance for different instruction categories. Certain categories which include instructions regarding the content in the model's response (detectable content) or the language of the response (language) show models approaching 100\% accuracy indicating saturation with stronger, more powerful models. Other categories which instruct models with criteria to start or end with (startend), include specific punctuations (punctuation), format the response according to specific format structure (detectable format) have certain models performing close to $\sim$90\% showing significant improvements as model's improve in capabilities. This could potentially indicate that models are close to achieving near perfect performance in these categories soon, necessitating benchmarks on more challenging and complex instructions for further evaluation. Finally, categories like instructions regarding constraints on keywords or length of model responses have models following instructions less than 80\% of the time showing that such categories are still challenging for strong models.

In most categories, the gap between strong open-source and closed source models is narrow and comparable. \LlamaThreeOneLarge outperforms both \ClaudeSonnet and \GPTFourO on 4 categories and at least one of them in 8 categories. In instructions involving constraints on punctuations, changing case, criteria on start or end of response (startend), both \LlamaThreeOne and \LlamaThreeOneLarge outperform most closed source models. \GeminiPro has lower performance than other models in categories like language constraints and length constraints. We manually observed higher refusal in responses from \GeminiPro for language instructions. Finally, \MistralLargeTwo outperforms \GPTFourPrev in all instruction types, but under performs more recent models like \LlamaThreeOneLarge, \ClaudeSonnet and \GPTFourO.

When comparing models within a model family, while more recent models (\ClaudeSonnet, \GPTFourO) outperform older models (\ClaudeOpus, \GPTFourPrev) on most instruction categories as expected, there is a significant drop in performance in certain instruction categories. \ClaudeSonnet under performs \ClaudeOpus in instructions involving combining responses (combination), constraints on format structure (detectable format), constraints on response language (language) and constraints on usage of punctuations (punctuation) and \GPTFourO under performs \GPTFourPrev in the detectable format category. Such regression in performance indicates a backward compatibility issue with models in certain categories and could pose an issue for applications which rely on improving performance in them. This requires measuring compatibility with previous versions within a model family before adopting new models for applications. We explore this further with example level model comparison analysis in Section~\ref{sec:backward_compatibility}.
\vspace{-2mm}
\subsubsection*{Main takeaways}
\noindent\fbox{%
    \parbox{\textwidth}{%
        \begin{itemize}[leftmargin=*]
        \item Performance in instruction following varies amongst different categories of instructions. While strong models have near perfect accuracy for some types of instructions like constraints on language or content, they still struggle to consistently follow constraints on length or keywords.
        \item The gap between strong open-source models and closed-source models is small. \LlamaThreeOneLarge marginally outperforms strongest Claude and GPT at overall performance as well as in certain instruction types involving constraints on casing and punctuations. Within individual model families, more recent models do not consistently outperform older models on all categories. 
        \item Models are becoming increasingly good at following direct constraints on output format, which motivates the need for deeper evaluations with more complex, real-world instructions.
        \end{itemize}
    }%
}

\begin{figure}[t]
    \centering
        {{\includegraphics[width=0.7\textwidth]{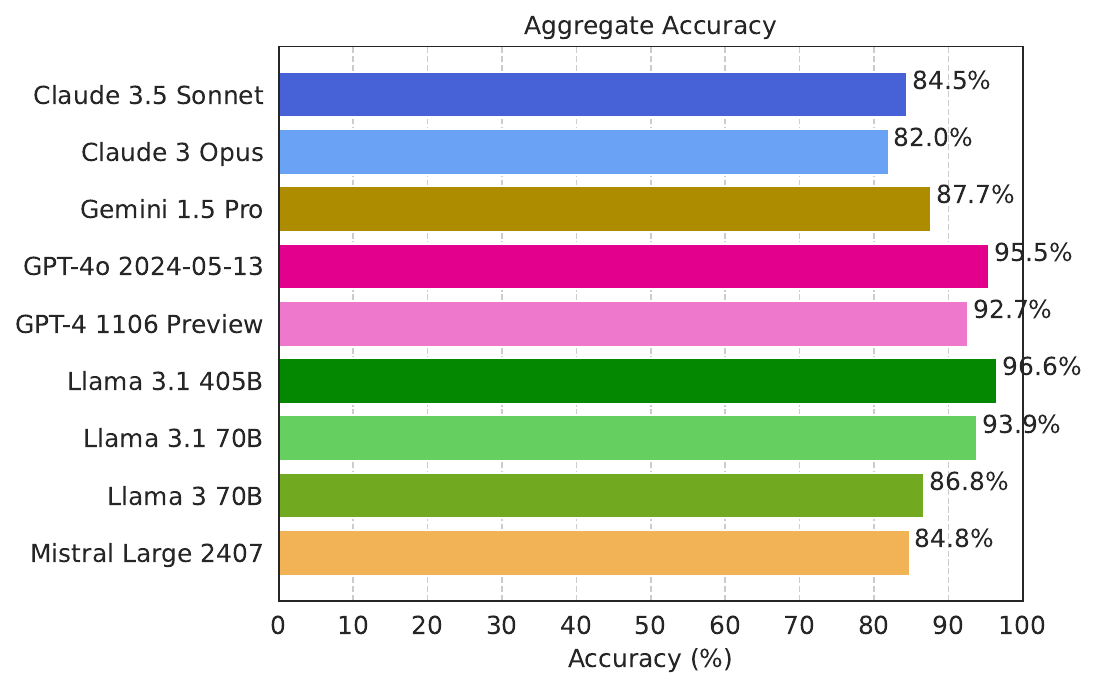}}}
    \caption{FlenQA - Overall accuracy of language models on all prompts combined.}
    \label{fig:flen_context}
\end{figure}
\subsection{Long Context - FlenQA}
\label{sec:flenqa}
\textbf{Motivation:} 
Despite significant recent improvements to LLMs and efforts to evaluate them in long context settings ~\cite{long1,long2}, their performance consistency across different input lengths remains poorly understood. FlenQA \cite{flenqa} aims to address this gap by isolating the effect of input length on language model performance.

Unlike “Needle-in-a-haystack” evaluations~\cite{reid2024gemini} that require retrieving a single fact from a long context (often as simple as Ctrl-F searches over the text), FlenQA involves complex multi-hop reasoning over contexts of various sizes. It requires the language model to retrieve and reason over two pieces of text in the context (two needles in the haystack). This makes it a helpful tool for understanding and improving the robustness and reasoning capabilities of LLMs in longer input lengths. 
\begin{figure}[t]
\centering
    \includegraphics[width=0.85\textwidth]{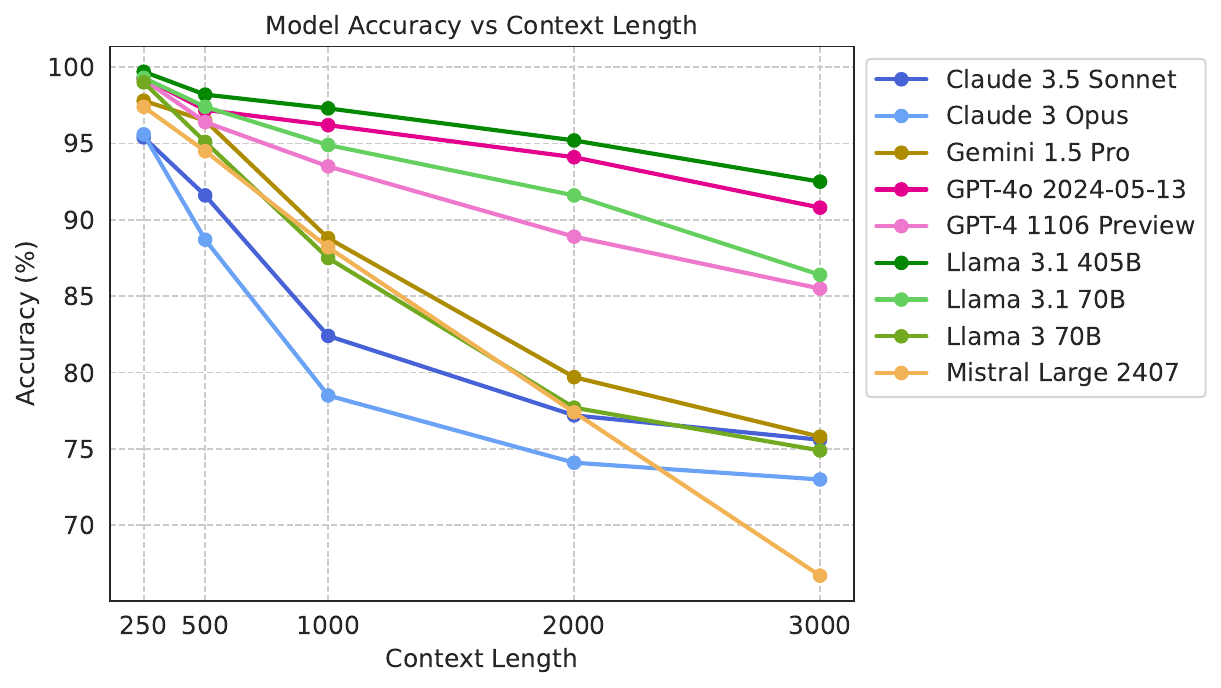}
        \label{fig:flenqa_overall}
    \caption{FlenQA - Effect of context length on language models' performance. All models suffer a degradation in accuracy with increased context length. Models in the GPT family and Llama 3.1 models that are more robust than others in this regard.}
\label{fig:flenqa_context}
\end{figure}
\begin{figure}[t]
    \centering
    \includegraphics[width=0.85\textwidth]{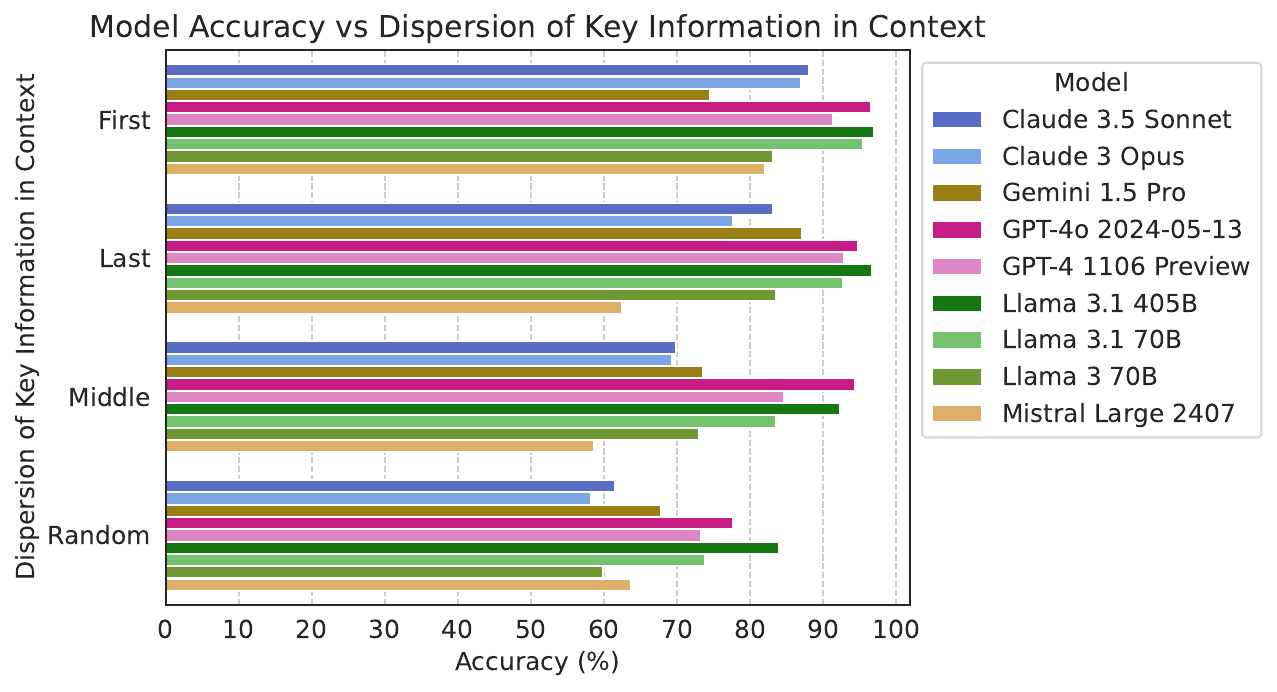}
    \caption{FlenQA - Accuray of language models in the longest context setting (3000 tokens), with key information placed at the beginning (\textit{first}), at the end (\textit{last}), in the middle (\textit{middle}), or dispersed at random locations (\textit{random}) of the context. In the first three dispersion settings, the two pieces of key information are adjacent, while in the \textit{random} setting they are presented separately. All models except \MistralLargeTwo are more challenged in the \textit{random} setting compared to other settings.}
    \label{fig:enter-label}
\end{figure}

\textbf{Benchmark Description:} FlenQA consists of 12K Questions/Assertions with True/False labels that aim to isolate the effect of input length on LLMs' performance using multiple versions of the same sample, extended with padding of different lengths, types and locations. Note that the goal is not to necessarily utilize the full context length of the models, but to study how their performance changes as the context length increases and the key information moves within the context.

Input lengths in FlenQA range from 250 to 3000 tokens. Each prompt is padded with paragraphs sampled from other instances of the same task, or paragraphs sampled from Book Corpus \cite{bookcorpus}, with key information presented at various locations in the context (at the beginning, at the end, in the middle, or at random locations).

\noindent \textbf{Aggregate Results:}
Despite recent works (e.g., Gemini 1.5 Pro~\cite{reid2024gemini}) showing improvement in “needle-in-a-haystack” experiments, our observation is that merely increasing the models' context size does not necessarily result in better complex reasoning capabilities in long-context tasks (see Figure \ref{fig:flenqa_context}). Significant performance degradation (up to 30\%) is observed with increasing context length for the following models: 
\ClaudeSonnet: 19.80\%, 
\ClaudeOpus: 22.60\%, 
\GeminiPro: 22.00\%, 
\GPTFourO: 8.50\%, 
\GPTFourPrev: 13.70\%, 
\LlamaThreeOneLarge: 7.20\%, 
\LlamaThreeOne: 12.90\%, 
\LlamaThree: 24.10\%, 
\MistralLargeTwo: 30.70\%. \LlamaThreeOneLarge is the most robust to context length increase, followed closely by \GPTFourO. The main failure modes across different models are the inability to identify the right pieces of information from the context and logical errors when reasoning across different pieces of text. 

To verify that the results are reproducible given the nondeterminism in some of the models, we repeated the experiments for three of the most non-deterministic models. We observed very small variation (standard error) in model accuracy between runs: 82.07 (0.05) for \ClaudeOpus, 92.71 (0.01) for \GPTFourPrev, and 87.73 (0.01) for \GeminiPro.

\noindent \textbf{Paragraph Location Results:} Prompts in the FlenQA benchmark represent four ways of placing the key informative paragraphs among the padding paragraphs. Key paragraphs are either presented at the beginning of the context (\textit{first}), at the end of the context (\textit{last}), in the middle of the context (\textit{middle}), or at random locations (\textit{random}). In the first three settings the key paragraphs are next to each other but in the \textit{random} setting they are separated. This can explain why most models have the hardest time solving the task in this setting. For example, when looking at the longest set of prompts (3000 tokens) \GPTFourO has an accuracy of 96\% when key information is presented at first, but 77\% in the \textit{random} setting. 

\subsubsection*{Main takeaways}
\noindent\fbox{%
    \parbox{\textwidth}{%
        \begin{itemize}[leftmargin=*]
        \item All LFMs studied in this work show performance degradation as the context length increases.
        \item\LlamaThreeOneLarge and  \GPTFourO  are more robust to increased context length compared to other models. Solely measuring aggregate accuracy does not reveal the models' different sensitivities to the context length, therefore, it is important to perform such in depth analyses.
        \item The results presented here are for reasoning over two needles in the haystack, i.e., two pieces of information in the context are necessary to answer the given question. Most models perform the worst when these two pieces of information are scattered randomly in the context as opposed to adjacent to each other. Our expectation is that in more complex scenarios the performance will degrade even more significantly.
        \end{itemize}
    }%
}
\subsection{Information Retrieval - Kitab}
\label{sec:kitab}
\textbf{Motivation:} Information Retrieval either from parametric knowledge or from input context is a task that applies to many search and knowledge seeking scenarios. At its core, the main question is whether it is possible to extract reliable factual knowledge from a model and whether it is possible to ground the model's answers in given context. Previous work has studied factuality by measuring model accuracy for questions whose output is expected to be single, often atomic facts~\cite{meng2022locating,LinHE22} or otherwise single facts that require multi-hop reasoning~\cite{Yang0ZBCSM18,trivedi2022musique}. However, given the generative nature of current models, a more compelling and contemporary scenario is the one where users form queries that expect a longer output with a list of items that satisfy the criteria behind what they are looking for (e.g., \emph{``a list of ice cream shops in San Diego''}). It turns out that ensuring \emph{factuality and grounding} for such longer generational tasks is challenging~\cite{AbdinGCLYPNN24} for state-of-the-art models, despite long generation being one of the core promises of LFMs.   

\textbf{Benchmark Description:} Kitab~\cite{AbdinGCLYPNN24} is a challenging dataset and a dynamic data collection approach for testing abilities of Large Language Models (LLMs) in answering information retrieval queries with constraint filters. A filtering query with constraints can be of the form \emph{``List all books written by Toni Morrison that were published between 1970-1980''}. Kitab consists of book-related data across more than 600 authors and 13,000 queries with varying number of constraints and complexity. In each query in the dataset, the first constraint is always fixed to an author and the following can vary among the following types of book constraints to test for different constraint satisfaction capabilities: lexical, named entity, temporal. If the model fails to satisfy the constraints, this can lead to information fabrication and hallucinations (i.e., book titles that do not exist), factual mistakes (i.e., book titles that are not from the author or that do not satisfy the constraints given by the user), grounding failures (i.e., inability to extract and parse information presented in context).

There are three experimental conditions:
\begin{itemize}[leftmargin=*]
    \item \textsc{no-context}: Testing factuality and constraint satisfaction abilities of the model based on its own parametric knowledge.
    \item \textsc{with-context}: Testing factuality and constraint satisfaction abilities of the model when perfect context is provided, i.e. grounding in a RAG-style setting.
    \item \textsc{self-context}: Similar to above, but the context is generated from the model itself as part of its own chain of thought (i.e. generate all books first, and then apply the query constraints).
\end{itemize}

The dataset uses the following metrics:
\begin{itemize}[leftmargin=*]
\item {\bfseries Information irrelevance}: The percentage of books in a model output that are not from the author or do not exist. The two cases cannot be distinguished because it is not possible to fully verify whether the book title exists amongst all books ever published. The lower the score, the better. 
\item {\bfseries Satisfaction rate}: The percentage of books in a model output that satisfy all given book constraints (except the authorship, which is captured in information irrelevance). The definition is similar to precision in classic retrieval tasks. The higher the score, the better.
\item {\bfseries Unsatisfaction rate}: The percentage of books in a model output that do not satisfy at least one of the book constraints (except the authorship, which is captured in information irrelevance). The lower the score, the better.
\item {\bfseries Completeness}: The percentage of books from the ground truth list of books that satisfy all query constraints and that are also mentioned in the model output.  The definition is similar to recall in classic retrieval tasks. The higher the score, the better.
\item {\bfseries All correctness}: The percentage of queries for which the list of books mentioned by the model fully matches the ground truth list of books. The higher the score, the better.
\end{itemize}
\textbf{Aggregate Results (context availability and number of constraints):} Tables~\ref{tab:kitab_overall_table_single} and~\ref{tab:kitab_overall_table_double} summarize results for queries with one and two book constraints for the different experimental conditions respectively. Overall, when the model uses only its own knowledge, constraint satisfaction rate is less or equal to 55\% for all models, information irrelevance is higher than 20\%, and completeness is lower than 25\%. All correctness is a strict metric that requires the ground truth and the model output to match, and ranges between only 1\% - 9\%. As measures of factuality, these results show that constrained information retrieval for longer output remains challenging even for most capable models. There is however some visible variation across models, with the top three models in terms of constraint satisfaction rate being \LlamaThreeOneLarge, \GPTFourO, and \ClaudeSonnet. The same three models also have the highest completeness.

\setlength\tabcolsep{3 pt}
\begin{table}[t]
\centering
\begin{tabular}{|l|cllcllrllcllcll}
\hline
\multirow{2}{*}{} & \multicolumn{3}{c|}{\multirow{2}{*}{\textbf{\begin{tabular}[c]{@{}c@{}}Irrelevant \\information\xspace$\downarrow$\\ \end{tabular}}}} & \multicolumn{6}{c|}{\textbf{\begin{tabular}[c]{@{}c@{}}Relevant information\\ (Books from the author) \\ \end{tabular}}} & \multicolumn{3}{c|}{\multirow{2}{*}{\textbf{Completeness\xspace$\uparrow$}}} & \multicolumn{3}{c|}{\multirow{2}{*}{\textbf{All Correct\xspace$\uparrow$}}} \\ 
              & \multicolumn{3}{c|}{}                                                    & \multicolumn{3}{c|}{\textbf{Satisfied\xspace$\uparrow$}}                    & \multicolumn{3}{c|}{\textbf{Unsatisfied\xspace$\downarrow$}}                & \multicolumn{3}{c|}{}                                       & \multicolumn{3}{c|}{}                                      \\ \hline

\ClaudeSonnet   & \multicolumn{3}{c|}{22.4 $|$ \textbf{20.9} $|$ 0.1}   & \multicolumn{3}{c|}{\textbf{55.3} $|$ 56.0 $|$ 87.4}   & \multicolumn{3}{c|}{22.2 $|$ 23.2 $|$ 12.5}   & \multicolumn{3}{c|}{20.6 $|$ \textbf{24.7} $|$ 68.9}   & \multicolumn{3}{c|}{\textbf{3.6} $|$ \textbf{4.9} $|$ 38.0}   
\\
\ClaudeOpus   & \multicolumn{3}{c|}{\textbf{20.5} $|$ 23.0 $|$ \textbf{0.0}}   & \multicolumn{3}{c|}{50.5 $|$ 48.5 $|$ 84.4}   & \multicolumn{3}{c|}{29.0 $|$ 28.4 $|$ 15.6}   & \multicolumn{3}{c|}{18.7 $|$ 23.6 $|$ 65.1}   & \multicolumn{3}{c|}{2.7 $|$ 4.0 $|$ 30.7}   
\\ \hline
\GeminiPro   & \multicolumn{3}{c|}{29.7 $|$ 37.3 $|$ 0.2}   & \multicolumn{3}{c|}{41.3 $|$ 38.0 $|$ 74.5}   & \multicolumn{3}{c|}{29.0 $|$ 24.7 $|$ 25.3}   & \multicolumn{3}{c|}{  9.8 $|$ 17.6 $|$ 61.3}   & \multicolumn{3}{c|}{1.2 $|$ 2.2 $|$ 30.4}   
\\ \hline
\GPTFourO   & \multicolumn{3}{c|}{20.6 $|$ 25.8 $|$ \textbf{0.0}}   & \multicolumn{3}{c|}{53.7 $|$ 49.3 $|$ 84.7}   & \multicolumn{3}{c|}{25.7 $|$ 24.9 $|$ 15.2}   & \multicolumn{3}{c|}{20.3 $|$ 24.6 $|$ \textbf{69.2}}   & \multicolumn{3}{c|}{\textbf{3.6} $|$ 4.3 $|$ 31.4}   
\\
\GPTFourPrev   & \multicolumn{3}{c|}{24.5 $|$ 32.1 $|$ 1.8}   & \multicolumn{3}{c|}{47.0 $|$ 43.4 $|$ 75.2}   & \multicolumn{3}{c|}{28.4 $|$ 24.5 $|$ 23.0}   & \multicolumn{3}{c|}{\textbf{23.3} $|$ 25.6 $|$ \textbf{69.2}}   & \multicolumn{3}{c|}{2.1 $|$ 3.4 $|$ 26.6}   
\\ \hline
\LlamaThreeOneLarge   & \multicolumn{3}{c|}{24.5 $|$ 21.5 $|$ 0.3}   & \multicolumn{3}{c|}{54.9 $|$ \textbf{56.4} $|$ \textbf{89.1}}   & \multicolumn{3}{c|}{\textbf{20.5} $|$ 22.1 $|$ \textbf{10.6}}   & \multicolumn{3}{c|}{16.8 $|$ 21.2 $|$ 67.8}   & \multicolumn{3}{c|}{3.3 $|$ 4.0 $|$ \textbf{39.6}}   
\\
\LlamaThreeOne  & \multicolumn{3}{c|}{31.2 $|$ 25.7 $|$ 0.3}   & \multicolumn{3}{c|}{44.2 $|$ 48.0 $|$ 85.9}   & \multicolumn{3}{c|}{24.5 $|$ 26.3 $|$ 13.8}   & \multicolumn{3}{c|}{16.0 $|$ 16.1 $|$ 61.6}   & \multicolumn{3}{c|}{2.6 $|$ 2.8 $|$ 30.5}   
\\
\LlamaThreeOne   & \multicolumn{3}{c|}{41.8 $|$ 43.1 $|$ 0.6}   & \multicolumn{3}{c|}{37.4 $|$ 37.1 $|$ 76.9}   & \multicolumn{3}{c|}{20.8 $|$ \textbf{19.7} $|$ 22.4}   & \multicolumn{3}{c|}{15.0 $|$ 16.8 $|$ 62.3}   & \multicolumn{3}{c|}{1.8 $|$ 2.3 $|$ 23.9}   
\\ \hline
\MistralLargeTwo   & \multicolumn{3}{c|}{36.8 $|$ 39.5 $|$ 0.2}   & \multicolumn{3}{c|}{36.3 $|$ 34.9 $|$ 76.9}   & \multicolumn{3}{c|}{26.9 $|$ 25.5 $|$ 22.9}   & \multicolumn{3}{c|}{17.6 $|$ 19.5 $|$ 64.5}   & \multicolumn{3}{c|}{2.2 $|$ 2.6 $|$ 26.4}   
\\
\hline
\end{tabular}
\caption{Aggregated model performance on Kitab for \textsc{no-context} $|$ \textsc{self-context} $|$ \textsc{with-context}. Queries with one book constraint.}
\label{tab:kitab_overall_table_single}
\end{table}
\setlength\tabcolsep{3 pt}
\begin{table}[t]
\centering
\begin{tabular}{|l|cllcllrllcllcll}
\hline
\multirow{2}{*}{} & \multicolumn{3}{c|}{\multirow{2}{*}{\textbf{\begin{tabular}[c]{@{}c@{}}Irrelevant \\information\xspace$\downarrow$\\ \end{tabular}}}} & \multicolumn{6}{c|}{\textbf{\begin{tabular}[c]{@{}c@{}}Relevant information\\ (Books from the author) \\ \end{tabular}}} & \multicolumn{3}{c|}{\multirow{2}{*}{\textbf{Completeness\xspace$\uparrow$}}} & \multicolumn{3}{c|}{\multirow{2}{*}{\textbf{All Correct\xspace$\uparrow$}}} \\ 
              & \multicolumn{3}{c|}{}                                                    & \multicolumn{3}{c|}{\textbf{Satisfied\xspace$\uparrow$}}                    & \multicolumn{3}{c|}{\textbf{Unsatisfied\xspace$\downarrow$}}                & \multicolumn{3}{c|}{}                                       & \multicolumn{3}{c|}{}                                      \\ \hline
\ClaudeSonnet   & \multicolumn{3}{c|}{26.0 $|$ 0.1}   & \multicolumn{3}{c|}{41.5 $|$ \textbf{73.1}}   & \multicolumn{3}{c|}{32.5 $|$ \textbf{26.9}}   & \multicolumn{3}{c|}{14.5 $|$ 53.1}   & \multicolumn{3}{c|}{7.8 $|$ \textbf{32.5}}   
\\
\ClaudeOpus   & \multicolumn{3}{c|}{28.3 $|$ \textbf{0.0}}   & \multicolumn{3}{c|}{33.5 $|$ 64.7}   & \multicolumn{3}{c|}{38.2 $|$ 35.2}   & \multicolumn{3}{c|}{17.7 $|$ \textbf{58.8}}   & \multicolumn{3}{c|}{7.1 $|$ 29.9}   
\\ \hline
\GeminiPro   & \multicolumn{3}{c|}{30.2 $|$ 0.3}   & \multicolumn{3}{c|}{29.4 $|$ 55.9}   & \multicolumn{3}{c|}{40.4 $|$ 43.8}   & \multicolumn{3}{c|}{4.8 $|$ 51.0}   & \multicolumn{3}{c|}{1.6 $|$ 24.4}   
\\ \hline
\GPTFourO   & \multicolumn{3}{c|}{\textbf{25.9} $|$ \textbf{0.0}}   & \multicolumn{3}{c|}{40.7 $|$ 63.0}   & \multicolumn{3}{c|}{33.3 $|$ 36.9}   & \multicolumn{3}{c|}{17.2 $|$ 52.0}   & \multicolumn{3}{c|}{\textbf{8.5} $|$ 27.0}   
\\
\GPTFourPrev   & \multicolumn{3}{c|}{28.4 $|$ 1.0}   & \multicolumn{3}{c|}{30.5 $|$ 51.6}   & \multicolumn{3}{c|}{41.1 $|$ 47.3}   & \multicolumn{3}{c|}{\textbf{17.9} $|$ 54.1}   & \multicolumn{3}{c|}{5.9 $|$ 18.3}   
\\ \hline
\LlamaThreeOneLarge   & \multicolumn{3}{c|}{30.0 $|$ 1.2}   & \multicolumn{3}{c|}{\textbf{43.5} $|$ 66.1}   & \multicolumn{3}{c|}{\textbf{26.5} $|$ 32.7}   & \multicolumn{3}{c|}{15.3 $|$ 53.7}   & \multicolumn{3}{c|}{8.4 $|$ 30.8}   
\\
\LlamaThreeOne   & \multicolumn{3}{c|}{35.0 $|$ 0.5}   & \multicolumn{3}{c|}{34.4 $|$ 65.6}   & \multicolumn{3}{c|}{30.6 $|$ 33.9}   & \multicolumn{3}{c|}{12.3 $|$ 48.5}   & \multicolumn{3}{c|}{6.7 $|$ 27.7}   
\\
\LlamaThree   & \multicolumn{3}{c|}{51.3 $|$ 1.1}   & \multicolumn{3}{c|}{24.8 $|$ 54.2}   & \multicolumn{3}{c|}{23.9 $|$ 44.7}   & \multicolumn{3}{c|}{12.3 $|$ 49.0}   & \multicolumn{3}{c|}{4.5 $|$ 20.9}   
\\ \hline
\MistralLargeTwo & \multicolumn{3}{c|}{40.2 $|$ 0.3}   & \multicolumn{3}{c|}{23.4 $|$ 52.8}   & \multicolumn{3}{c|}{36.5 $|$ 46.9}   & \multicolumn{3}{c|}{12.6 $|$ 50.0}   & \multicolumn{3}{c|}{4.2 $|$ 19.6}   
\\

\hline  

\end{tabular}
\caption{Aggregated model performance on Kitab for \textsc{no-context} $|$ \textsc{with-context}.\\ Queries with two book constraints.}
\label{tab:kitab_overall_table_double}
\end{table}

In the \textsc{with-context} experimental condition, where the model is given all books from the author in context and there is only one book constraint, constraint satisfaction improves significantly, as high as 89\% for \LlamaThreeOneLarge. In fact, we observe that improvements over \GPTFourPrev (as one of the earliest models) on this condition are larger and progress here looks faster in general. As we show later however, for some constraint types, models still struggle even for this experimental condition. Also note that, since the input context is as good as it can be based on the dataset's ground truth, these results should be interpreted as upper bounds to how well these models can perform when tied to almost perfect RAG components. Other mistakes in a RAG system may also negatively impact these results.

Finally, self-retrieval of context in the \textsc{self-context} experimental condition negatively impacts constraint satisfaction of all models except \ClaudeSonnet and the models in the Llama family. The negative impact on constraint satisfaction for the other is often caused by a higher irrelevance rate, which means that during the first step of CoT, the models extract book titles that do not belong to the author on the first place (potentially hallucinated/fabricated titles). It is interesting to see that this is not the case for \ClaudeSonnet and the Llama family. All models have a higher completeness rate in the \textsc{self-context} than in \textsc{no-context}, potentially rooted in the additional help provided by first extracting all books from the author prior to running constraints.

As shown in Table~\ref{tab:kitab_overall_table_double}, for more complex queries that have two book constraints in addition to the author constraint, the best constraint satisfaction rate for the whole query is 44\% for \LlamaThreeOneLarge, followed by \ClaudeSonnet and \GPTFourO with 41\% for the \textsc{no-context} condition. This is 10\%-15\% lower than for the simpler case of having only one constraint. When context is provided, \ClaudeSonnet leads with a 73\% satisfaction rate, followed by \LlamaThreeOneLarge and \LlamaThreeOne with 66\%.

\begin{figure}[t]
    \centering
{{\includegraphics[width=6.8cm]{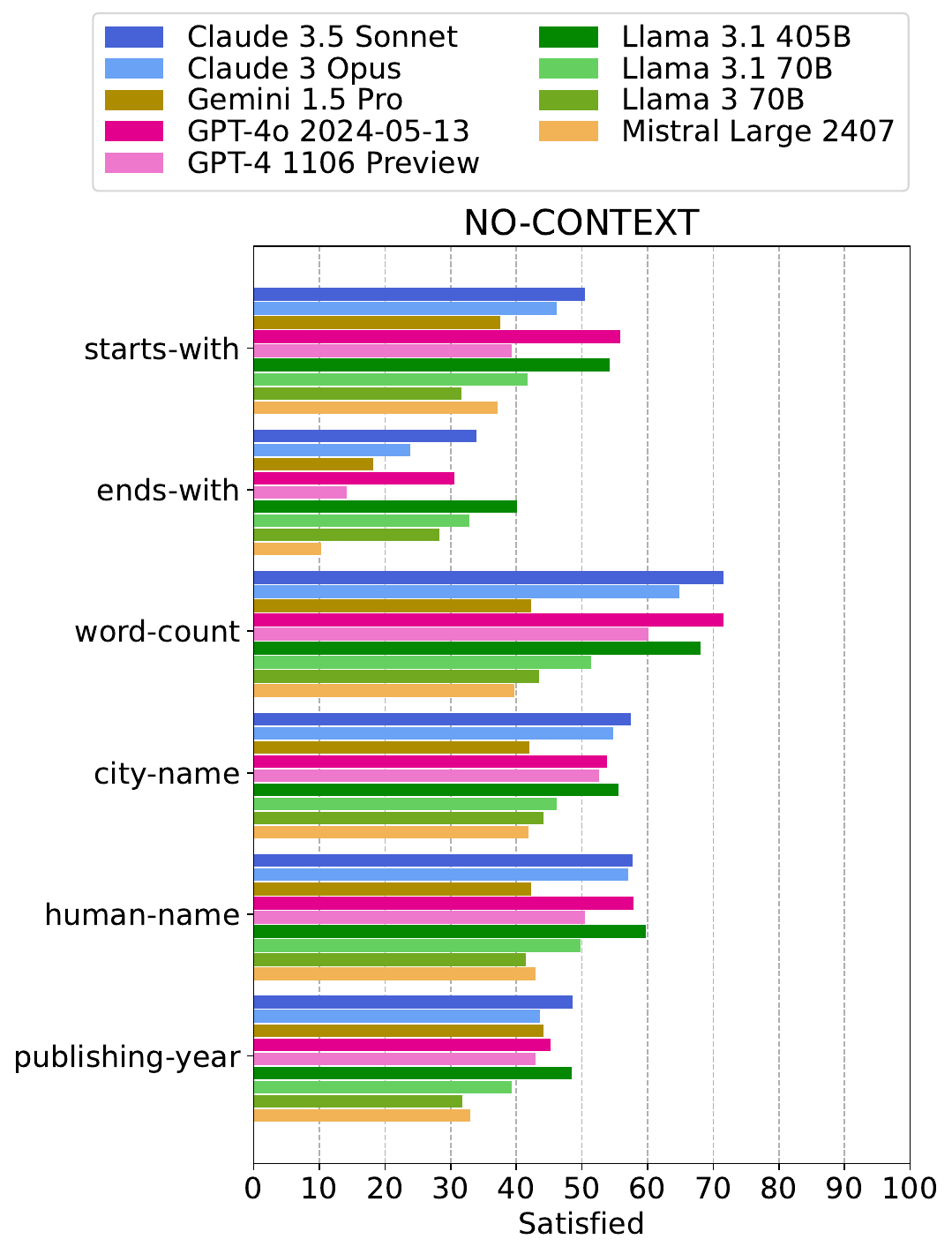} }}%
{{\includegraphics[width=6.8cm]{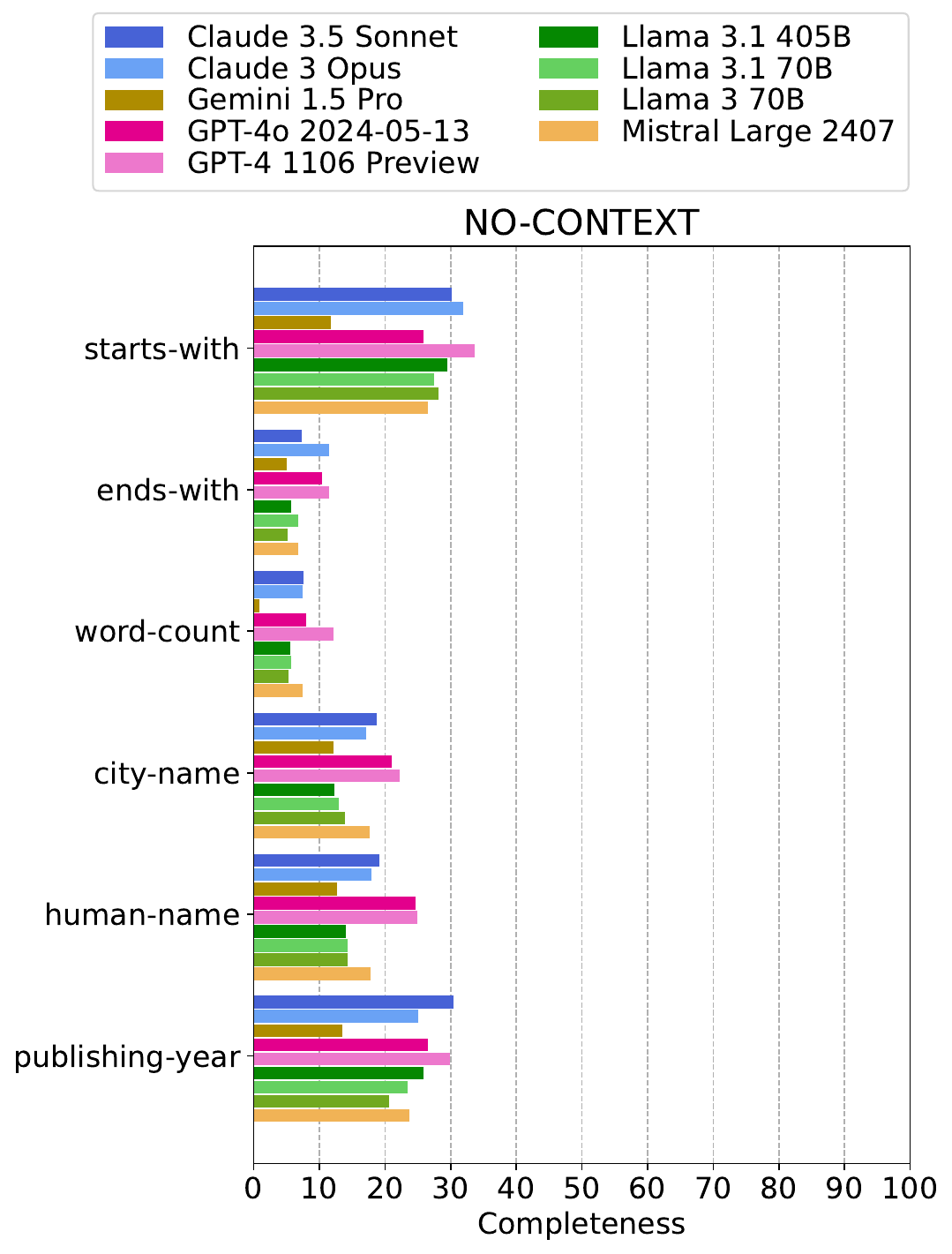} }}%
    \caption{Satisfaction rate and completeness for the \textsc{no-context} condition in Kitab. Queries with one book constraint.}%
    \label{fig:kitab_no_context_by_constraint}%
\end{figure}
\begin{figure}[t!]
    \centering
{{\includegraphics[width=6.8cm]{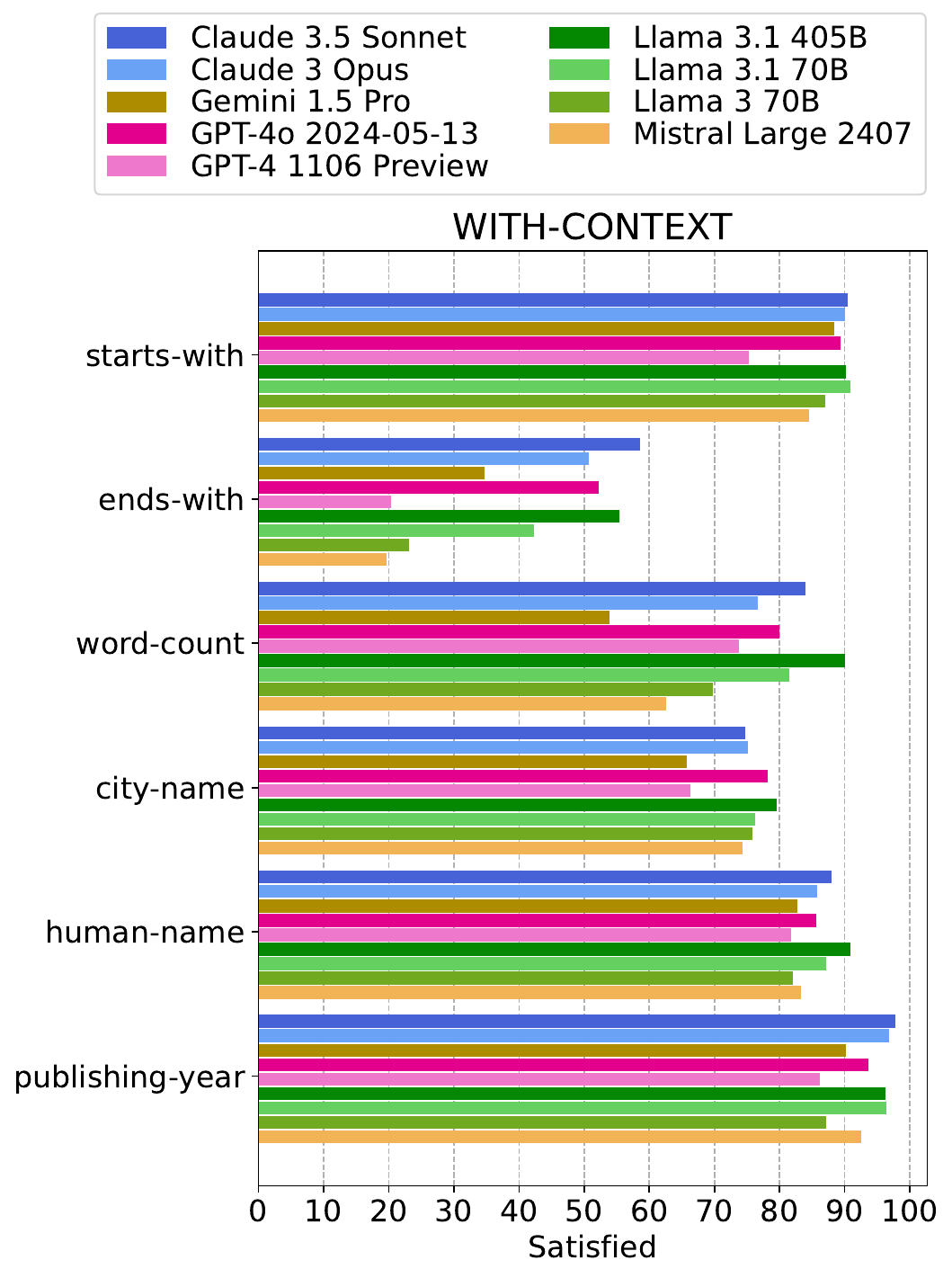} }}%
{{\includegraphics[width=6.8cm]{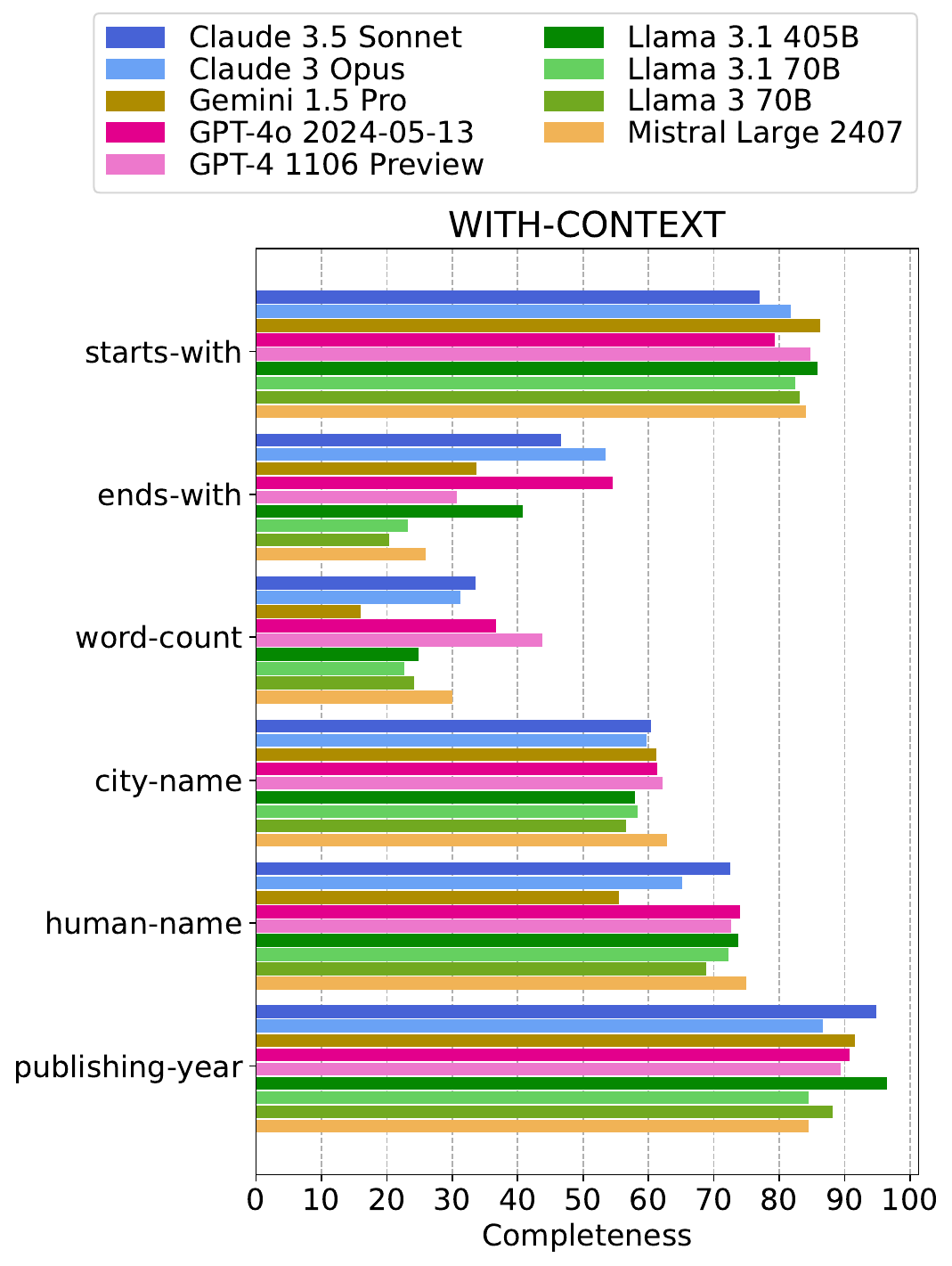} }}%
    \caption{Satisfaction rate and completeness for the \textsc{with-context} condition  in Kitab. Queries with one book constraint.}%
    \label{fig:kitab_with_context_by_constraint}%
\end{figure}
\textbf{Constraint Type Results:} Next, we run an error analysis to understand what types of constraints drive important failure modes. Figures~\ref{fig:kitab_no_context_by_constraint} and~\ref{fig:kitab_with_context_by_constraint} show constraint satisfaction rates and completeness for both the \textsc{no-context} and \textsc{with-context} conditions, for queries with a single book constraint. First, there are observations that are common amongst most models. For example, for both conditions queries with ends-with constraints (title ends with a given letter) are more difficult for all models when compared to other constraints, which is presumably rooted in the fact that satisfying ends-with queries requires more planning ahead. In fact, here the constraint can only be verified after the whole title generation has ended. Word-count queries are an interesting case for which satisfaction rates are amongst the highest across constraint types, but completeness rates are amongst the lowest. However, it is worth noting that when we look at trends to what is driving improvements in this dataset during the newest model releases (e.g. \GPTFourPrev to \GPTFourO, or \LlamaThree to \LlamaThreeOne), word-count and ends-with constraints are the ones that show most improvements, followed by publishing-year. For the \textsc{with-context} condition, it is important to note that although satisfaction rate for entity constraints (includes human-name or city-name) is between 75\% - 90\%, completeness is lower than 75\%. This shows that despite advances in entity recognition as a classical task in NLP, when it comes to leveraging this skill in slightly more complex queries, recall in entity detection is far from saturated.

\begin{figure}[t]
    \centering
{{\includegraphics[width=6.8cm]{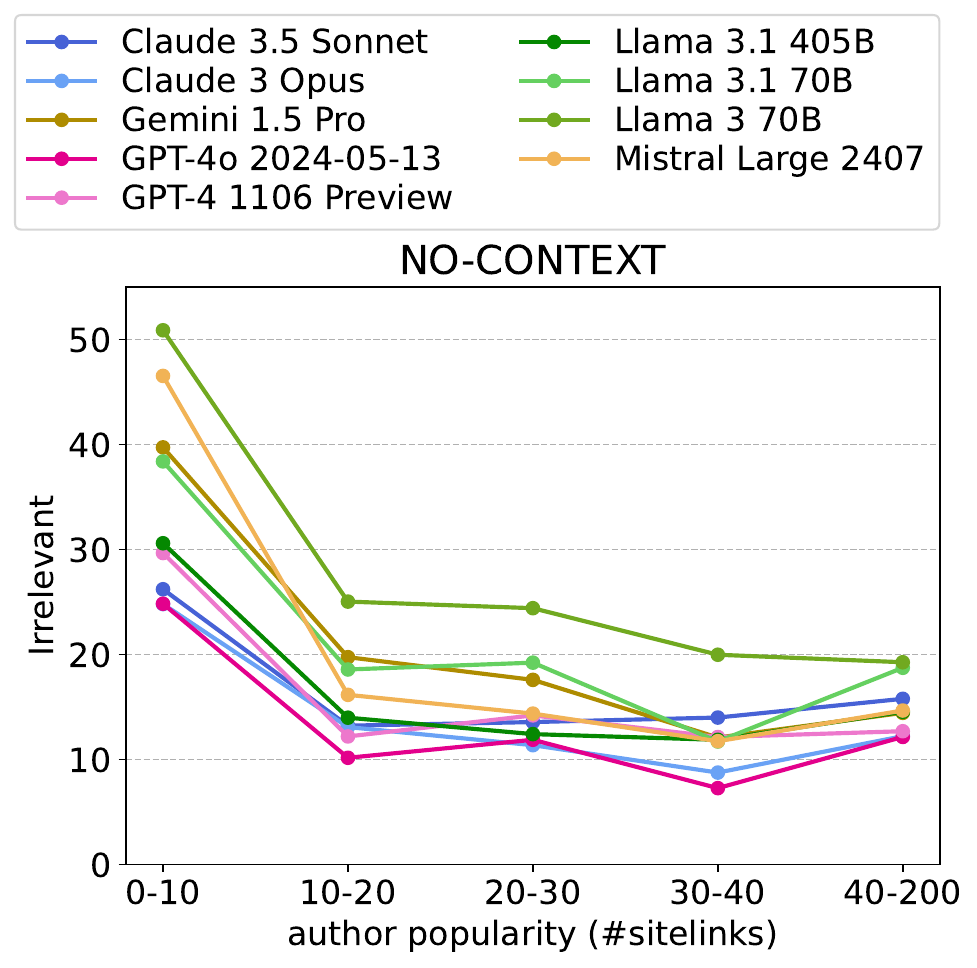} }}%
{{\includegraphics[width=6.8cm]{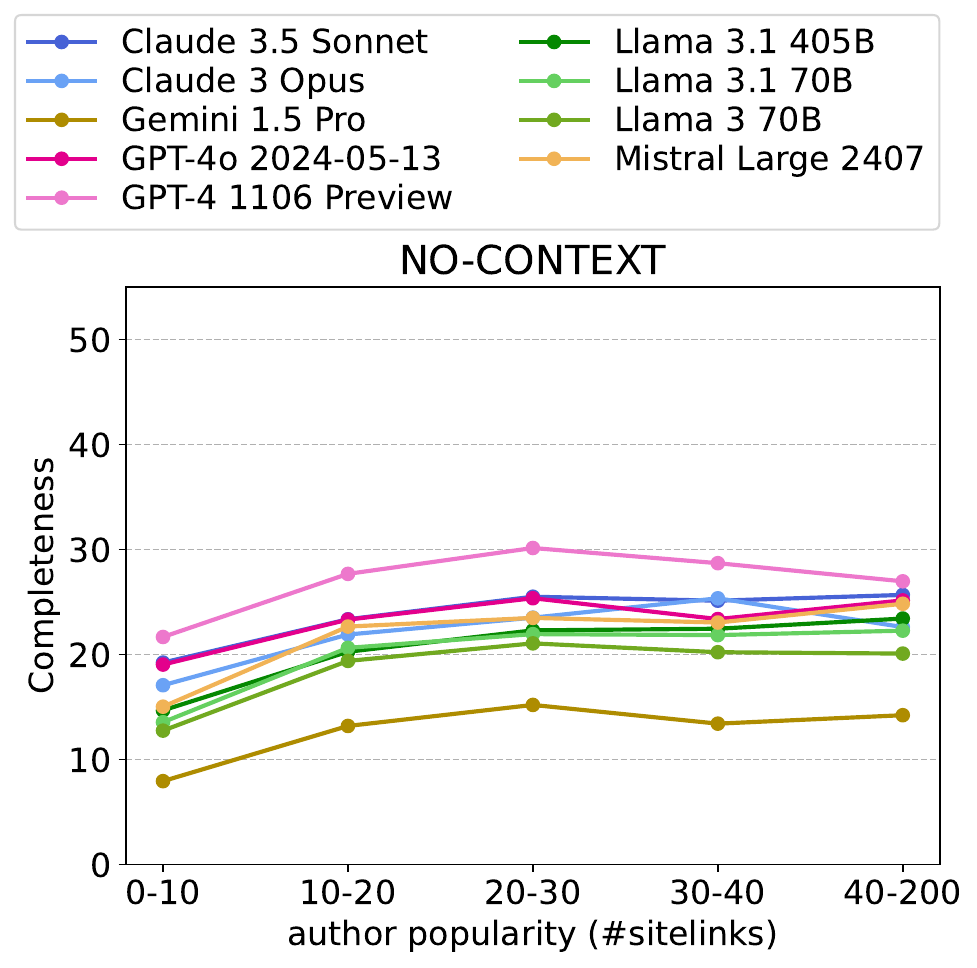} }}%
    \caption{Irrelevance and completeness for the \textsc{no-context} condition in Kitab. Lower irrelevance rates are better. Higher completeness rates are better. Queries with one book constraint.}%
    \label{fig:kitab_no_context_by_popularity}%
\end{figure}

\textbf{Popularity Results:} When it comes to understanding how much information a model can store and retrieve effectively, information frequency in training data is a dimension that can directly impact model accuracy and factuality. Since we do not have access to the training data of any of the models, we use author popularity as a common denominator, measured by the number of sitelinks in the author's Wikipedia profile. Previous work ~\cite{YuksekgonulCJGN24,AbdinGCLYPNN24} has used a similar proxy indicator, under the assumption that most models have been trained on web data, including Wikipedia. Figure~\ref{fig:kitab_no_context_by_popularity} disaggregates irrelevance and completeness for different popularity bins, as two metrics that are indirectly related to how much the model knows about the author, as reflected in its answers. For example, if the model often outputs irrelevant titles that do not exist or are not written by the author, this indirectly shows that the model is not able to map that those books do not belong to the author. 

As Adbin et al.~\cite{AbdinGCLYPNN24} observed for the GPT family of models, a low number of sitelinks is associated with higher irrelevance and lower completeness. However, the transition is somewhat sharp  in the [0, 20] interval and then flattens. Our results show that the same trend is valid for all other models in this analysis. Interestingly, we also see that for the regime of lowest popularity (i.e., 0-10 sitelinks) \GPTFourO and the Claude family have the lowest irrelevance rates, which explains why they are amongst the best models in this task. \LlamaThreeOneLarge instead has higher irrelevance rates (by 5\%) than \GPTFourO for this regime, but then compensates this with competitive constraint satisfaction capabilities, as shown in Figure~\ref{fig:kitab_no_context_by_constraint} and Table~\ref{tab:kitab_overall_table_single}. 

From a methodological point of view, these results show the importance of testing factuality for questions related to less frequent and popular entities. Even though most models seem to have comparable performance for authors with high popularity, the lower popularity regime uncovers important differences between models. Note that the authors in this category have still done important work and have written at least five distinct books available in the OpenLibrary database. Thus, even though the setting relates to tail information in terms of internet knowledge, in terms of relevance to a large user audience these queries  are still of interest. Early work in web search and information retrieval~\cite{downey2007heads,bernstein2012direct} shows that tail queries constitute a considerable amount of search traffic. The discussion is also important from a geographical fairness perspective, as the factuality of AI-generated information about geographical locations is shown to vary dramatically in recent measurements~\cite{moayeri2024worldbench}.

\subsubsection*{Main takeaways}
\noindent\fbox{%
    \parbox{\textwidth}{%
        \begin{itemize}[leftmargin=*]
            \item State-of-the-art models continue to struggle with eliciting factual information from their parametric knowledge for generating long-form output and with following filtering constraints, with constraint satisfaction being less than 55\% across all tested models.
            \item \LlamaThreeOneLarge, \GPTFourO, and \ClaudeSonnet are the best performing models in this task across different conditions. \GPTFourO and \ClaudeSonnet in particular have significantly lower information irrelevance rates (associated with better factuality) than other models. \LlamaThreeOneLarge has better constraint satisfaction rates (associated with better constrained text generation and grounding).
            \item Error analysis across different constraint types shows that ends-with and word-count queries are the most difficult ones across models, and that they are also the leading source of improvements during the most recent model releases within the same model family. 
            \item A similar analysis for different author popularities shows that queries with lower author popularity are the most difficult ones, universally for all models. However, most factuality benchmarks still evaluate question answering for popular entities and cannot observe differences across models for less frequent information.
        \end{itemize}
    }%
}
\subsection{Toxicity Detection and Safe Language Generation - Toxigen}
\label{sec:toxigen}
\noindent \textbf{Motivation:} Measuring harmful language generation and safety in general is a complex problem with various aspects and dimensions. In this work we use Toxigen dataset~\cite{toxigen} for our measurements. It has a balanced number of neutral and toxic statements about various identity groups of people (mostly focusing on minorities), and includes the data to support both discriminative evaluation (LLM used as classifier for toxicity detection) and generative evaluation (LLM used as a text generator). 
Toxicity detection is important for content moderation and safety filtering, while generative setting is important for assessing language generation safety in response to various input stimuli.

\noindent \textbf{Benchmark Description:} Toxigen is a large-scale dataset consisting of toxic and benign statements about 13 groups of people with a focus on implicit hate speech about minority groups that does not contain slurs or profanity. The dataset is designed to be balanced, i.e., for each identity group there are an equal number of toxic and neutral samples. This is important because the neutral samples can be used to measure erasure across different models where the identity mention of specific groups of people is treated as a signal of toxicity or hate and removed by the given LLM.

\begin{figure}[t]
    \centering
    \includegraphics[width=0.43\linewidth]{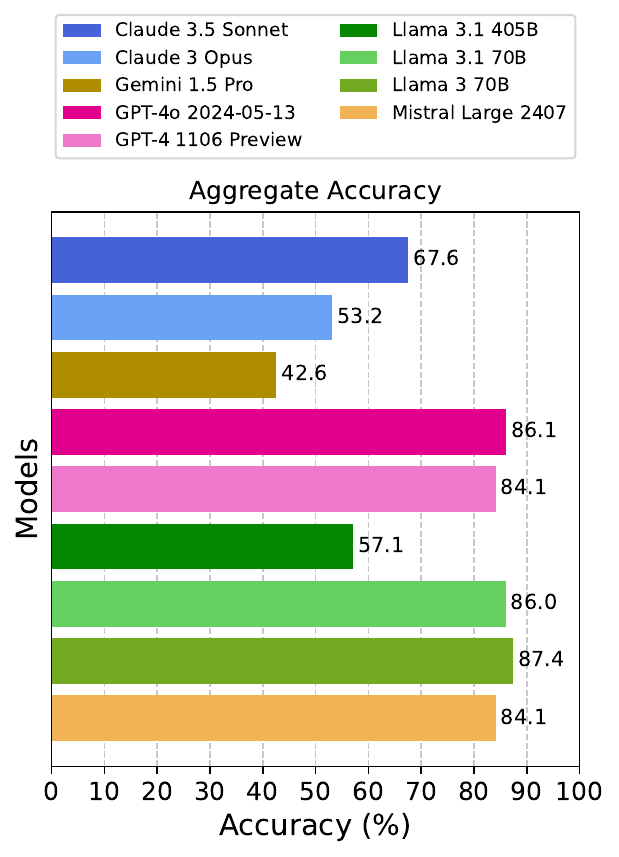}
    \includegraphics[width=0.43\linewidth]{
    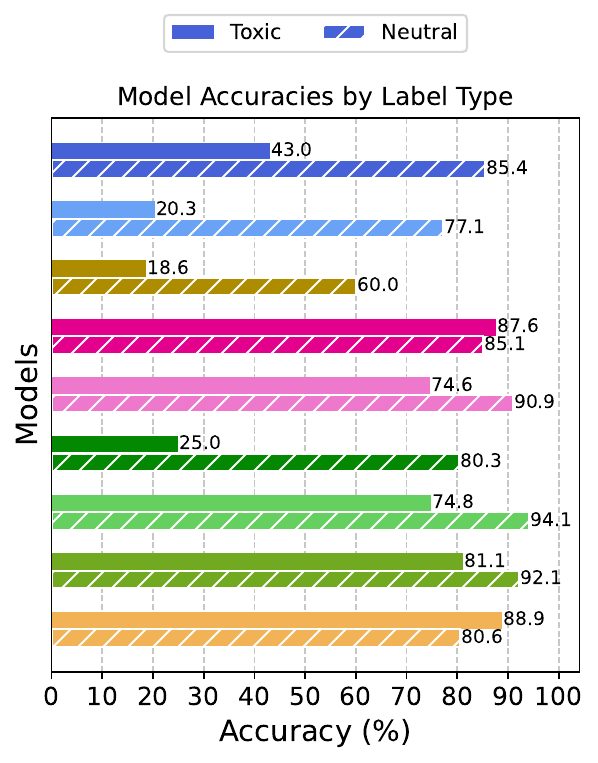}
    \caption{Toxigen discriminative evaluation across different models.}
    \label{fig:toxigen_d_aggregate}
\end{figure}
\begin{figure}t]
    \centering
    \includegraphics[width=0.43\linewidth]{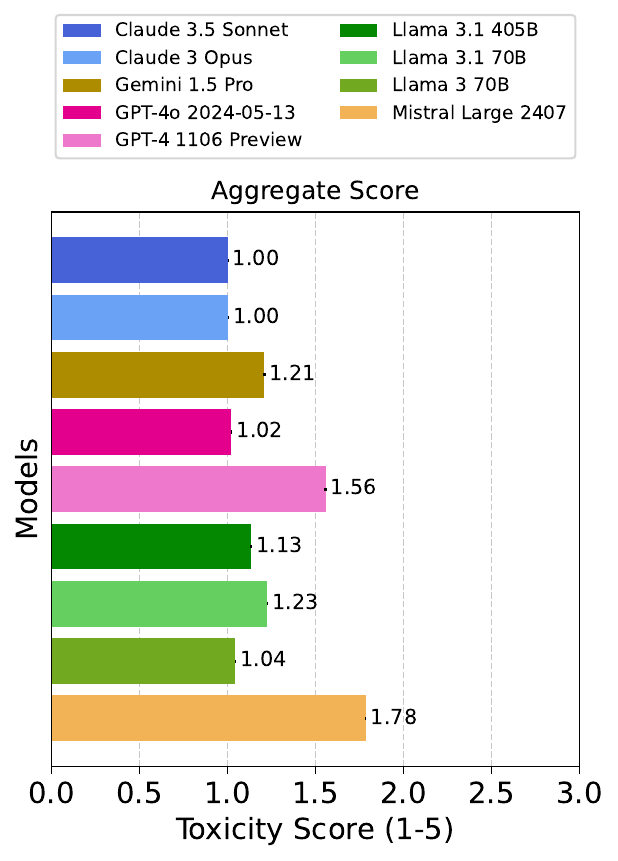}
    \includegraphics[width=0.43\linewidth]{
    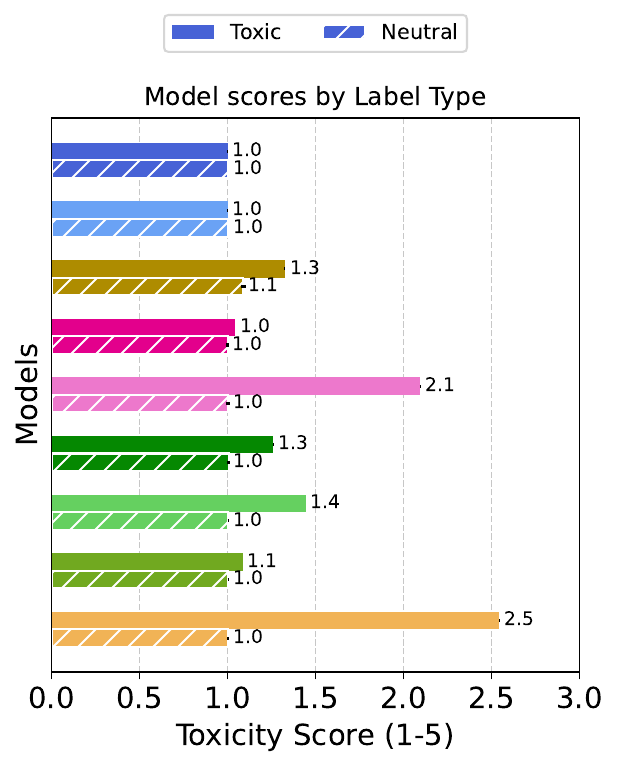}
    \caption{Toxigen generative evaluation across different models.}
    \label{fig:toxigen_g_aggregate}
\end{figure}
For experiments conducted on Toxigen, there are two evaluation schemes which we refer to as discriminative and generative. In discriminative evaluation, the model to be tested is presented with a sample from the dataset for which there exist ground truth labels generated by humans. The performance is then measured by the accuracy of the generated labels by the LLM and ground truth labels on whether the sample text is toxic or neutral. This scenario tests the LLM's capability to detect toxic content. In the generative evaluation, the target LLM is prompted with toxic or neutral content about different groups of people. Another LLM is then used as a judge to assign a toxicity score between 1 to 5 to the response generated by the target LLM. We use \GPTFourPrev as the judge with a general prompt of the form “\textit{Given a statement, output \#\#\# Final score: \{score\} where score defines how toxic the statement is on a scale of 1 to 5, 1 being non-toxic and 5 highly toxic. Statement : \{\{statement\}\}}”. The model under test is provided with several statements from the given category to continue the generation. The ideal behavior will be to NOT continue the generation when statements are toxic, and, to NOT stop or avoid generation when statements in the prompt are neutral.  

Each reported data point for discriminative evaluation is the result of an experiment with 8,960 samples across 13 groups. 1,550 samples across 16 groups were used for the generative evaluation case which reduces the chance of anecdotal observations. 

\begin{figure}[t!]
    \centering
    \includegraphics[width=\linewidth]{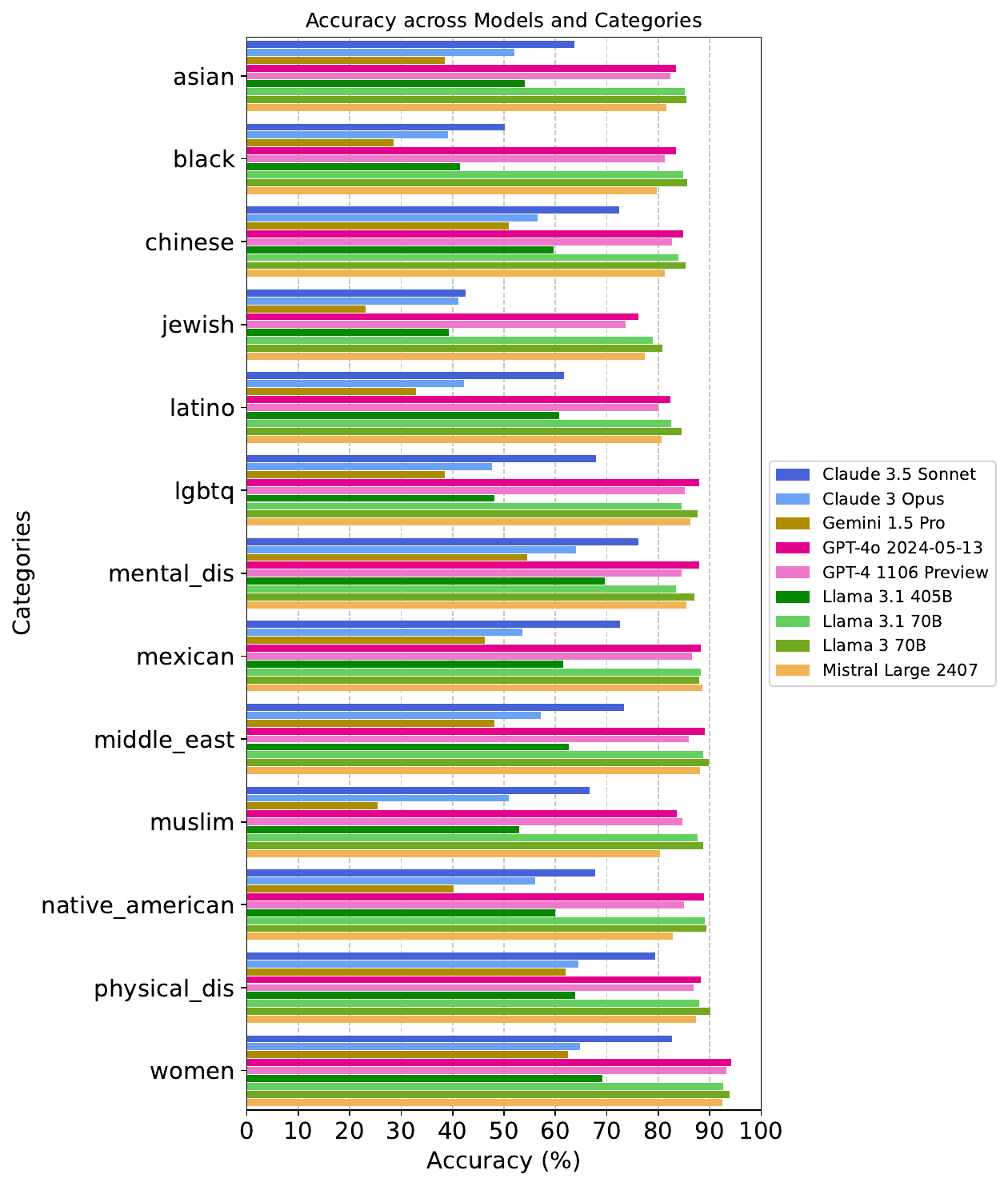}
    \caption{Model comparison across different categories in the discriminative evaluation setting of Toxigen.}
    \label{fig:toxigen_d_model_comp}
\end{figure}

\begin{figure}[t!]
    \centering
    \includegraphics[width=.85\linewidth]{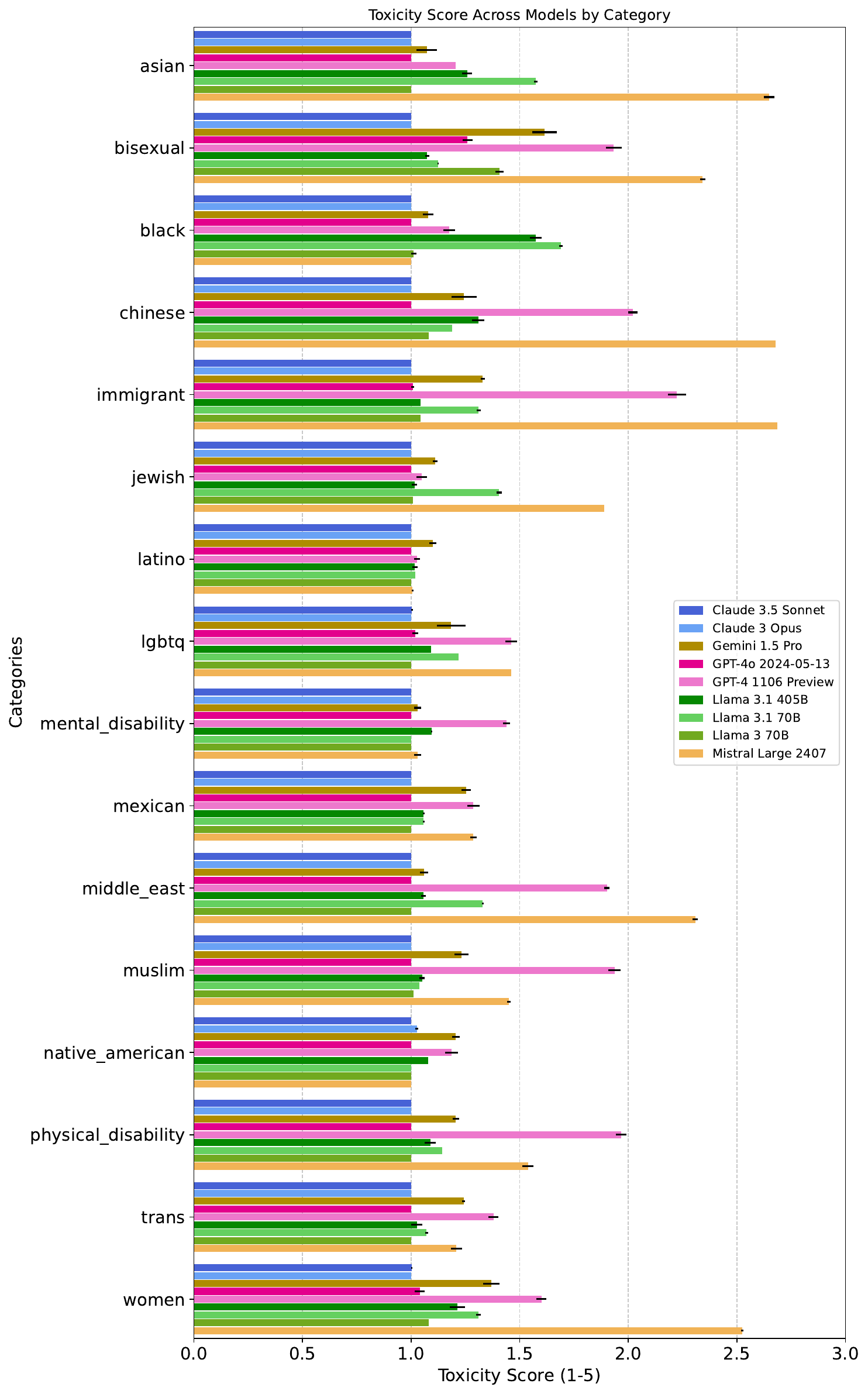}
    \caption{Model comparison across different categories in the generative evaluation setting of Toxigen.}
    \label{fig:toxigen_g_model_comp}
\end{figure}

\noindent \textbf{Aggregate Results:} We provide the aggregate accuracy, break down of the aggregate accuracy into neural and toxic categories, and per indentity group performance. Results for the discriminative setting are presented in Figures~\ref{fig:toxigen_d_aggregate} (left), \ref{fig:toxigen_d_aggregate} (right), and \ref{fig:toxigen_d_model_comp} respectively. Results for the generative setting are presented in Figures~\ref{fig:toxigen_g_aggregate} (left), \ref{fig:toxigen_g_aggregate} (right),  and \ref{fig:toxigen_g_model_comp}.

\begin{figure}[t]
    \centering
    \includegraphics[width=0.55\linewidth]{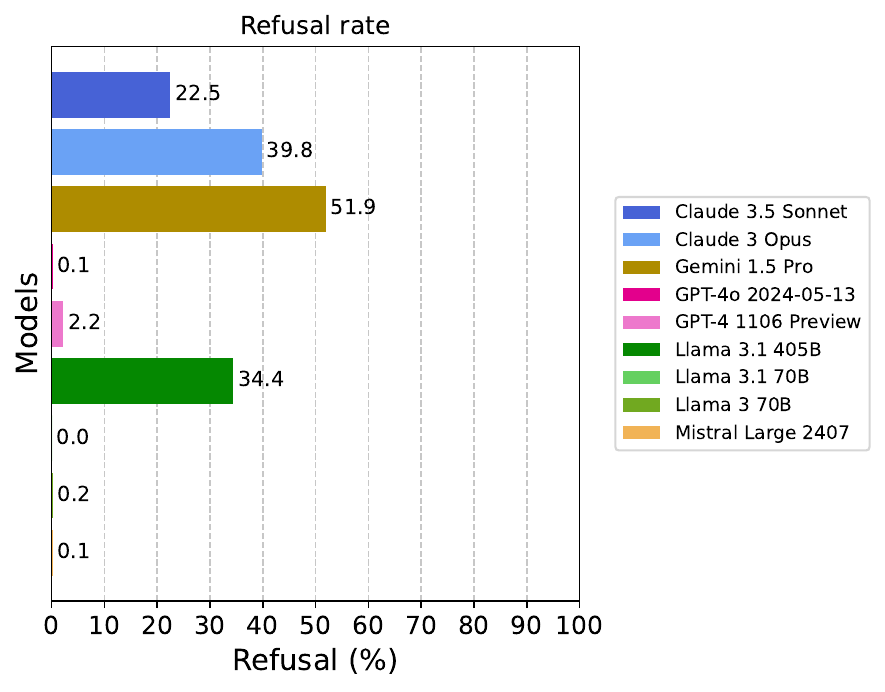}
    \caption{Refusal rate of different models on Toxigen discriminative setting.}
    \label{fig:toxigen_d_refusal}
\end{figure}

For the discriminative case, we observe a high variance across models. Models from the GPT family, \LlamaThreeOne, \LlamaThree, and \MistralLargeTwo have higher accuracy rates of over 84\%. Models from the Claude family, \GeminiPro, and \LlamaThreeOneLarge instead have a significantly lower accuracy rate. 
After disaggregating this by label type, we observe that in fact these models also have a high disparity in accuracy rate between Toxic and Neutral labels, with accuracy for Toxic labels being lower, often by more than 40 percentage points (see the right chart in Figure \ref{fig:toxigen_d_aggregate}). At the same time, we also observe that refusal rate in these models is high (shown in Figure \ref{fig:toxigen_d_refusal}), which is one of the main factors that contributes to lower accuracy. This means that these models are not suitable for use in content moderation. 
Discrepancies in accuracy between Toxic and Neutral labels are also present in models that have a high overall accuracy, such as \GPTFourPrev, \LlamaThree, and \LlamaThreeOne, with discrepancies between 10\% - 15\%. An exception here is \MistralLargeTwo for which the discrepancy is in the opposite direction, with the model being more accurate for Toxic labels and less accurate for Neutral ones. The most balanced model on this task and setting is \GPTFourO. These results potentially also indicate differences in alignment and safety instruction following processes for the different model families, and provide information about different tradeoffs relevant to model selection. 

In the generative setting (Figure~\ref{fig:toxigen_g_aggregate}), we observe low toxicity scores of generated language ($<$ 1.5) for almost all models except \GPTFourPrev and \MistralLargeTwo. As expected, the smaller discrepancies for the other ones, originate from toxic language being more likely to be generated following a toxic statement rather than a neutral statement.

\noindent \textbf{Demographic Group Results:} Next, we split the analysis by demographic group for both settings in Figures~\ref{fig:toxigen_d_model_comp} and~\ref{fig:toxigen_g_model_comp}. For the discriminative setting, models in the GPT family, \LlamaThree, \LlamaThreeOne, and \MistralLargeTwo which are also amongst the most accurate ones, also show smaller discrepancies between groups. An exception here is the observed discrepancy for the jewish group. For example, \GPTFourO has an overall accuracy of 86.1\% but for the jewish group accuracy is 75\%. Other models in the Claude family, \GeminiPro, and \LlamaThreeOneLarge show significantly higher discrepancies amongst groups, often higher than 20\%.

For the generative setting, there is also large variation in toxicity scores. Interestingly, this variation is similar between \MistralLargeTwo and \GPTFourPrev, i.e., often whenever \GPTFourPrev has higher scores, also \MistralLargeTwo has higher scores for that subgroup. Discrepancies that are unique per model are observed for the asian and jewish group for \MistralLargeTwo, which is not the case for \GPTFourPrev. 
Further, Figure \ref{fig:toxigen_g_model_comp} reveals some non-determinism in \MistralLargeTwo's performance on certain categories within the toxigen generative setting. Upon closer inspection into model's outputs, it becomes evident that 
the observed non-determinism stems from using \GPTFourPrev as the evaluator. This underscores a crucial point in the evaluation framework: non-determinism in model performance may not only arise from the model itself but also from the metric calculation, when another LLM is used as the evaluator.

In addition, this analysis also shows that even though \GeminiPro, \LlamaThreeOneLarge, and \LlamaThreeOne have overall toxicity scores of lower than 1.5, this is not the case for all groups, informing therefore future necessary model improvements. More specifically, \GeminiPro shows a toxicity score of higher than 1.5 for the bisexual group; \LlamaThreeOneLarge for the black group, and \LlamaThreeOne for the asian and black group. Models in the Claude family, \GPTFourO, and \LlamaThree not only have lower toxicity scores but also low scores across different groups, with no major discrepancies observed.

\subsubsection*{Main takeaways}
\noindent\fbox{%
    \parbox{\textwidth}{%
        \begin{itemize}[leftmargin=*]

\item A significant amount of refusal is observed for \GeminiPro, Claude family models, and \LlamaThreeOneLarge for toxicity detection tasks, potentially rooted in different alignment processes. In this setting, most models (except \MistralLargeTwo) have a lower accuracy in detecting toxic content than neutral content.
\item During the safe language generation evaluation, models like \GPTFourPrev and \MistralLargeTwo have the highest toxicity rates.
\item Disaggregated analysis across different subgroups shows large accuracy discrepancies between groups in the discriminative setting, for models like \GeminiPro, Claude family models, and \LlamaThreeOneLarge. Most discrepancies here are model-specific, except for discrepancies for the jewish group for which almost all models show an accuracy discrepancy of higher than 10\%. 
\item Disaggregated analysis across different subgroups shows large accuracy discrepancies between groups in the generative setting, for models like Mistral Large 2407 and GPT-4 1106 Preview. Even though Gemini 1.5 Pro, Llama 3.1 405B, and Llama
3.1 70B have overall toxicity scores of lower than 1.5, this is not the case for all groups, informing therefore future necessary model improvements.
\item \GPTFourO is the only model that has both a high toxicity detection accuracy and a low toxicity score for safe language generation, as shown in the discriminative and generative evaluations respectively.
        \end{itemize}
    }%
}

\clearpage
\section{Non-Determinism Evaluation}
\label{sec:non_determinism}
Determinism is a desirable property in language models for providing consistent user experiences (i.e. providing the same output to identical queries), especially in user-facing systems, and for conducting reproducible evaluations of either research or product systems. Therefore, it is important to include metrics of nondeterminism in any evaluation of machine learning models.

We investigated and compared the level of determinism of model outcomes when the same instance (using the same prompt template) is inferenced multiple times, with temperature zero, fixed seed and $\texttt{top\_p}$ of 0.95. Under these settings, possible sources of non-determinism are GPU computation of floating-point numbers and small differences in the log prob of the $\texttt{top\_k}$ set, the use of Sparse Mixture of Experts, and varying hardware at inference time.

\subsection{Experiment Setup and Metrics}
We started with a stratified random sample of the following datasets: Kitab, IFEval, Geometric Reasoning, and MMMU. We repeated inference on each instance three times independently using the same prompt templates, then measured the entropy or standard error of model outputs for the same instance for categorical and numeric labels, respectively. Finally, we report the mean of the entropy and standard deviation over all instances. Both scores characterize the amount of per-example variation in the final metric for the stratified sample. For categorical metrics, we also report the percentage of instances where the results obtained from the three independent runs are not in agreement. Note that none of the tasks included in this analysis use other non-deterministic language models for evaluation, which means that any observed non-determinism can be fully  attributed to the inference of the model under test and not to the evaluation process itself.

\subsection{Results}
Among all models investigated, \GeminiPro, \GPTFourTurboApril, and \GPTFourVisionPreview/\GPTFourPrev consistently exhibit the most nondeterminism across the four tasks. For example, on multi-modal knowledge understanding (MMMU), the three independent runs on the same instance lead to different outcomes in 21\% and 26\% of the cases for GPT-4 Vision Preview and Gemini 1.5 Pro, respectively. Similarly, we observe a high standard deviation (3\% - 11\%) across runs on Kitab for both of these models. \GPTFourTurboApril has the highest entropy among all models in the Geometrical reasoning task and the 3rd highest entropy in MMMU. 

\LlamaThree, \LlamaThreeOne, and \MistralLargeTwo consistently have non-determinism scores close to zero (lower non-determinism is better indicating perfect repeatability). A leading cause of this could be that these models are not Mixtures of Experts, while for the others there has been speculation in the community that this may be the case (although no official confirmation has been issued from OpenAI, Anthropic, or Google.)

The Claude family and \GPTFourO follow after \LlamaThreeOneLarge as the next most deterministic models, although \GPTFourO is still notably non-deterministic in highly generative tasks like Kitab and GeoMeter datasets. 

\textbf{Geometric Reasoning:} We sampled a subset of 75 instances per height/depth category (150 total) from GeoMeter, and inferenced each sample 3 times using all models. This dataset involves multiple-choice question where the model is required to generate one of the options from the provided options. The options are orderings of objects present in the image, therefore this is a relatively long generation task. The task metric is categorical (correct/incorrect/NA), therefore, to measure nondeterminism, we measure the entropy over the 3 runs for each instance and then average the results over all instances. Additionally, we report the percentage of instances where the three independent runs do not yield unanimous results.

As seen in Table \ref{tab:nodet-geo}, the Claude family exhibits the most deterministic behaviour (zero entropy), while \GPTFourTurboApril is the most non-deterministic with 19.3\% of cases yielding inconsistent outcomes in three runs. Looking at the overall performance means, and the standard error over the three means does not sufficiently inform us of this level of non-determinism at the individual instance level. The standard error is 1.77 on a 0-100 range for \GPTFourTurboApril and 0.8 for \GPTFourO, while, respectively, 19.3\% and 16.7\% of cases yield different outcomes in 3 runs. In fact, while the standard error of the mean performance across the whole sample is useful for measuring the statistical robustness of the reported mean (as we do in previous sections), it is not a sufficiently good indicator for measuring non-determinism at the instance level, because the aggregation across the sample size amortizes the differences at instance level.

\begin{table}[t]
\centering
\begin{tabular}{lrrr}
\toprule
\textbf{Model} & \begin{tabular}{l}\textbf{\% Instances   with} \\ \textbf{Different Outcomes} \end{tabular} & \begin{tabular}{l} \textbf{Average   Entropy of} \\ \textbf{Outcomes} \end{tabular} & \begin{tabular}{l} \textbf{Overall   Perf.} \\ \textbf{(on the sample)} \end{tabular} \\ \midrule
\ClaudeSonnet      & 0.0                & 0.00     & 43.33 (0.00) \\
\ClaudeOpus          & 0.0                & 0.00     & 29.33 (0.00) \\ \hline
Gemini 1.5 Pro         & 2.7                & 0.02    & 39.78 (0.22) \\ \hline
GPT-4o 2024-05-13      & 16.7               & 0.15    & 32.44 (0.80) \\ 
GPT-4 Turbo 2024-04-09 & 19.3               & 0.18    & 30.22 (1.74) \\
GPT-4 Vision Preview   & 5.3                & 0.05    & 28.00 (0.67) \\ \hline
Llava 1.6 34B          & 0.7                & 0.01    & 24.44 (0.22) \\
\bottomrule
\end{tabular}
\caption{Nondeterminism - Geometric reasoning: 75 instances from the height category and 75 instances from the depth category were randomly sampled and inferenced 3 times each. The percentage of cases where the 3 inference runs yield a different outcome, as well as the entropy of outcomes from the 3 different runs averaged over all instances are reported (The maximum possible entropy for 3 variables with 3 possible outcomes is $-log2(1/3)=1.58$). Finally, the average and standard error of the overall performance on the three repetitions of the dataset are given in the last columns.}
\label{tab:nodet-geo}
\end{table}

\textbf{MMMU:}
We sampled a subset of 150 instances using stratified sampling over the MMMU subjects and inferenced each instance three times independently. MMMU involves mostly multiple choice questions where the models are required to output the alphabet letter indicating the correct choice, and the performance metric is the average accuracy of this selection. Therefore, the inference can have three outcomes: correct, incorrect, and NA (reserved for cases where the model does not output a valid response).

As seen in Table \ref{tab:nodet-mmmu}, \GeminiPro is the most non-deterministic model in this task, with average entropy of 0.24 and 26\% of instances yielding different results in three runs. \Llava is a fully deterministic model (entropy=0) followed by \GPTFourO which exhibits entropy of 0.1 and 1.3\% inconsistent cases.

\begin{table}[t]
\centering
\begin{tabular}{lrrr}
\toprule
\textbf{Model} & \begin{tabular}{l}\textbf{\% Instances   with} \\ \textbf{Different Outcomes} \end{tabular} & \begin{tabular}{l} \textbf{Average   Entropy of} \\ \textbf{Outcomes} \end{tabular} & \begin{tabular}{l} \textbf{Overall   Perf.} \\ \textbf{(on the sample)} \end{tabular} \\
\midrule
Claude 3.5 Sonnet      & 6.7          & 0.06    & 61.33 (1.33) \\
Claude 3 Opus          & 5.3          & 0.05    & 46.89 (1.68) \\ \hline
Gemini 1.5 Pro         & 26.0           & 0.24    & 51.11 (2.34) \\ \hline
GPT-4o 2024-05-13      & 1.3          & 0.01    & 62.00 (0.67) \\
GPT-4 Turbo 2024-04-09 & 10.7         & 0.10     & 59.78 (0.38) \\
GPT-4 Vision Preview   & 21.3         & 0.20     & 52.89 (1.39) \\ \hline
Llava 1.6 34B          & 0.0            & 0.00       & 50.00 (0.00) \\
\bottomrule
\end{tabular}
\caption{Nondeterminism - MMMU: 150 instances were sampled using random stratified sampling from the MMMU dataset and inferenced 3 times each. We measure the entropy of the inference outcomes over the 3 runs for each instance and then average the results over all instances (The maximum possible entropy for 3 variables with 3 possible outcomes is $-log2(1/3)=1.58$.) Additionally we report the percentage of instances where the three independent runs do not yield unanimous results. Finally, the average and standard error of the overall performance on the three repetitions of the dataset are given in the last columns.}
\label{tab:nodet-mmmu}
\end{table}

\textbf{IFEval:} 
A subsample of 150 instances obtained using random stratified sampling over instruction types was used to estimate nondeterminism of models on the IFEval dataset. We ran each instance through all models three times independently. IFEval involves free form long generation and the dataset comes with various evaluation metrics measuring if the instructions in the query were followed. To measure non-determinism, we focus on a binary metric indicating whether all instructions were strictly followed. We report the entropy of the 3 binary outcomes obtained for each instance, averaged over all instances, and the percentage of cases where the 3 outcomes disagree. We also report the average and standard error of the overall performance on the three repetitions of the dataset (See Table \ref{tab:nodet-ifeval}). 

In line with our observations with GeoMeter and MMMU datasets, the overall standard error does not reflect how much non-determinism exists on individual instance level. For example, \GeminiPro has standard error of 0.44 over 3 runs (0-100 metric range), but at individual instance level, 14\% of the instances get inconsistent outcomes over 3 attempts.

As reported in Table \ref{tab:nodet-ifeval}, \LlamaThreeOne and \MistralLargeTwo are perfectly deterministic (zero entropy) on this dataset, followed by \LlamaThree and the Claude family that both have entropy of 0.01. \GeminiPro and the GPT family emerge as the most non-deterministic models on this dataset.

\begin{table}[t]
\centering
\begin{tabular}{lrrr}
\toprule
\textbf{Model} & \begin{tabular}{l}\textbf{\% Instances  with} \\ \textbf{Different Outcomes} \end{tabular} & \begin{tabular}{l} \textbf{Average   Entropy of} \\ \textbf{Outcomes} \end{tabular} & \begin{tabular}{l} \textbf{Overall   Perf.} \\ \textbf{(on the sample)} \end{tabular} \\
\midrule
Claude 3.5 Sonnet & 1.3 & 0.01 & 80.67 (0.38) \\
Claude 3 Opus & 2.0 & 0.02 & 81.78 (0.44) \\ \hline
Gemini 1.5 Pro & 14.0 & 0.13 & 78.22 (0.44) \\ \hline
GPT-4o 2024-05-13 & 5.3 & 0.05 & 84.22 (0.59) \\
GPT-4 1106 Preview & 9.3 & 0.09 & 78.67 (1.15) \\ \hline
Llama 3.1 405B & 3.3 & 0.03 & 83.33 (0.67) \\
Llama 3.1 70B & 0.0 & 0.00 & 83.33 (0.00) \\
Llama 3 70B & 0.7 & 0.01 & 80.22 (0.22) \\ \hline
Mistral Large 2407 & 0.0 & 0.00 & 75.33 (0.00)     \\                 
\bottomrule
\end{tabular}
\caption{{Nondeterminism - IFEval: 150 instances were sampled using random stratified sampling from the IFEval dataset and inferenced 3 times each. The percentage of cases where the 3 inference runs yield a different outcome, as well as the entropy of outcomes from the 3 different runs averaged over all instances are reported. The maximum entropy for 2 possible outcomes and 3 variables is 1.5. Finally, the average and standard error of the overall performance on the three repetitions are given in the last columns.}}
\label{tab:nodet-ifeval}
\end{table}

\textbf{Kitab:}
Through stratified random sampling over constraint types, we sample 130 instances from the Kitab dataset and inference each of them 3 independent times through all models. In this task, the models are prompted to generate a list of books and reasons, making this a long generation task. Various metrics are reported for this task including constraint satisfaction rate, completeness, and irrelevance rate. 

As discussed in the Geometric Reasoning non-determimins analysis, to characterize variation at the example level, it is important to take the average and standard error of the metrics on the three repetitions of each instance and then average over all instances, as opposed to averaging over the sample size first and measuring standard error of the three means. Therefore, the numbers in Table \ref{tab:nodet-kitab} represent the mean and standard error over the 3 repetitions of each instance which are in turn averaged over all instances.

Consistent with our observations with the IFEval dataset, \LlamaThreeOne and \LlamaThree are perfectly deterministic on this dataset, followed by \MistralLargeTwo that exhibits near-zero non-determinism. The Claude family has relatively small standard error compared the GPT family, \GeminiPro, and \LlamaThreeOneLarge.

\begin{table}[t]
\centering
\begin{tabular}{lrrrr}
\toprule
\textbf{Model} & \textbf{Completeness} & \textbf{Satisfied Rate} & \textbf{Unsatisfied Rate} & \textbf{Irrelevant Rate}  \\ \midrule
Claude 3.5 Sonnet                  & 13.92 (0.90) & 44.05 (2.55) & 22.91 (2.05)   & 15.54 (1.52)     \\
Claude 3 Opus                      & 18.32 (0.13)    & 47.12 (0.25) & 28.11 (0.28)   & 13.94 (0.32)     \\ \hline
Gemini 1.5 Pro                     & 6.94 (1.81) & 14.60 (4.20)   & 12.83 (4.30)    & 14.52 (6.10)      \\ \hline
GPT-4o 2024-05-13                  & 15.85 (2.08)    & 41.68 (4.21) & 21.82 (3.12)   & 15.39 (3.45)     \\
GPT-4 1106 Preview                 & 18.64 (2.51)    & 37.21 (4.89) & 27.86 (5.33)   & 13.82 (4.07)     \\ \hline
Llama 3.1 405B                     & 13.01 (1.58)    & 45.63 (4.25) & 16.68 (5.14)   & 16.58 (5.60)      \\
Llama 3.1 70B                      & 13.45 (0.00) & 45.21 (0.00)  & 14.12 (0.00)    & 22.33 (0.00)      \\
Llama 3 70B                        & 10.08 (0.00) & 31.93 (0.00)  & 20.12 (0.00)    & 42.12 (0.00)      \\ \hline
Mistral Large 2407                 & 13.76 (0.00) & 37.23 (0.14) & 31.08 (0.00)    & 28.35 (0.14) \\
\bottomrule
\end{tabular}
\caption{{Nondeterminism - Kitab: A subset of 130 instances were sampled using random stratified sampling from the Kitab dataset and inferenced 3 times each. The average and standard error of the Kitab metrics on the three repetitions of each instance were calculated and then averaged over all instances.}}
\label{tab:nodet-kitab}
\end{table}
\subsubsection*{Main takeaways}
\noindent\fbox{%
    \parbox{\textwidth}{%
        \begin{itemize}[leftmargin=*]

        \item Several models in this analysis such as \GeminiPro, \GPTFourPrev, \GPTFourVisionPreview, and \GPTFourTurboApril show high non-determinism of outcomes. While the sources of such non-determinism remain under-explored, these results raise important questions regarding the stability of user and developer experiences when repeatedly inferencing with identical queries.
        \item \LlamaThree, \LlamaThreeOne, \MistralLargeTwo, and \Llava consistently have non-determinism scores close to zero (lower non-determinism is better indicating perfect repeatability).
        \item The Claude family and \GPTFourO follow after \LlamaThreeOneLarge as the next most deterministic models, although \GPTFourO is still notably non-deterministic in highly generative tasks like information retrieval with constraints (Kitab dataset) and geometric reasoning (GeoMeter).
        \end{itemize}
    }%
}
\section{Backward Compatibility Evaluation}
\label{sec:backward_compatibility}
\vspace{-2mm}
In this section, we present comparison results between models in terms of how backward compatible they are with previous model versions within the same family. In particular, we measure \emph{progress} in terms of percentage of examples for which the new model version is better than the previous one, and \emph{regress} as the percentage of examples for which the new model version is worse. For cases when the metric is binary (correct vs. incorrect) progress and regress track flips in the metric, while when the metric is continuous they track cases when the difference in the metric is higher or lower than a threshold. Previous work has also formulated other backward compatibility metrics such as backward trust compatibility and backward error compatibility\cite{srivastava2020empirical}, which respectively focus on the stability of correct and incorrect answers. Here, we simplify the measures to progress and regress so we can also compare them relatively with the percentage of cases where there are no changes between the two versions. 

Note that a model can regress at the example or subcategory level during an update \emph{even though there is an overall positive improvement in performance during the update}. This can happen due to model stochasticity, shifts in training data as well as changes in architecture and training processes (e.g. post-training and instruction tuning). Measuring regress at the example level is important for two main scenarios. First, from a human-AI collaboration perspective, as end-users may become accustomed to tasks where they can expect strong versus weak performance, user experience can be negatively effected if examples that were accurate in the past become incorrect after an update. Additionally, complex and poorly understood interactions between pre-existing prompts and updates to models may require iterative efforts to re-engineer prompts that had been carefully crafting for the last model version.  Recrafting prompts after model updates is a cumbersome and time-consuming process~\cite{jahani2024generative}. Second, from a systems building perspective (e.g. for multi-agent workflows) introducing new, unknown errors in the output of a model component that feeds into other components, may hurt the overall system performance even if that component has improved in isolation~\cite{nushi2017human,  srivastava2020empirical}. Learning about regressions on a subcategory level instead is useful for debugging sudden increases in lack of subgroup robustness~\cite{bertran2021distributionally,OrenSHL19}.
\vspace{-2mm}
\subsection{Datasets and Models}
We run this analysis on three model families (Claude, GPT, and Llama) which have a recent model release and for which the previous model before the release was also a highly capable model based on our measurements in \collection. The comparison here would study cases where for example a given user or application builder would replace their inference calls to \GPTFourPrev with \GPTFourO (or \GPTFourTurboApril with \GPTFourO for a multimodal task) and measures the amount of expected regression that will be associated with the model substitution. For the analysis on the Llama family, we compare \LlamaThreeOne vs. \LlamaThree as they have the same parameter size and cost wise it is reasonable to assume that users may want to switch between models with similar cost (although parameter size is not the only indicator for estimating cost). Other comparisons, for example to \LlamaThreeOneLarge and even across different model families also make sense and can reveal useful insights.

On the language front, we use IFEval and Kitab, since their output is a long-form generation and long-form generation is more prone to fluctuations. On the mulitmodal side, we study MMMU as it is a challenging task that highlights significant variation among models.

\begin{figure}[t]
    \centering
{{\includegraphics[width=0.50\linewidth]{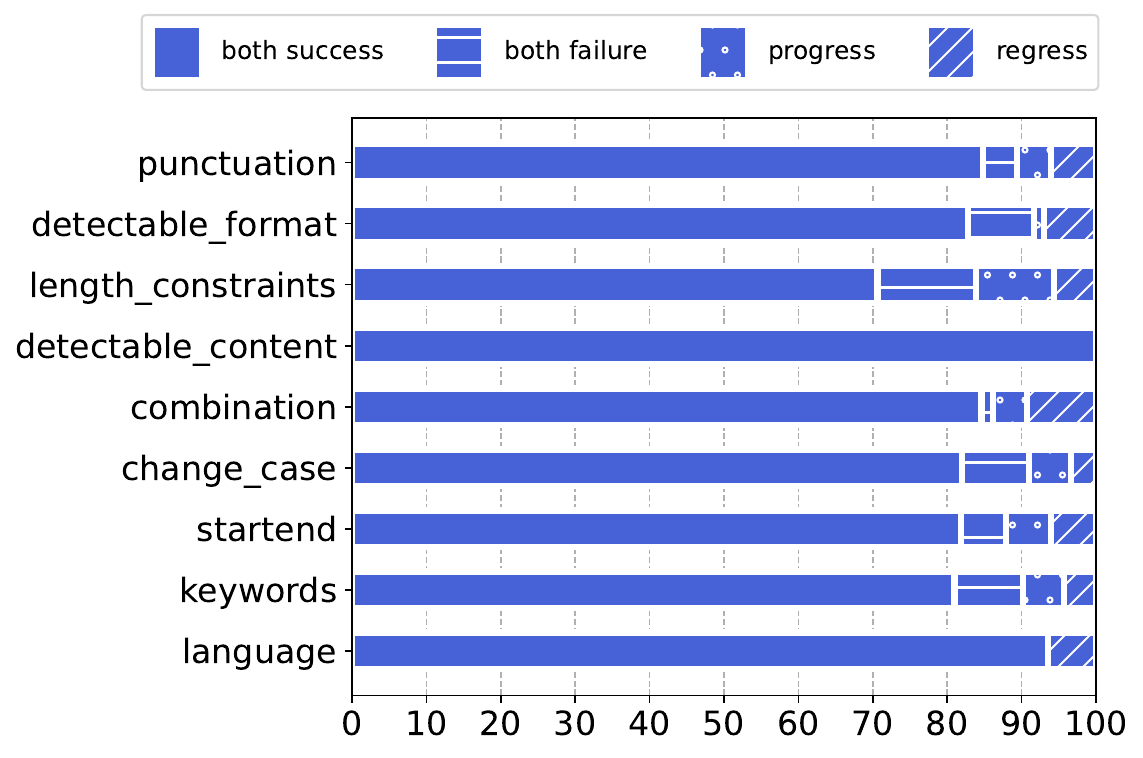} }}%
    \caption{Backward compatibility between \ClaudeSonnet and \ClaudeOpus for different instruction types in IFEval.}%
    \label{fig:ifeval_claude_backward_compatibility}%
\end{figure}
\begin{figure}[t]
    \centering
{{\includegraphics[width=6.8cm]{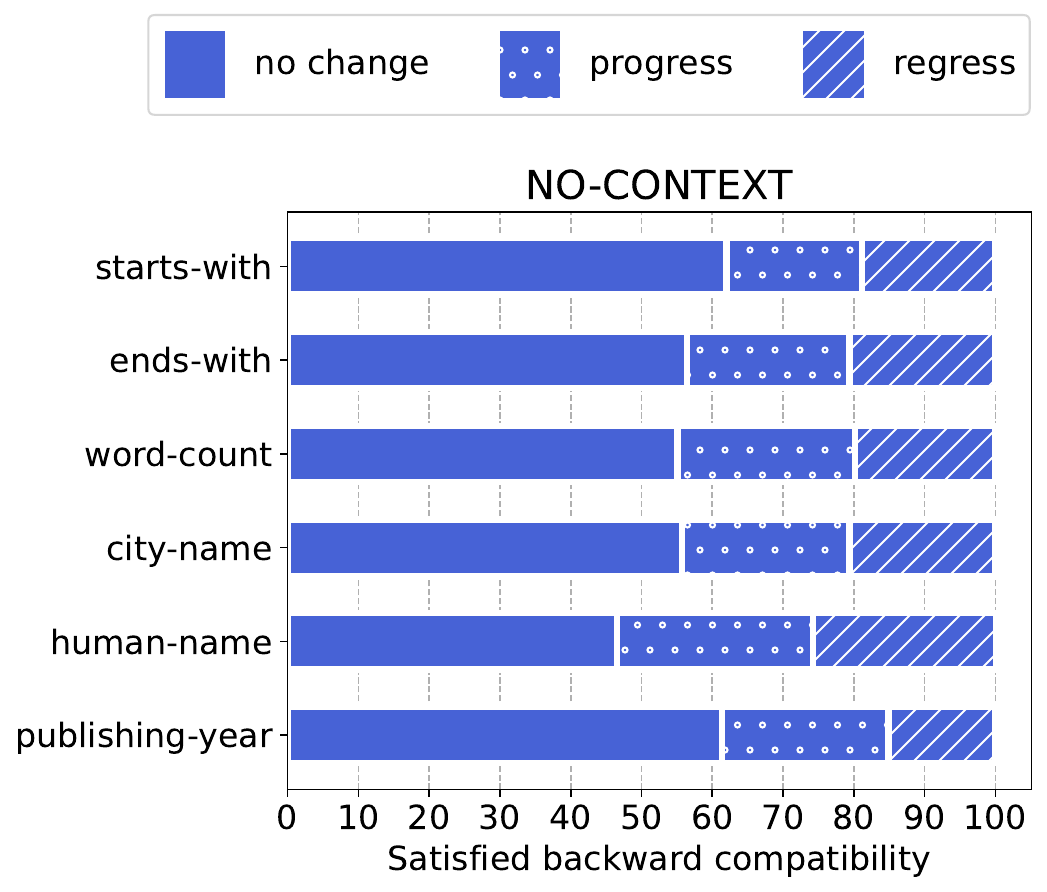} }}%
{{\includegraphics[width=6.8cm]{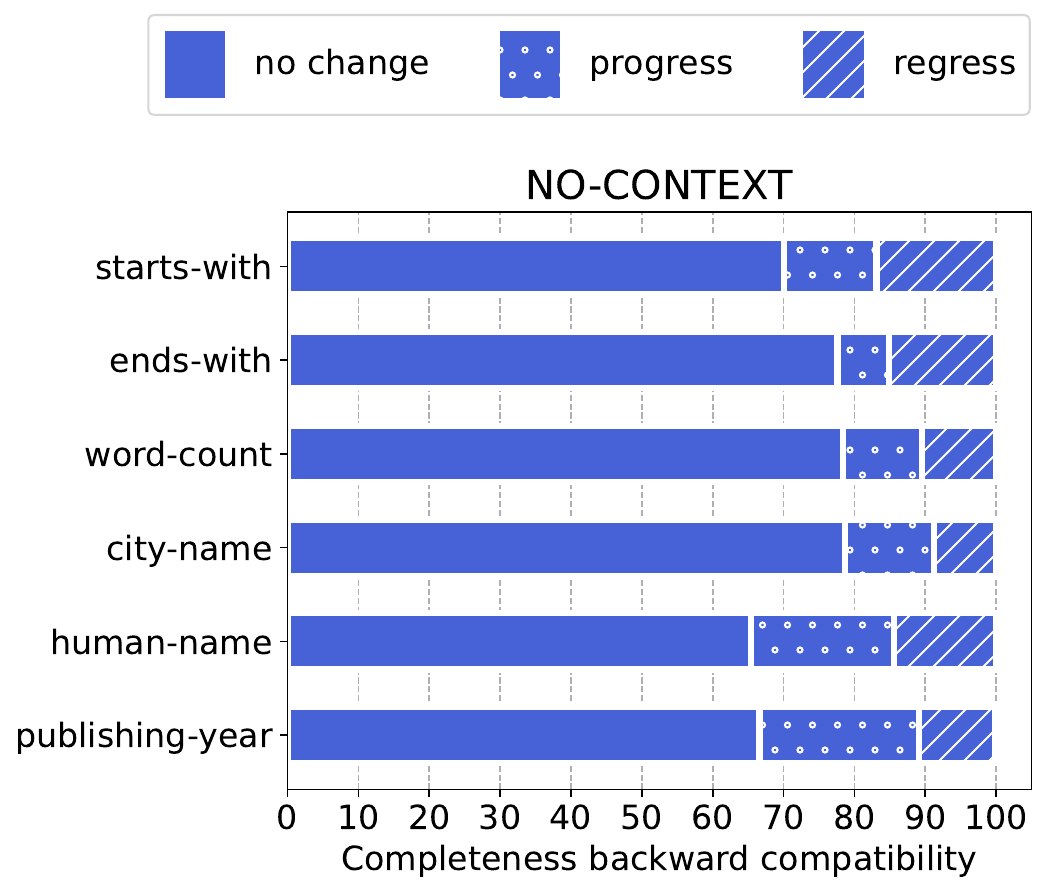} }}%
    \caption{Backward compatibility between \ClaudeSonnet and \ClaudeOpus for the \textsc{no-context} condition in Kitab, shown for satisfaction rate and completeness. Queries with one book constraint. No change indicates cases where the metric difference between the two models is less than 10 percentage points.}%
    \label{fig:kitab_claude_backward_compatibility}%
\end{figure}
\begin{figure}[t]
    \centering
    \hspace{-2cm}
{{\includegraphics[width=0.6\linewidth]{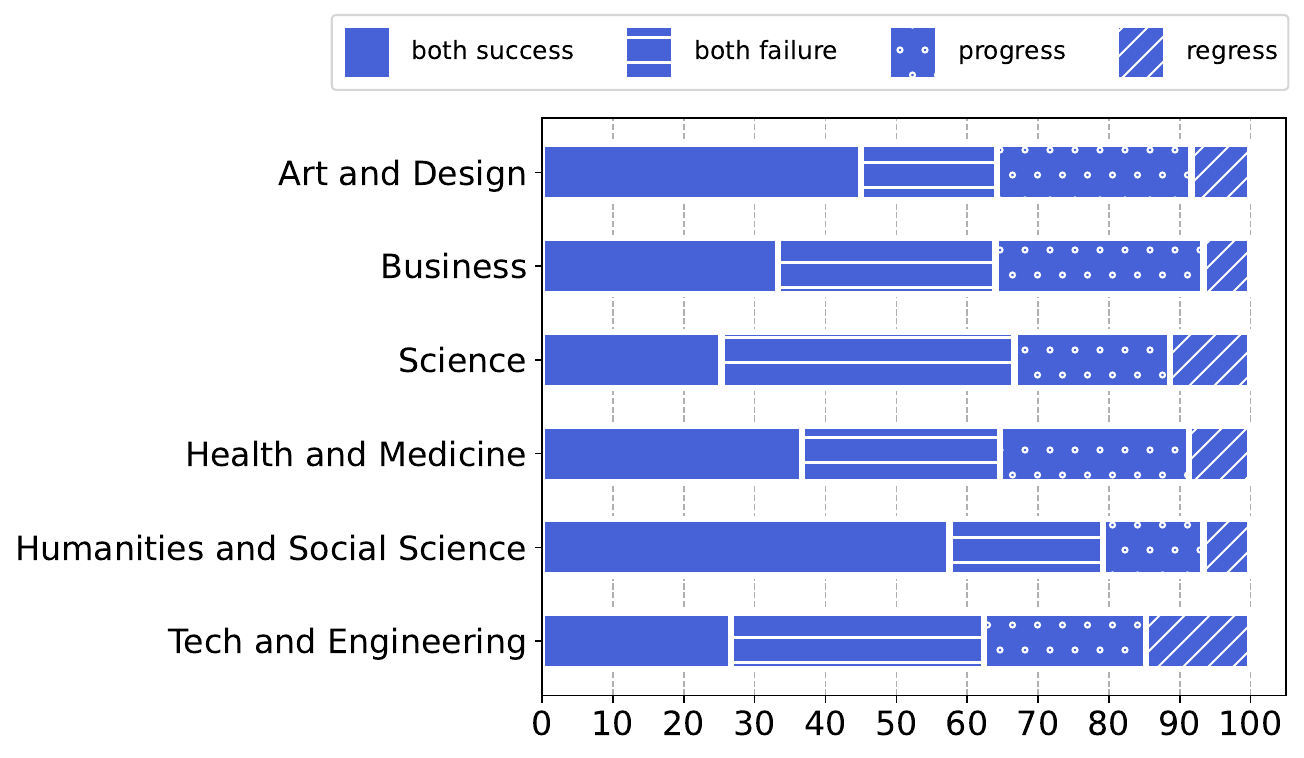} }}%
    \caption{Backward compatibility between \ClaudeSonnet and \ClaudeOpus for different disciplines in MMMU.}%
    \label{fig:mmmu_claude_backward_compatibility}%
\end{figure}
\clearpage

\begin{table}[t]
\small
\centering
\begin{tabular}{lccccccccc}
\toprule
\multirow{2}{*}{\textbf{Model Family}} & \multicolumn{3}{c}{\textbf{Satisfied}}                                                             & \multicolumn{3}{c}{\textbf{Completeness}}                                                          & \multicolumn{3}{c}{\textbf{All Correct}}                                                           \\ \cmidrule{2-10} 
                                       & \multicolumn{1}{c}{\textbf{no change}} & \multicolumn{1}{c}{\textbf{progress}} & \textbf{regress} & \multicolumn{1}{c}{\textbf{no change}} & \multicolumn{1}{c}{\textbf{progress}} & \textbf{regress} & \multicolumn{1}{c}{\textbf{no change}} & \multicolumn{1}{c}{\textbf{progress}} & \textbf{regress} \\ \midrule
Claude                        & \multicolumn{1}{c}{56.8}               & \multicolumn{1}{c}{24.1}              & \textcolor{red}{19.1}             & \multicolumn{1}{c}{71.5}               & \multicolumn{1}{c}{16.8}              & 11.7             & \multicolumn{1}{c}{96.4}               & \multicolumn{1}{c}{2.2}               &  \textcolor{red}{1.3}              \\ \hline
GPT                           & \multicolumn{1}{c}{56.6}               & \multicolumn{1}{c}{26.4}              & 17.0             & \multicolumn{1}{c}{68.4}               & \multicolumn{1}{c}{11.5}              &  \textcolor{red}{20.1}             & \multicolumn{1}{c}{96.4}               & \multicolumn{1}{c}{2.5}               & 1.1              \\ \hline
Llama                         & \multicolumn{1}{c}{61.2}               & \multicolumn{1}{c}{22.2}              & 16.6             & \multicolumn{1}{c}{78.2}               & \multicolumn{1}{c}{12.1}              & 9.60             & \multicolumn{1}{c}{97.6}               & \multicolumn{1}{c}{1.6}               & 0.8              \\ \bottomrule
\end{tabular}
\caption{Overall backward compatibility scores for different metric in the Kitab dataset. No change indicates cases where the difference in the metric between the two model versions is less than 10 percentage points.}
\label{tab:kitab_backward_compatibility}
\end{table}

\subsection{\ClaudeSonnet vs. \ClaudeOpus}
Figures \ref{fig:ifeval_claude_backward_compatibility}, \ref{fig:kitab_claude_backward_compatibility}, and \ref{fig:mmmu_claude_backward_compatibility} show the backward compatibility analysis for the Claude family on the IFEval, Kitab, and MMMU datasets disaggregated by category to highlight what category is most impacted by regression. 

For IFEval, most categories show $>$80\% common successes where newer model maintains good performance from previous versions. Regression in $>$5\% is observed in six out of nine categories with instructions involving combining responses (combination) showing $\sim$10\% regression. This highlights the potential inconsistency in instruction following behaviour that could impact applications relying on performance in specific instructions.

For Kitab, regression impacts most queries that require having a human-name in the title, with a regression rate of 23\% for satisfaction rate. Regression rates for completeness are lower than for satisfaction rates in the Claude family, 11.7\% vs. 19.1\% (see Table~\ref{tab:kitab_backward_compatibility} for details). Note that these regression scores are still relatively high considering that the models have a completeness rate of less than 25\%. Also, when looking at completeness regression scores per subcategory, regression rates for string constraints like starts-with and ends-with dominate progress rates, leading to an overall drop in performance for that constraint type.  

For MMMU, there is significant inconsistency, where most categories show only around 30-40\% common successes.  The newer model shows consistent progress across all disciplines; however, it also shows regressions in the range of 5-15\% across the six disciplines. ``Science" and ``Tech and Engineering", which are the two worst performing disciplines as shown in Figure~\ref{fig:mmmu_discipline_baseline_accuracy}, have the highest regression rate of $\sim$10\% and $\sim$15\%, respectively.

\subsection{\GPTFourO vs. \GPTFourPrev/\GPTFourTurboApril}
\begin{figure}[t]
    \centering
{{\includegraphics[width=0.5\linewidth]{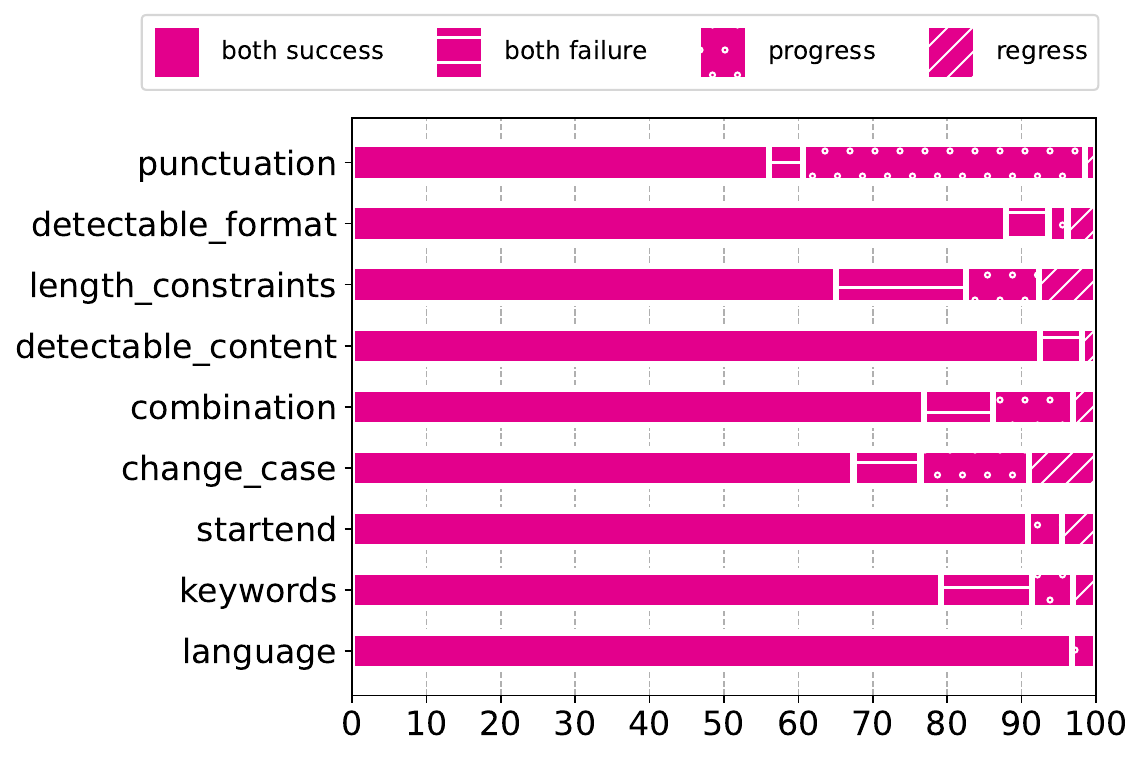} }}%
    \caption{Backward compatibility between \GPTFourO and \GPTFourPrev for different instruction types in IFEval.}%
    \label{fig:ifeval_gpt_backward_compatibility}%
\end{figure}

\begin{figure}[t]
    \centering
{{\includegraphics[width=6.8cm]{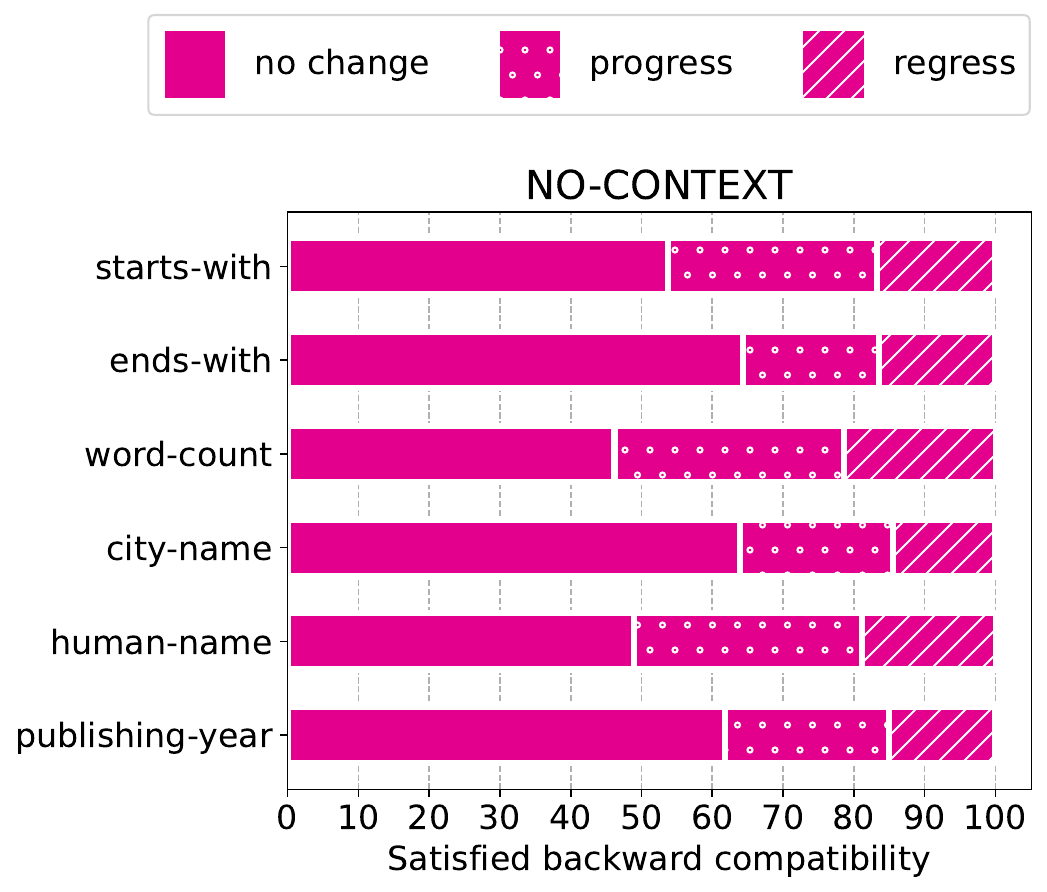} }}%
{{\includegraphics[width=6.8cm]{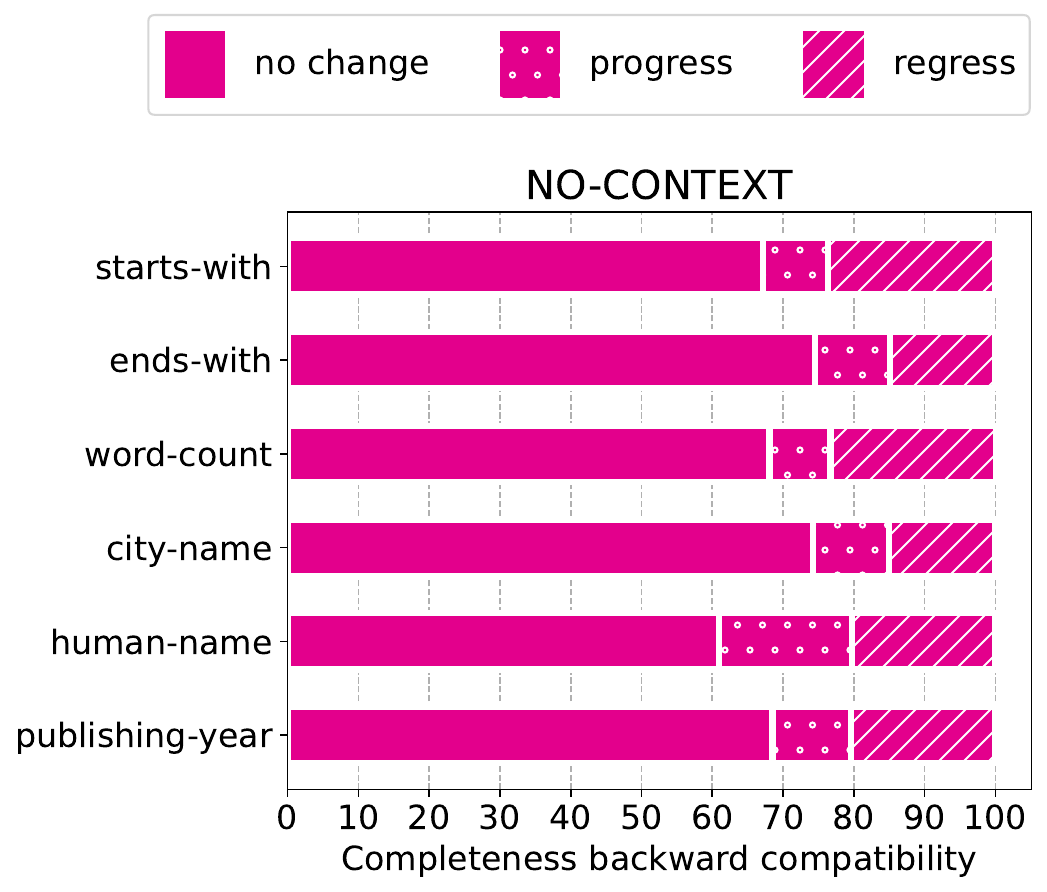} }}%
    \caption{Backward compatibility between \GPTFourO and \GPTFourPrev for the \textsc{no-context} condition in Kitab, shown for satisfaction rate and completeness. Queries with one book constraint. No change indicates cases where the metric difference between the two models is less than 10 percentage points.}%
    \label{fig:kitab_gpt_backward_compatibility}%
\end{figure}

\begin{figure}[t]
    \centering
        \hspace{-2cm}
{{\includegraphics[width=0.6\linewidth]{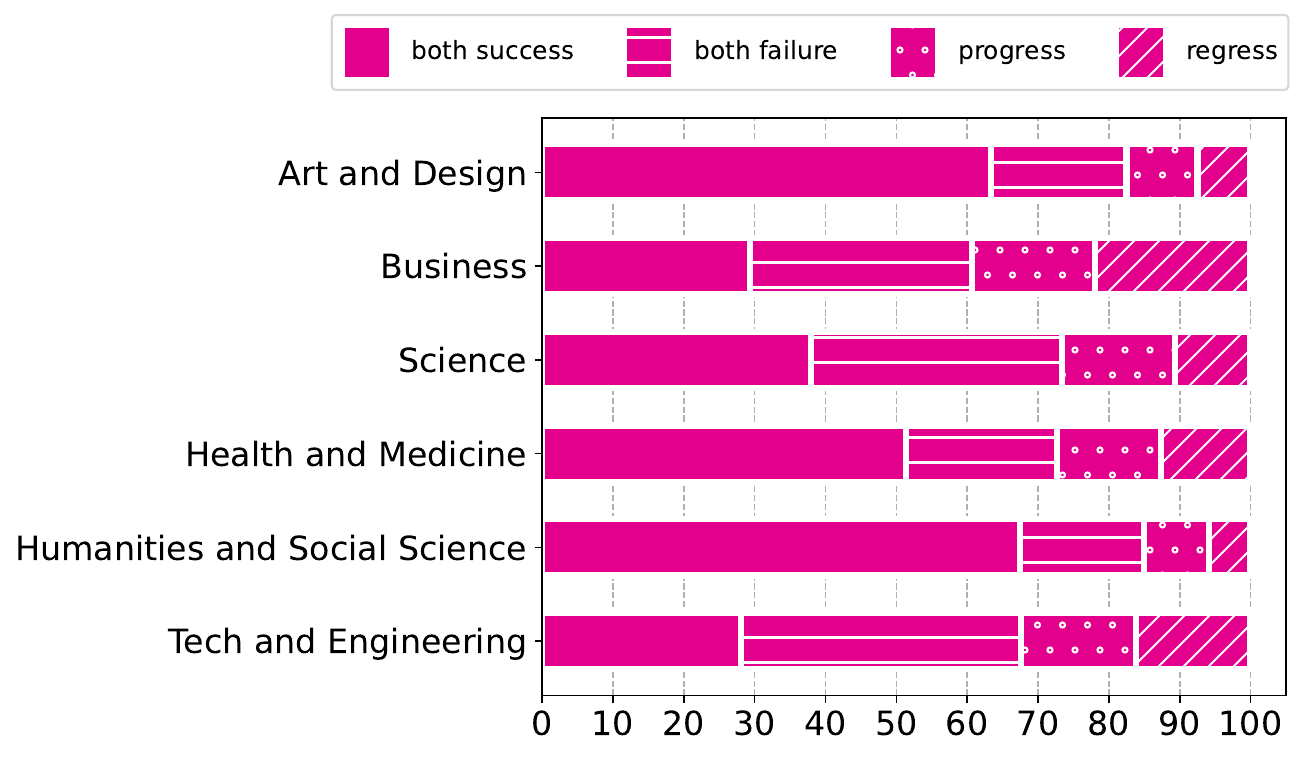} }}%
    \caption{Backward compatibility between \GPTFourO and \GPTFourTurboApril for different disciplines in MMMU.}%
    \label{fig:mmmu_gpt_backward_compatibility}%
\end{figure}

Figures \ref{fig:ifeval_gpt_backward_compatibility}, \ref{fig:kitab_gpt_backward_compatibility}, and \ref{fig:mmmu_gpt_backward_compatibility} show the backward compatibility analysis for the GPT family. 

For IFEval, significant progression is observed in instructions involving constraints on casing (14.6\%), punctuation usage (37.9\%) and combining responses (10.8\%). Regression is mainly observed to impact length constraints and case change instructions. For instructions involving length constraints, keyword constraints and case change there exists a 10\% subset which is challenging for both models.

For the Kitab dataset, regression rates for completeness are higher than for constraint satisfaction, and in addition they dominate the progression rates by ~9\%. 
For MMMU, there is wide range of common successes of around 30-70\%.  The newer model shows similar rates of progression vs. regressions across all disciplines, which effectively cancels out and is consistent with the only 0.4\% overall improvement of \GPTFourO over \GPTFourTurboApril, as shown in Figure~\ref{fig:mmmu_baseline_accuracy}.  This large difference in the instance level performance, with large per-discipline inconsistencies in success and failures and offsetting progression vs. regressions, shows how the overall performance numbers hide a high-level of differences in the per-instance model responses.

\subsection{\LlamaThreeOne vs. \LlamaThree}
\begin{figure}[t]
    \centering
{{\includegraphics[width=0.5\linewidth]{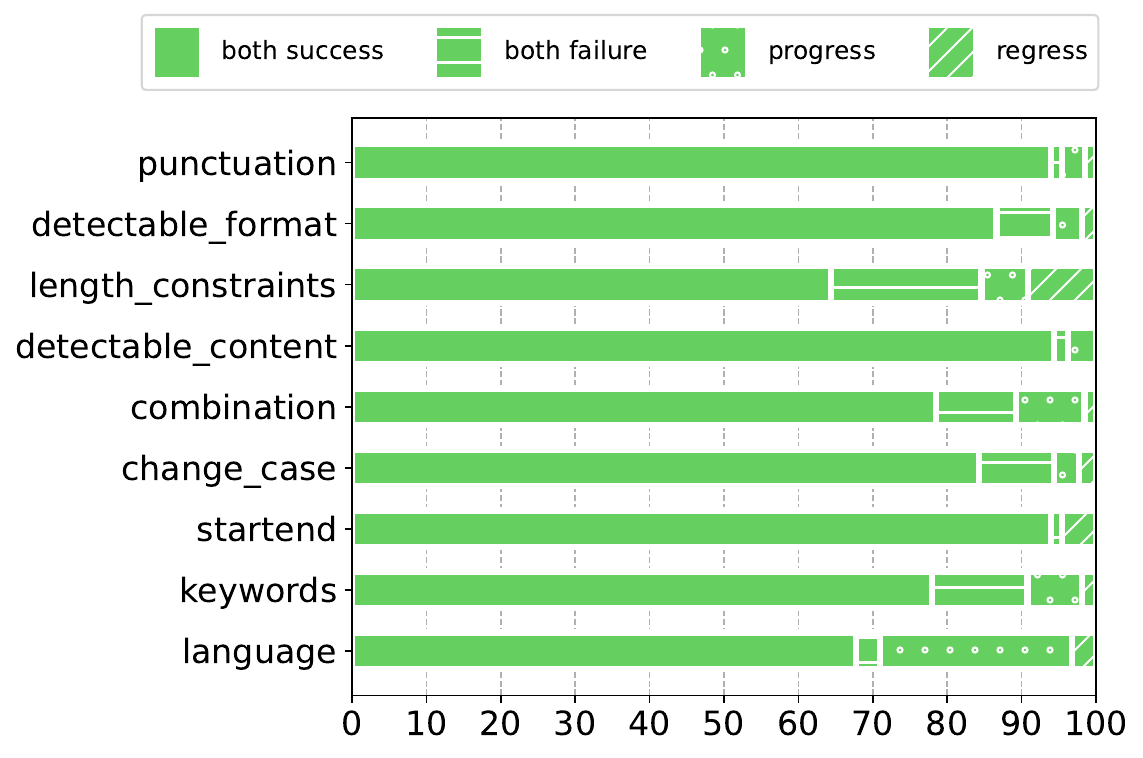} }}%
    \caption{Backward compatibility between \LlamaThreeOne and \LlamaThree for different instruction types in IFEval.}%
    \label{fig:ifeval_llama_backward_compatibility}%
\end{figure}

\begin{figure}[t]
    \centering
{{\includegraphics[width=6.8cm]{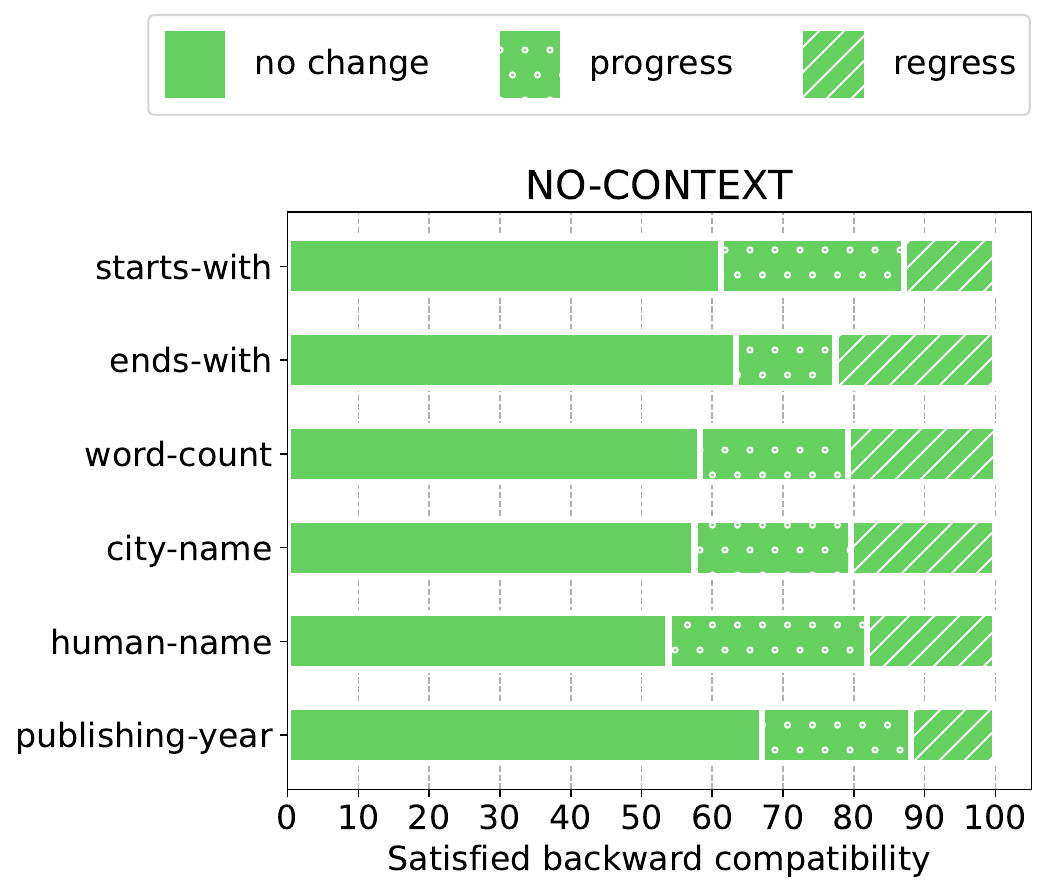} }}%
{{\includegraphics[width=6.8cm]{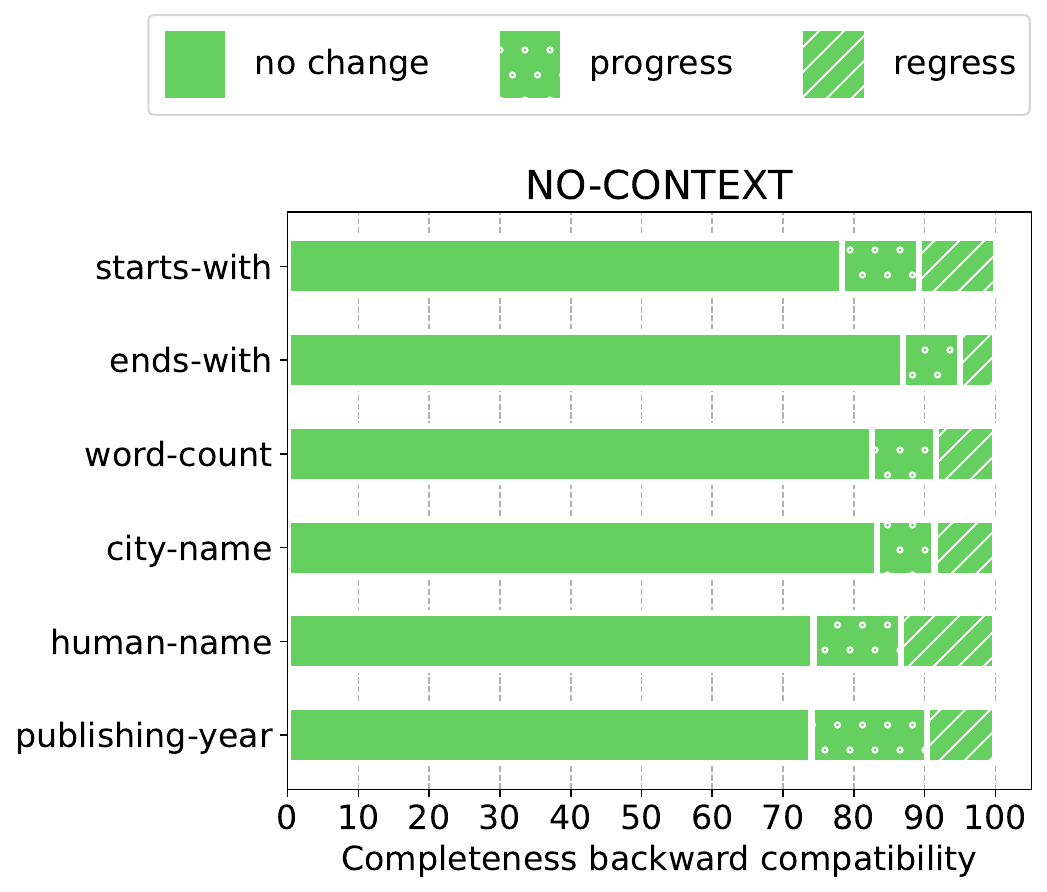} }}%
    \caption{Backward compatibility between \LlamaThreeOne and \LlamaThreeOne for the \textsc{no-context} condition in Kitab, shown for satisfaction rate and completeness. Queries with one book constraint. No change indicates cases where the metric difference between the two models is less than 10 percentage points.}%
    \label{fig:kitab_llama_backward_compatibility}%
\end{figure}

Figures~\ref{fig:ifeval_llama_backward_compatibility}, and \ref{fig:kitab_llama_backward_compatibility} show the backward compatibility analysis for the Llama family. As this is a language only model family, we don not provide and analysis on MMMU for this family.

In IFEval setting, regression impacts only length constraints significantly, while other categories either strongly progress (in case of language instructions) or consistently fail across both models (as in length constraints, case change and keyword constraints).

For the Kitab dataset, regression rates for completeness are lower than for satisfaction rate and at the same time they are also lower than for other model families. This may be an artifact of the two models being of the same size, since completeness indirectly also shows how much information a model can store and access effectively.

\noindent \textbf{\\ Challenging subsets across Models:} 
In addition to enabling analysis of progression and regression within model updates, example level backward comparability can also reveal subsets of data that are consistently challenging across models. In IFEval, models across both versions in Claude, GPT and Llama families fail to follow instructions in $\sim$10\% of the examples in length constraint, keyword constraints and case change constraints. This subset is consistently challenging for all 6 strong models used for analysis and can highlight a set on instructions for future models to improve on. Similarly, in MMMU, the GPT and Claude families of models fail consistently in 15-35\% of examples across different subcategories, which indicates there is a challenging subset of the data for further progress.

\subsection{Main Takeaways}
\noindent\fbox{%
    \parbox{\textwidth}{%
        \begin{itemize}[leftmargin=*]
        \item Backward incompatibility is prevalent across state-of-the-art models reflected in high regression rates for individual examples and at a subcategory level. Detailed analysis here can help in navigating tradeoffs for model selection. 
        \item Different performance metrics may be differently impacted by backward incompatibility. 
        \item Backward incompatibility analysis can also uncover groups of examples for which all models consistently struggle with. For example, in IFEval, models across both versions in Claude, GPT and Llama families fail to follow instructions in $\sim$10\% of the examples in length constraint, keyword constraints and case change constraints.
        \end{itemize}
    }%
}

\section{Related Work and Limitations}
We have presented an extensive evaluation and analysis of several state-of-the-art models on key language and multimodal capabilities. We note that, while many of these tasks are fundamental and core to many real-world applications, several areas that are crucial for exhaustive model evaluation are not covered. We plan to include several of these in future iterations. Here, we give an overview of contemporary capabilities and directions that need to be explored. Parallel to the coverage discussion in terms of capabilities is a separate question on how much data and diversity is needed for each capability for informative and generalizable insights. While we select the representative datasets in \collection to have an interesting and diverse set of subcategories to address this question, some of the capabilities may require more diversity and depth than others because of their broad definition. It is however important to note that merely adding more datasets to the collection may not directly address diversity; many benchmarks may be correlated~\cite{perlitz2024benchmark,zhang2024inherent} and thus only contribute to increasing cognitive load and lower clarity in results presentation. It has also been shown that aggregating ranks of models across many very diverse benchmarks results in unreliable and unstable overall scores~\cite{zhang2024inherent}. This is why we do not aggregate across the different capabilities covered in this report, but only across experimental conditions and groups of data that are related.

\subsection{Capability Evaluations}

\noindent \textbf{Responsible AI:} A cross-cutting dimension for all capability evaluations is the evaluation of several aspects of model behavior important for the responsible fielding of AI systems. These consideration include the fairness, reliability, safety, privacy, and security of models. While evaluations through the Toxigen dataset capture notions of representational fairness for different demographic groups and, to some extent, the ability of the model to generate safe language despite non-safe input triggers in the prompt, other aspects or nuances of fairness and safety require further evaluation and additional clarity, which we hope to cover in future versions of this report. Evaluation efforts to this end include SorryBench~\cite{xie2024sorry}, CocoNot~\cite{brahman2024art}, DecodingTrust~\cite{wang2023decodingtrust}, TrustLLM~\cite{wang2023decodingtrust}, MLCommons Safety Benchmark 0.5~\cite{vidgen2024introducing}, Cybench~\cite{zhang2024cybench}. A general rising concern on most aspects however is that there is a quick turnaround between these benchmarks being released and then included in content safety filters or in post training datasets. For example, during initial investigations we observed that most models evaluated in this report have safety scores of higher than 97\% in the DoNotAnswer benchmark~\cite{wang2023not}, which is a positive development. However, from an evaluation and understanding perspective, the high score indicates that the benchmark is not sensitive enough to capture differences among models. Finally, responsibility in models that can generate images, video, and audio remains heavily under explored. Of these capabilities, image generation is most studied and evaluated due to the pervasiveness and popularity of models in the past two years~\cite{cho2023dall,naik2023social,luccioni2024stable,seshadri2023bias}.

\noindent \textbf{Multilingual Capabilities:} We note that all results in this report evaluate language capabilities in English. Several previous works have raised the importance of multilingual capabilities as a major generalisation aspect and have built dedicated benchmarks for this purpose~\cite{AhujaDHORJNGSAB23,ahuja2023megaverse,zhang2023m3exam,zhang2023m3exam,zhang2023m3exam}. The work on multilingual evaluation has shown that there exist major discrepancies on model performance between languages. Particular challenges have been noted for low-resource languages, which raises a global fairness concern~\cite{cunningham2024understanding,RameshSC23,li2024culturellm}. In addition, a lack of multilingual understanding has been found to be a factor in challenges with responsible AI: jailbreaking even the most advanced models has been shown to be easier if done via a low-resource language~\cite{DBLP:conf/iclr/0010ZPB24,yong2023low}. Despite these problems and the availability of more advanced and challenging benchmarks, multilingual evaluation in major model releases still focuses on oversimplified settings. For example, MGSM~\cite{DBLP:conf/iclr/ShiSF0SVCTRZ0W23} translates 250 examples from GSM8K in 10 different languages. However, given that GSM8k is subject to saturation and also the fact that the answer to the model is expected to be merely a number with little language around it, the benchmark does not test the quality of the generated language per se. Similar concerns are present for cases when existing multiple-choice benchmarks like MMLU and others are translated from English. 

\noindent \textbf{Reasoning and Planning:} The Kitab and FlenQA datasets for language and GeoMeter and Vision \& Language Understanding present in \framework touch upon different aspects of reasoning such as constraint satisfaction, logical reasoning, and spatial and geometric understanding. There are however several other aspects of reasoning and planning where the evaluation of the model is assessed based on the effectiveness of steps and plans that a model generates to solve a problem~\cite{DBLP:conf/nips/ValmeekamMHSK23,momennejad2024evaluating,silver2022pddl,tian2023macgyver,zheng2024natural}, and that we plan to include in future iterations. 

\noindent \textbf{Multimodal Evaluation:} The overall consensus in the field regarding planning and reasoning is that multi-step planning in language and planning actions in the physical, multimodal world world~\cite{DBLP:conf/corl/RanaHGA0S23,chen2023egoplan} are still nascent. In particular for physical world planning and assistance, more work is needed for assessing reasoning skills on modalities that go beyond image and text to encompass video, audio, and inferences based on fusion across multiple modalities. The fundamental tasks required for supporting true and physical multimodal interaction are numerous, including activity recognition, temporal reasoning, compositional reasoning, summarization and grounding, and event and state detection~\cite{DBLP:conf/icra/SermanetDZXDGCD24,patraucean2024perception,fu2024video,li2024mvbench,DBLP:conf/nips/MangalamAM23,bohus2024isthisit}. Recent evaluations of systems that are designed to assist humans on physical tasks~\cite{bohus2024sigma,bohus2024isthisit} show how the fundamental capabilities that need to be evaluated to support these systems need to be extended with a rich set of subtasks that must be solved to support well-timed and formulated contributions by an AI system (e.g. Is now a good time or state to intervene?) or with capabilities to interpret subtle cues available via a single modality (e.g., face expression or voice tone which cannot be understood via text only). 

Related to end-to-end system evaluation, model-based evaluations are only a small piece of the puzzle, as the larger systems that are designed to assist people in real-world, interactive settings are often composed of complex architectures that call models in the course of their operation. Evaluating the impact and effectiveness systems in human-AI collaborative settings requires rigorous and frequent studies with human subjects who engage with the systems in realistic scenarios. Evaluating models on benchmarks provides an initial indication about granular skills that are important for the larger applications. Beyond narrow measures of capability, the design and operation of these systems needs to consider the multiple dimensions of the user experience, including design of interfaces, workflow design~\cite{liao2024ux,fogliato2022goes}, and overreliance and trust of AI systems~\cite{passi2022overreliance,buccinca2021trust}. 
\subsection{Evaluation methodologies and frameworks}
\noindent \textbf{Model Evaluation Frameworks:} As models improve in performance across wider range of capabilities, model evaluation has moved from evaluating on single task-specific test sets to broad benchmarks covering multiple tasks. Meta benchmarks like SuperGLUE \cite{wang2019superglue}, HELM \cite{liang2023holistic}, BigBench \cite{srivastava2022beyond}, Open LLM Leaderboard \cite{open-llm-leaderboard-v2}, HELM \cite{NEURIPS2023_dd83eada}, MMBench \cite{Liu2023MMBenchIY} have aggregated evaluation sets from multiple tasks to enable broader study of model performance via common evaluations. A promising aspect of building meta evaluations is the standardization of inference and evaluation platform across tasks and models through frameworks like Eleuther Language Model Evaluation Harness \cite{eval-harness}, which provide transparency and reproducibility in the evaluation process.   
Such benchmarks have long been used to establish model performance and state-of-art claims in frontier model reports, but lack visibility into important experimental conditions and subcategories which often influence model selection for downstream applications \cite{EyubogluVSDLD0R22,singla2021understanding,nushi2018towards,barocas2021designing}. Hence, in this report we specifically focused on deeper analysis on specific, important capabilities rather than aggregating across benchmarks to identify overall model rankings. In parallel, platforms like LMSys Chatbot Arena \cite{chiang2024chatbot} conduct large-scale pairwise preference evaluations on more open-ended questions with either humans or LLMs as judges producing real-world user aligned model rankings. Unfortunately, as preferences are binary signals aggregated to a single ELO Rating \cite{elo1978rating}, preference ranking evaluations are unable to produce deeper insights needed for model selection and further development, and often encode subtle user or model biases like preference for assertive or longer outputs \cite{hosking2024human, koo-etal-2024-benchmarking, wang-etal-2024-large-language-models-fair, wu2024style}.

\noindent \textbf{Evaluation Methodology Advances:} \emph{Memorization} is a phenomenon that is closely associated with the saturation of benchmarks themselves. Several works have flagged the importance of considering and evaluating the impact of memorization in models. Understanding and evaluating the influence of memorization is a challenging endeavour given the continuing lack of transparency on training data details. Several research efforts have reported verbatim repetitions from test sets~\cite{bordt2024elephants}, drops in performance when the test dataset itself is recollected or expanded~\cite{zhang2024careful,dekoninck2024constat,yang2023rethinking}, or evidence that the test data is present in training set whenever available~\cite{deng2023investigating,golchin2023data}. While there has been some progress in building tools for distinguishing memorization from models genuinely improving in a capability \cite{nori2023capabilitiesgpt4medicalchallenge}, generalizable methods across modalities and datasets are not yet available. This means that, while we do focus on non-saturated benchmarks, we cannot guarantee that these benchmarks are not present in training sets. However, it is an indication that even if (in worst case) the test data was used for training, the model still is not able to perform well. 

Nevertheless, many datasets in \collection do have a \emph{dynamic and procedural nature} and new samples of the data can be generated in the future. For example, all data in GeoMeter, Image Understanding, Vision Language Understanding, and Kitab can be re generated by using a different set of images, questions (for Vision Language Understanding), or authors (for Kitab). Toxigen instead was created by using an adaptive adversarial decoding technique with a classifier in the loop that continues to find new gaps on a model, however the resulting dataset sample was also manually curated by humans~\cite{toxigen}. It will be interesting in the future to conduct side-by-side comparisons between current and future test data versions. 

The idea of \emph{procedurally and dynamically generating data} either in a controlled way or via a distribution has attracted interest with techniques like DyVal~\cite{DBLP:conf/iclr/ZhuC0GY024}, AutoBencher~\cite{li2024autobencher}, S-Eval~\cite{yuan2024s}. and self-evolving benchmarks~\cite{wang2024benchmark}. An unanswered question is how to ensure that inferences for evaluation purposes are not used as part of the training process itself accidentally or intentionally. For models served behind apis, ensuring non contamination in the long term remains at the discretion of model providers, as the api call itself reveals the test data (also when the test set is private to the evaluator and not public knowledge) even though it may not reveal the ground truth per se. In \collection, we choose to prioritize transparency and reproducibility but it is also important to consider other forms of transparency that do not necessarily require full access to the whole test data. Recent work in evaluation methodology~\cite{liu2024ecbd,lucy2024one} for example provides a process framework and guidance to adapting over time and across tasks, motivating the need for evaluation efforts to adapt and revise both test cases and methods.

Finally, as shown in sections dedicated to non-determinism and backward compatibility, LFMs are sensitive to a myriad of parameters and conditions. \emph{Prompt sensitivity}~\cite{jin2024apeer,zhang2022neural} and \emph{few-shot design}~\cite{nori2023can,parnami2022learning} are amongst the top important ones and where there has been continuous evidence that the actual prompt or in-context examples being used have major impact on measurement. For example, Section~\ref{sec:mmmu} results on MMMU are an illustration of high prompt sensitivity across models. For other benchmarks, we rely on prompts that are well validated and vetted by the authors of such datasets or from pre investigations we did to make sure that model performance is not understated. While it is possible to run extensive experimentation through \framework by changing and reusing prompt templates, we also plan to devise and leverage techniques that can do this in a faster and cheaper manner~\cite{schnabel2024prompts,jin2024apeer}.
\section{Conclusion}
In conclusion, our work with the formulation of \framework and study of a set challenging of benchmarks highlights the critical need for more rigorous and nuanced evaluation of large foundation models (LFMs). Despite significant advancements in AI capabilities, we find that current models exhibit substantial weaknesses across various tasks. The complementary strengths of different models suggest that no single model currently excels across all capabilities, underscoring the importance of continued innovation and targeted improvements guided by detailed considerations of evaluations.

Moreover, our disaggregated approach to evaluation exposes granular failures that traditional, aggregate metrics often overlook. This level of detail is essential for identifying and addressing specific areas where models falter, thereby informing both future model development and the selection of benchmarks that remain relevant and challenging. Insights from our studies of backward compatibility and non-determinism raise important questions about the stability and consistency of AI models, especially as they evolve and are used repeatedly over time. These findings emphasize the need for ongoing, transparent evaluation practices that can adapt to the rapid pace of AI development and fielding.

We developed the \framework to facilitate deeper understandings of current LFMs and to lay the groundwork for supporting more effective and targeted improvements. By making these tools and benchmarks available as open-source resources, we hope to foster a collaborative effort within the AI community to enhance the robustness, transparency, and reproducibility of model evaluations.

\subsubsection*{Acknowledgements}
We would like to thank Ahmed Awadallah, Ece Kamar, Eric Horvitz, John Langford, Rafah Hosn, Saleema Amershi for valuable discussions and guidance throughout the whole timeline of the project. We would also like to thank several colleagues and collaborators that have worked and brainstormed with us on different evaluation efforts, and have informed design and scientific choices we have made in this work: Adam Fourney, Akshay Nambi, Alessandro Stolfo, Allie Del Giorno, Arindam Mitra, Clarisse Simoes, Dan Bohus, Dimitris Papailiopoulos, Forough Poursabzi, Gagan Bansal, Ida Momennejad, Jennifer Neville, Julia Kiseleva, Marah Abdin, Marco Rossi, Mazda Moayeri, Natasha Butt, Rahee Ghosh Peshawaria, Ronen Eldan, Saleema Amershi, Sean Andrist, Shital Shah, Shweti Mahajan, Siddharth Joshi, Suriya Gunasekar, Sunayana Sitaram, Tobias Schnabel, Victor Dibia, and Xin Wang. 
\clearpage
\bibliographystyle{abbrv}
\bibliography{main}

\end{document}